\documentclass[]{template}

\usepackage[utf8]{inputenc}             % allow utf-8 input
\usepackage[T1]{fontenc}                % use 8-bit T1 fonts
\usepackage{url}                        % simple URL typesetting
\usepackage{booktabs}                   % professional-quality tables
\usepackage{multirow}
\usepackage{colortbl}
\usepackage{multicol}
\usepackage{amsfonts}                   % blackboard math symbols
\usepackage{nicefrac}                   % compact symbols for 1/2, etc.
\usepackage{microtype}                  % microtypography
\usepackage[dvipsnames]{xcolor}         % colors

\usepackage{latexsym}
\usepackage{etoc}

\usepackage{graphicx}
\usepackage{float}
\usepackage{subcaption}
\usepackage{wrapfig}
\usepackage{lipsum}

\usepackage{bm}

\usepackage{tabularx} 
\usepackage{ragged2e} 
\newcolumntype{L}{>{\RaggedRight\hangafter=1\hangindent=0em}X}

\usepackage{enumitem}
\definecolor{promptblue}{RGB}{220,235,252}
\definecolor{responsegreen}{RGB}{223,245,229}
\definecolor{analysisred}{RGB}{252,224,224}

\usepackage{amsmath}
\usepackage{amssymb}
\usepackage{mathtools}
\usepackage{amsthm}

\setboolean{logo}{true}    % 设为 true 表示显示 logo；设为 false 则不显示

\usepackage[linesnumbered,ruled,vlined]{algorithm2e}

\hypersetup{
    colorlinks=true,
    linkcolor=red,
    citecolor=Cerulean,
    filecolor=magenta,      
    urlcolor=magenta,
}

\usepackage[capitalize,noabbrev]{cleveref}
\crefname{section}{§}{§§}
\Crefname{section}{§}{§§}

\usepackage{calligra}
\DeclareMathAlphabet{\mathcalligra}{T1}{calligra}{m}{n}

\usepackage{pifont}

\theoremstyle{plain}

\theoremstyle{definition}

\theoremstyle{remark}

\renewcommand{\paragraph}[1]{\vspace{1mm}\noindent\textbf{#1}}

\DeclareCaptionLabelFormat{cont}{#1~#2\alph{ContinuedFloat}}
\usepackage[most]{tcolorbox}
\tcbset{
  promptbox/.style={
    top=10pt,
    colback=lightgray!20,
    colframe=Black,
    colbacktitle=NavyBlue,
    enhanced,
    center,
    attach boxed title to top center={yshift=-0.1in,xshift=0.0in},
    boxed title style={boxrule=0pt,colframe=white,},
  }
}
\newtcolorbox{promptbox}[2][]{promptbox, title=#2,#1}
\usepackage[most]{tcolorbox}

\tcbset{
  takeawaybox/.style={
    enhanced jigsaw,
    breakable,
    top=10pt,
    colback=lightgray!20,
    colframe=Black,
    colbacktitle=BurntOrange,
    center,
    attach boxed title to top center={
      yshift=-0.1in,
      xshift=0in
    },
    boxed title style={
      boxrule=0pt,
      colframe=white,
    },
    title after break={\tcbtitle\ (continued)},
    extras middle and last={
      colbacktitle=BurntOrange
    },
    extras first and middle={
      bottom=1mm
    },
    overlay first and middle={
      \node[
        anchor=south east,
        font=\small\itshape
      ] at ([xshift=-2mm,yshift=1mm]frame.south east){};
    },
  }
}

\newtcolorbox{takeawaybox}[2][]{
  takeawaybox,
  title={#2},
  #1
}

\tcbset{
  observationbox/.style={
    top=10pt,
    colback=lightgray!20,
    colframe=Black,
    colbacktitle=YellowGreen,
    enhanced,
    center,
    attach boxed title to top center={
      yshift=-0.1in,
      xshift=0.0in
    },
    boxed title style={
      boxrule=0pt,
      colframe=white,
    },
  }
}

\newtcolorbox{observationbox}[2][]{observationbox, title=#2, #1}

\usepackage{xspace}

\newcommand{\best}[1]{\cellcolor{green!20}\textbf{#1}}
\newcommand{\second}[1]{\cellcolor{blue!12}#1}
\newcommand\blfootnote[1]{%
  \begingroup
  \renewcommand\thefootnote{}\footnote{#1}%
  \addtocounter{footnote}{-1}%
  \endgroup
}

\usepackage{CJK}

\title{Benchmarking Fine-Grained Image-Text Alignment in Interleaved Multimodal Contexts}

\author[1,2]{Bingli Wang*}
\author[2]{Huanze Tang*}
\author[2]{Haijun Lv}
\author[2]{Zhishan Lin}
\author[2]{Lixin Gu}
\author[1]{Lei Feng$\dagger$}
\author[2]{Qipeng Guo$\dagger$}
\author[2]{Kai Chen}

\affil[1]{Southeast University}
\affil[2]{Shanghai AI Laboratory}

\begin{abstract}
In recent years, Multimodal Large Language Models (MLLMs) have achieved remarkable progress on a wide range of multimodal benchmarks. Despite these advances, most existing benchmarks mainly focus on single-image or multi-image comprehension. In real-world scenarios such as document reading, information is often presented as interleaved multimodel contexts. This requires MLLMs not only to recognize the content of individual images, but also to identify relevant textual and visual evidence, establish fine-grained alignments between them, and reason over these aligned signals in interleaved contexts based on contextual evidence. However, there is still a lack of systematic benchmarks for quantifying the fine-grained understanding ability of MLLMs in interleaved image-text contexts. To fill this gap, we propose \textbf{COHERENCE}, a benchmark designed to evaluate the ability of MLLMs to recover fine-grained image-text correspondences in interleaved multimodal contexts. COHERENCE covers interleaved image-text content from four representative domains and contains 6,161 high-quality questions. Moreover, we perform a six-type error analysis, enabling fine-grained attribution of failures in interleaved image-text understanding to the specific capabilities missing in current MLLMs.
\end{abstract}

\begin{document}

\blfootnote{$\dagger$ Corresponding authors, $*$ Equal contribution.}
\blfootnote{{Source code is available at \href{https://github.com/Katono5/COHERENCE}{GitHub}. Dataset is available at \href{https://huggingface.co/datasets/BingliW/COHERENCE}{Huggingface}}}

\maketitle

\section{Introduction}
Humans construct their understanding of complex concepts not through isolated sensory inputs, but by continuously synthesizing structured, multi-source information. Deep cognitive processing occurs specifically when textual explanations and visual elements are structurally interleaved, requiring the brain to actively build fine-grained connections between textual spans and their corresponding visual representations\cite{Theory, Dual-Coding}. Such complex synthesis and reasoning capabilities have increasingly been extended to artificial intelligence systems. In recent years, Multimodal Large Language Models (MLLMs)~\cite{GLM46,Kimi-K2.5,Qwen3-VL,StepVL, Intern-S1-Pro} have made major progress in multimodal understanding~\cite{MMBench,MME,MMMU,MMMU-Pro,Seed-Bench,ScienceQA,Seed-Bench-v2,II-Bench,CII-Bench} and generation tasks~\cite{MMIE}. However, most existing multimodal evaluation settings treat modalities in a relatively disjointed manner, focusing primarily on traditional Visual Question Answering (VQA)~\cite{VQA} tasks. In these settings, models are typically asked to answer an isolated text question based on one or multiple images acting as standalone context. While this flashcard-style evaluation has helped improve basic visual perception, it fundamentally lacks the structural complexity of true multimodal reasoning. This leaves a significant gap in evaluating how well models can navigate and align the continuous, interleaved image-text formats that are essential for deep comprehension.

In real-world internet environments and user interactions, such as reading news articles, analyzing financial reports with numerous charts, or browsing product pages and tutorials containing both images and text, information is often presented as interleaved contexts of visual and textual content~\cite{MarkupLM,Pix2Struct}. This has made the ability to understand image-text interleaved contexts a crucial requirement for modern multimodal large language models. Reflecting this growing recognition, interleaved image-text data have been increasingly incorporated into the pretraining corpora of recent MLLMs~\cite{StepVL,Qwen3-VL,Kimi-K2.5}. However, effectively modeling such complex contexts remains challenging. In image-text interleaved scenarios, information from a single modality is often insufficient. MLLMs must accurately identify the relationships between specific textual spans and corresponding images in long multimodal contexts, integrate fragmented evidence from different sources for joint reasoning, and generate answers strictly grounded in the provided context, rather than relying on parametric knowledge acquired during pretraining, which may lead to hallucinations~\cite{ObjectHallucination,FaithScore,HallusionBench}.

To address the lack of systematic evaluation for alignment and reasoning in interleaved contexts, we propose COHERENCE, a large-scale benchmark specially designed to evaluate the fine-grained understanding ability of MLLMs in interleaved image-text contexts. Different from previous evaluations~\cite{MuirBench,MIBench,MMIU} that simply place several images together or use multi-turn conversations over images, COHERENCE is designed to better capture the structure and challenges of complex interleaved image-text contexts.  The benchmark covers four representative complex domains and contains 6,161 high-quality questions. These questions are used to study two key alignment abilities of models in interleaved image-text understanding: 
\textbf{(i) global image-text alignment}, primarily reflected by \textbf{exact match}, which tests whether the model can capture the overall cross-modal structure and coherence of the full interleaved context; and 
\textbf{(ii) local image-text alignment}, primarily reflected by \textbf{partial match}, which tests whether the model can resolve fine-grained references between textual mentions and specific images and extract locally relevant information. Beyond these two alignment abilities, we perform a six-type error analysis to provide a more comprehensive understanding of model's underlying capability deficiencies in interleaved image-text settings. Based on extensive experimental evaluations, we summarize our key contributions as follows:

\begin{itemize}
    \item \textbf{A New Benchmark.} We present \textsc{COHERENCE}, the first interleaved text-image understanding benchmark, covering four representative domains and consisting of 6,161 high-quality examples.
    \item \textbf{A Systematic Evaluation Framework.} We propose a systematic evaluation framework that disentangles global and local alignment in interleaved image-text understanding. This decomposition enables fine-grained error analysis and systematic attribution of model failures to specific capability deficits, yielding interpretable insights into the limitations of current MLLMs.
    \item \textbf{Key Empirical Findings on MLLM Capabilities.} We perform a comprehensive study of both open-source and closed-source models, and obtain three main findings:
    \begin{enumerate}
        \item While models of different sizes already show strong capability in local image-text alignment, global image-text alignment over complex interleaved multimodal contexts appears to be an emergent ability that arises only at larger scales.
        \item Compared with LLaVA-style modular MLLMs that attach a pre-trained vision encoder to an existing LLM via an adapter, native MLLMs jointly trained from scratch on both text-only and multimodal data generally perform better on complex image-text alignment tasks.
        \item A clear gap still exists between the strongest open-source and closed-source models. For instance, Qwen3.5-397B-A17B, the best-performing open-source model, achieves \textbf{64.81} on \textsc{COHERENCE}, while Gemini-3.1-Pro-Preview-Thinking reaches \textbf{71.82}.
    \end{enumerate}
\end{itemize}

\section{Related Work}
\subsection{MLLM}

Early multimodal large language models (MLLMs) mainly followed a modular design, where a pretrained vision encoder was coupled with a large language model, and training relied largely on image-text pair data. Representative examples include Flamingo~\cite{Flamingo}, which introduced cross-attentional visual conditioning to handle arbitrarily interleaved image-text sequences, and BLIP-2~\cite{BLIP-2}, which used a lightweight Query Transformer to bridge frozen vision and language backbones. Building on this paradigm, models such as LLaVA~\cite{Llava} and IDEFICS2~\cite{IDEFICS2} further improved general-purpose multimodal capabilities through visual instruction tuning and better data construction. More recently, research has increasingly incorporated naturally interleaved image-text documents into the pretraining stage, as reflected in datasets\cite{MINT-1T,OmniCorpus,OBELICS} and studies\cite{StepVL,Qwen3-VL,GLM46}. Along with this shift in training data, the field has also begun moving beyond modular pipelines toward more native multimodal designs, where images and text are modeled in a more unified manner~\cite{Chameleon,ANOLE,GPT-4o,Kimi-K2.5}. Overall, this evolution reflects a broader trend in MLLM research: from modular systems trained on paired data, to pretraining with interleaved documents, and gradually toward native architectures designed for interleaved multimodal understanding.

\subsection{MLLM Benchmark}

Existing benchmarks have significantly expanded multimodal evaluation, yet most general-purpose MLLM benchmarks (e.g., MMBench~\cite{MMBench}, SEED-Bench~\cite{Seed-Bench}, MMMU~\cite{MMMU}, MMStar~\cite{MMStar}) primarily focus on single-image or short-context settings. Recent efforts have begun to explore more complex scenarios, including multi-image understanding (MuirBench~\cite{MuirBench}, MIBench~\cite{MIBench}, MMIU~\cite{MMIU}), long-context evaluation (MileBench~\cite{MileBench}), and interleaved multimodal reasoning (LLaVA-Interleave Bench~\cite{LLaVA-NeXT-Interleave}, MMIE~\cite{MMIE}, InterleavedBench~\cite{InterleavedBench}). However, these benchmarks mainly assess high-level capabilities such as reasoning or generation, and many tasks can still be partially solved using unimodal cues~\cite{Lost-in-the-Middle}, lacking a systematic evaluation of fine-grained image-text alignment in interleaved contexts~\cite{BLINK,Mementos}. COHERENCE is designed to address this limitation by explicitly targeting fine-grained grounding and context-dependent reasoning in complex interleaved multimodal settings.

\section{COHERENCE}
\begin{figure}
    \centering
    \includegraphics[width=1\linewidth]{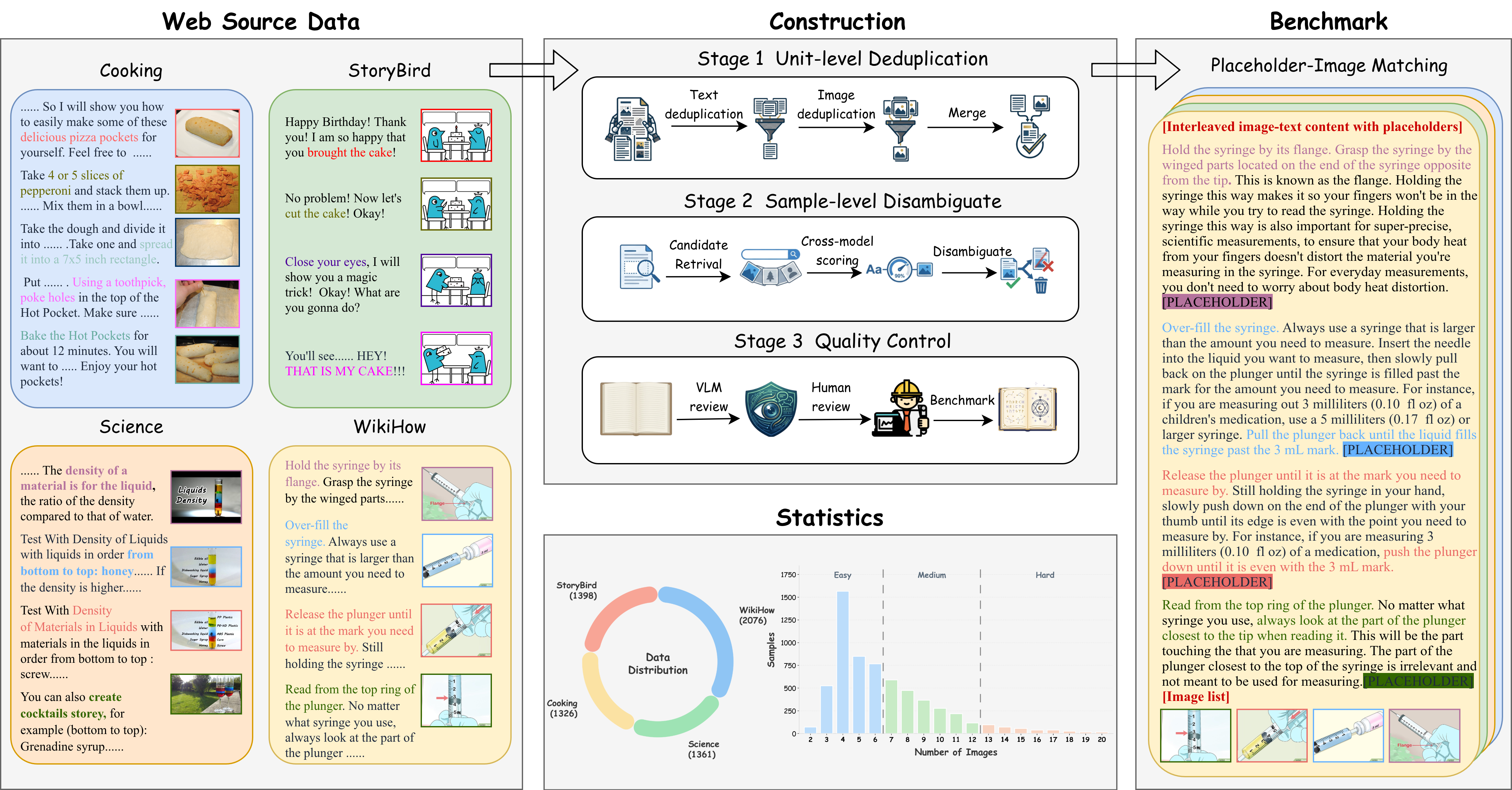}
    \caption{Overview of the COHERENCE benchmark construction pipeline. Starting from naturally interleaved image-text documents collected, we transform the original data into a structured interleaved sequence matching task and apply a three-stage filtering pipeline.}
    \label{fig:overview}
\end{figure}

\subsection{Benchmark Overview}
COHERENCE is designed to evaluate the ability of MLLMs to achieve both local image-text semantic alignment and global consistency understanding in complex interleaved multimodal contexts. Unlike traditional single-image visual question answering settings, where the input consists of an isolated image and its associated question, COHERENCE considers complex contexts composed of multiple images and text segments interleaved together. The evidence required to solve a given instance is often distributed across different textual segments and multiple candidate images. Consequently, models must establish stable image-text correspondences at the global context level and answer questions by grounding their reasoning in contextual evidence. An overview of the COHERENCE benchmark construction pipeline and the dataset statistics is shown in Figure~\ref{fig:overview}.

\subsubsection{Task Definition}
We study the problem of \emph{global vision-language coherence modeling} in complex interleaved multimodal contexts. Unlike traditional single-image visual question answering (VQA), the input to our task is a sequence of alternating text segments and images:
\[
C = (T_1, I_1, T_2, I_2, \dots, T_n, I_n),
\]
where \(T_i \in \mathcal{T}\) denotes the \(i\)-th text segment and \(I_i \in \mathcal{I}\) denotes its corresponding image.

To characterize global image-text alignment ability, we transform the original context into a structured interleaved sequence matching task. Specifically, we replace each image \(I_i\) with a placeholder \(\langle p_i \rangle\), yielding
\[
\tilde{C} = (T_1, \langle p_1 \rangle, T_2, \langle p_2 \rangle, \dots, T_n, \langle p_n \rangle).
\]
Meanwhile, the images are randomly permuted to form a candidate sequence
\[
\mathbf{I}^{\mathrm{cand}} = (I_{\sigma(1)}, I_{\sigma(2)}, \dots, I_{\sigma(n)}),
\]
where \(\sigma\) is a permutation over \(\{1,\dots,n\}\).
The goal of the model is to recover a bijection
\[
\pi : \{1,\dots,n\} \to \{1,\dots,n\},
\]

which is mathematically equivalent to learning the inverse permutation \(\pi = \sigma^{-1}\).

\subsubsection{Benchmark Construction}
We construct the COHERENCE dataset from multi-source interleaved image-text corpora following CoMM\cite{CoMM}, resulting in 6,161 instances. The data span diverse semantic structures and reasoning patterns, increasing both task diversity and generalization difficulty. The dataset includes four domains. WikiHow focuses on everyday commonsense and procedural knowledge, emphasizing procedural steps and local causal relations. StoryBird centers on narrative content, highlighting narrative coherence and cross-segment dependencies. We further derive two subsets, Cooking and Science, from the Instructables web data. Cooking focuses on recipes and procedures, requiring fine-grained action–visual grounding, while Science consists of experimental and procedural materials, emphasizing structured reasoning and alignment between abstract concepts and visual evidence.
More detailed analyses of domain differences are provided in Appendix~\ref{app:domain-analysis}. In addition, we partition the benchmark into three subsets: \emph{easy}, \emph{medium}, and \emph{hard}, based on the number of images per sequence.

Given an original interleaved document \(C = \{T_1, I_1, \dots, T_n, I_n\}\), we construct a matching instance \((\tilde{C}, \mathbf{I}^{\mathrm{cand}})\) through the following process:

\begin{enumerate}
    \item \textbf{Placeholder Substitution.}  
    We remove all images and replace them with indexed placeholders \(\langle p_i \rangle\), while preserving the original contextual structure, yielding \(\tilde{C}\).

    \item \textbf{Candidate Pool Construction.}  
    We collect all images \(\{I_i\}_{i=1}^n\) and randomly shuffle them to form an unordered candidate pool \(\mathbf{I}^{\mathrm{cand}}\), thereby eliminating positional bias.

    \item \textbf{Instance Formation.}  
    We use \((\tilde{C}, \mathbf{I}^{\mathrm{cand}})\) as the input and the original alignment relations as answers, thereby constructing a global matching task.
\end{enumerate}

\subsubsection{Quality Control}
To ensure the uniqueness, identifiability, and controllable difficulty of the dataset, we design a three-stage filtering pipeline to systematically refine the raw constructed samples.

\paragraph{Stage 1: Unit-Level Uniqueness.}
We first enforce a deduplication constraint within each sample. Specifically, for every interleaved document, we remove samples that contain repeated text segments or duplicated images. This step ensures that each text unit and each image candidate is unique within the same sample.

\paragraph{Stage 2: Semantic Identifiability.}
We then filter out semantically ambiguous samples to ensure that each instance admits a clear and unique matching solution. In practice, we use multiple MLLMs to score and verify candidate alignments, and remove samples in which multiple candidate matchings receive similarly high confidence or where the correct alignment depends on vague, underspecified, or weak semantic cues. This stage improves answer uniqueness and ensures that the task evaluates alignment ability rather than annotation ambiguity.

\paragraph{Stage 3: Difficulty Calibration.}
Finally, we calibrate sample difficulty to construct a benchmark with strong discriminative power. We first use multiple MLLMs to automatically assess the difficulty of the constructed samples, and filter out instances that are either too easy or too difficult based on model performance. We then conduct manual inspection as a final quality-control step to verify the effectiveness of the automatic filtering and remove remaining problematic cases. Through this process, we retain samples within an appropriate difficulty range, so that the final benchmark remains both challenging and capable of distinguishing models with different levels of multimodal understanding ability.

\section{Experiments}
\subsection{Experiment Setup}
\noindent\textbf{Evaluated Models.}
To systematically evaluate the COHERENCE benchmark, we conduct large-scale comparative experiments on both open-source and closed-source MLLMs, aiming to cover representative differences in model scale, architectural design, and capability profiles. Furthermore, we analyze only those models whose multimodal training pipelines are sufficiently documented. Among them, we distinguish between two broad multimodal modeling paradigms: \emph{modular MLLMs} and \emph{native MLLMs}. In general, modular MLLMs pretrain the vision encoder and the language model separately, and then connect them through additional cross-modal alignment or fusion training, whereas native MLLMs adopt a more unified multimodal training pipeline from an earlier stage.

\noindent\textbf{Evaluation.}
We cast our evaluation task as a structured sequence matching problem. Given the placeholder-based interleaved context \(\tilde{C}\) and candidate image sequence \(\mathbf{I}^{\mathrm{cand}}\), a model \(M\) predicts an assignment:
\[
\hat{\pi}_M = \arg\max_{\pi} \mathcal{S}_M(\tilde{C}, \mathbf{I}^{\mathrm{cand}}, \pi),
\]
where \(\pi\) is a bijection between placeholders and candidate images, and \(\mathcal{S}_M\) measures global coherence.

We evaluate with two complementary metrics. \emph{Exact Match} (EM) checks whether \(\hat{\pi}_M\) exactly equals the ground-truth assignment \(\pi^*\), reflecting the global image-text alignment:
\[
\tau_{\mathrm{EM}}(\hat{\pi}_M, \pi^*) = \mathbb{I}[\hat{\pi}_M = \pi^*].
\]

To capture partial correctness, we compute Kendall's Tau~\cite{Kendall} between \(\hat{\pi}_M\) and \(\pi^*\), and linearly rescale it to \([0,1]\) as the \emph{Partial Match} (PM) score:
\[
\tau(\hat{\pi}_M, \pi^*) = \frac{\#\text{concordant} - \#\text{discordant}}{\binom{n}{2}}, \qquad
\tau_{\mathrm{PM}}(\hat{\pi}_M, \pi^*) = \frac{\tau(\hat{\pi}_M, \pi^*) + 1}{2}.
\]
A higher PM score indicates better local relative ordering consistency.

\subsection{Main Results}
We conduct comprehensive experiments on COHERENCE, the results are summarized in Table \ref{tab:coherence-main-results} and \ref{tab:coherence-difficulty-results}. The evaluation prompt is provided in Appendix \ref{app:prompt}. We analyze the performance of the models from three perspectives:

\begin{table*}[t]
    \centering
    \caption{Performance of different models on COHERENCE. \textbf{Exact} denotes exact-match accuracy, and \textbf{Partial} denotes Kendall-based partial score. Within each group, \textcolor{green!70!black}{\textbf{Green}} cells indicate the best performance in each column, while \textcolor{blue!70!black}{\textbf{blue}} cells indicate the second-best performance.}
    \label{tab:coherence-main-results}
    \resizebox{0.98\linewidth}{!}{
    \begin{tabular}{lcccccccccc}
        \toprule
        \multirow{2}{*}{\textbf{Model}} 
        & \multicolumn{2}{c}{\textbf{WikiHow}} 
        & \multicolumn{2}{c}{\textbf{StoryBird}} 
        & \multicolumn{2}{c}{\textbf{Cooking}} 
        & \multicolumn{2}{c}{\textbf{Science}} 
        & \multicolumn{2}{c}{\textbf{Overall}} \\
        \cmidrule(lr){2-3} \cmidrule(lr){4-5} \cmidrule(lr){6-7} \cmidrule(lr){8-9} \cmidrule(lr){10-11}
        & \textbf{Exact} & \textbf{Partial}
        & \textbf{Exact} & \textbf{Partial}
        & \textbf{Exact} & \textbf{Partial}
        & \textbf{Exact} & \textbf{Partial}
        & \textbf{Exact} & \textbf{Partial} \\
        \midrule
        \rowcolor{gray!15}
        \multicolumn{11}{c}{\textbf{Open-Source Models}} \\
        \midrule
        Qwen3-VL-4B-Instruct & 58.57 & 85.47 & 22.46 & 63.89 & 26.24 & 71.01 & 17.19 & 64.83 & 34.28 & 72.90 \\
        Qwen3-VL-8B-Instruct & 57.42 & 85.82 & 24.18 & 68.03 & 31.30 & 74.42 & 17.56 & 70.09 & 35.45 & 75.86 \\
        Qwen3-VL-30B-A3B & 62.67 & 87.35 & 27.68 & 71.25 & 52.26 & 80.47 & 34.39 & 75.03 & 46.24 & 79.50 \\
        Qwen3-VL-235B-A22B & 64.45 & 88.72 & 32.12 & 75.09 & 46.83 & 80.92 & 32.77 & 78.12 & 46.32 & 81.61 \\
        Step-VL-10B & 57.47 & 84.91 & 25.25 & 70.14 & 42.84 & 76.66 & 26.60 & 71.18 & 40.19 & 76.75 \\
        GLM-4.6V & 62.96 & 87.84 & 29.18 & 71.98 & 40.27 & 78.00 & 29.83 & 75.00 & 43.09 & 79.29 \\
        Intern-S1-Pro & 64.26 & 88.56 & 28.90 & 72.50 & 52.34 & 80.66 & 35.19 & 76.55 & 47.25 & 80.56 \\
        Kimi-K2.5 & \best{75.67} & \best{93.26} & 46.21 & 82.75 & 58.97 & 83.12 & \second{52.09} & \second{84.43} & \second{60.19} & \second{86.74} \\
        Qwen3.5-4B & 62.76 & 88.58 & 36.41 & 77.86 & 43.14 & 78.04 & 31.45 & 76.24 & 45.64 & 81.16 \\
        Qwen3.5-35B-A3B & 69.70 & 90.69 & 48.21 & 82.80 & 54.15 & 81.75 & 44.09 & 82.00 & 55.82 & 85.06 \\
        Qwen3.5-122B-A10B & \second{71.87} & \second{91.89} & \second{49.93} & \second{83.91} & \second{61.54} & \second{84.08} & 48.79 & 83.38 & 59.57 & 86.52 \\
        Qwen3.5-397B-A17B & 69.85 & 91.19 & \best{55.58} & \best{86.09} & \best{71.27} & \best{87.63} & \best{60.32} & \best{87.14} & \best{64.81} & \best{88.37} \\
        \midrule
        \rowcolor{gray!15}
        \multicolumn{11}{c}{\textbf{Closed-Source Models}} \\
        \midrule
        doubao-seed-2-0-mini-260215 & 70.81 & 91.71 & 37.98 & 78.61 & 43.89 & 78.34 & 37.62 & 79.99 & 50.24 & 83.27 \\
        doubao-seed-2-0-lite-260215 & 72.74 & 92.36 & 44.35 & 82.04 & 51.58 & 78.75 & 45.78 & 81.97 & 55.79 & 84.79 \\
        doubao-seed-2-0-pro-260215 & 76.59 & \second{93.60} & 52.22 & 79.81 & 63.27 & 83.38 & 49.01 & 77.68 & 62.10 & 84.76 \\
        Claude-sonnet-4-6-thinking & 74.18 & 92.71 & 52.72 & \second{84.88} & 73.15 & 88.42 & 58.93 & \second{87.05} & 65.72 & \second{88.76} \\
        GPT-5.2 & 68.50 & 84.68 & 37.34 & 60.59 & 55.28 & 78.02 & 40.78 & 71.36 & 52.46 & 74.84 \\
        GPT-5.4-high & \second{77.50} & 88.77 & \second{57.37} & 76.42 & \best{79.19} & \best{91.83} & \best{68.41} & \best{88.39} & \second{71.29} & 86.54 \\
        Gemini-3.1-pro-preview-thinking & \best{80.73} & \best{94.70} & \best{63.59} & \best{88.52} & \second{73.45} & \second{88.71} & \second{65.10} & 86.11 & \best{71.82} & \best{90.11} \\
        \bottomrule
    \end{tabular}
    }
\end{table*}

\subsubsection{Performance Comparison Across Different Model Scales}
\begin{observationbox}{Finding 1}
Compared with larger models, small models are more prone to losing global information in long interleaved image-text scenarios, making it difficult for them to maintain a coherent overall understanding.
\end{observationbox}
We use the \emph{Partial score} to reflect models' ability in local image-text alignment, and find that model performance does not improve with scale to the same extent as observed for Exact Match. We analyze the scaling behavior of three model families, namely Qwen3-VL~\cite{Qwen3-VL}, Qwen3.5\cite{Qwen3.5}, and doubao-seed-2.0. Taking Qwen3-VL as an example, the 235B model improves over the 4B model by only 8.71 in Partial score. In the Qwen3.5 series, the gain from 4B to 397B-A17B is similarly limited to 7.21, while the doubao-seed-2.0 series shows only a 1.49 improvement. By contrast, the corresponding gains in Exact Match reach 12.04, 19.17, and 11.86, respectively. These results suggest that the advantage of larger models lies more in avoiding global mismatches and maintaining overall consistency, rather than in substantially improving local alignment judgments themselves. Smaller models already exhibit relatively strong local cross-modal alignment ability, while the gains brought by scaling are primarily reflected in stronger global context integration, rather than more fine-grained local grounding ability.

\subsubsection{Performance Comparison Between Modular and Native MLLMs}
\begin{observationbox}{Finding 2}
Compared with modular MLLMs, native MLLMs trained end-to-end demonstrate stronger in-context learning ability and better global image-text alignment in complex interleaved multimodal contexts.
\end{observationbox}
To compare the performance of modular and native MLLMs, we evaluate a range of representative models from both categories. The native MLLMs in our evaluation include models such as Kimi K2.5, the Qwen3.5 series, and Gemini-3.1-pro-preview-thinking, while the modular MLLMs include representative models such as the Qwen3-VL~\cite{Qwen3-VL} series, Step3-VL and Intern-S1-Pro~\cite{Intern-S1-Pro}. The overall results show that end-to-end trained native MLLMs exhibit a clear advantage on this benchmark. Even a relatively smaller native multimodal model, such as Qwen3.5-35B-A3, significantly outperforms the much larger Qwen3-VL-235B. Similarly, Kimi K2.5 also surpasses Intern-S1-Pro, despite their comparable model scales. In addition, Gemini-3.1-pro-preview-thinking achieves the best overall performance on this benchmark, further confirming the advantage of native MLLMs under this task setting.

\subsubsection{Performance Comparison Between Open-Source and Closed-Source Models}
\begin{observationbox}{Finding 3}
Closed-source models still maintain an overall lead on this benchmark, but open-source models represented by the Qwen3.5 series and Kimi K2.5 have already demonstrated substantial progress.
\end{observationbox}
From the overall comparison between open-source and closed-source models, the best-performing closed-source models, such as Gemini-3.1-pro-preview-thinking and GPT-5.4-high, still maintain a clear performance gap over most open-source models. For complex-context interleaved image-text understanding, closed-source models continue to hold certain advantages in training paradigms, the quality of interleaved multimodal data, cross-modal alignment optimization, and long-context modeling capabilities.

It is also important to note the substantial progress made by open-source models in recent years. Open-source models represented by the Qwen3.5 series and Kimi K2.5 have already demonstrated strong competitiveness on this benchmark, indicating that the open-source community is rapidly approaching the capability frontier of closed-source models in multimodal modeling, long-context training, and cross-modal alignment. Nevertheless, there remains considerable room for improvement for open-source models.

\subsection{The Effect of Context Length on Task Difficulty}
\begin{figure}[h]
    \centering
    \begin{subfigure}[b]{0.48\linewidth}
        \centering
        \includegraphics[width=\linewidth]{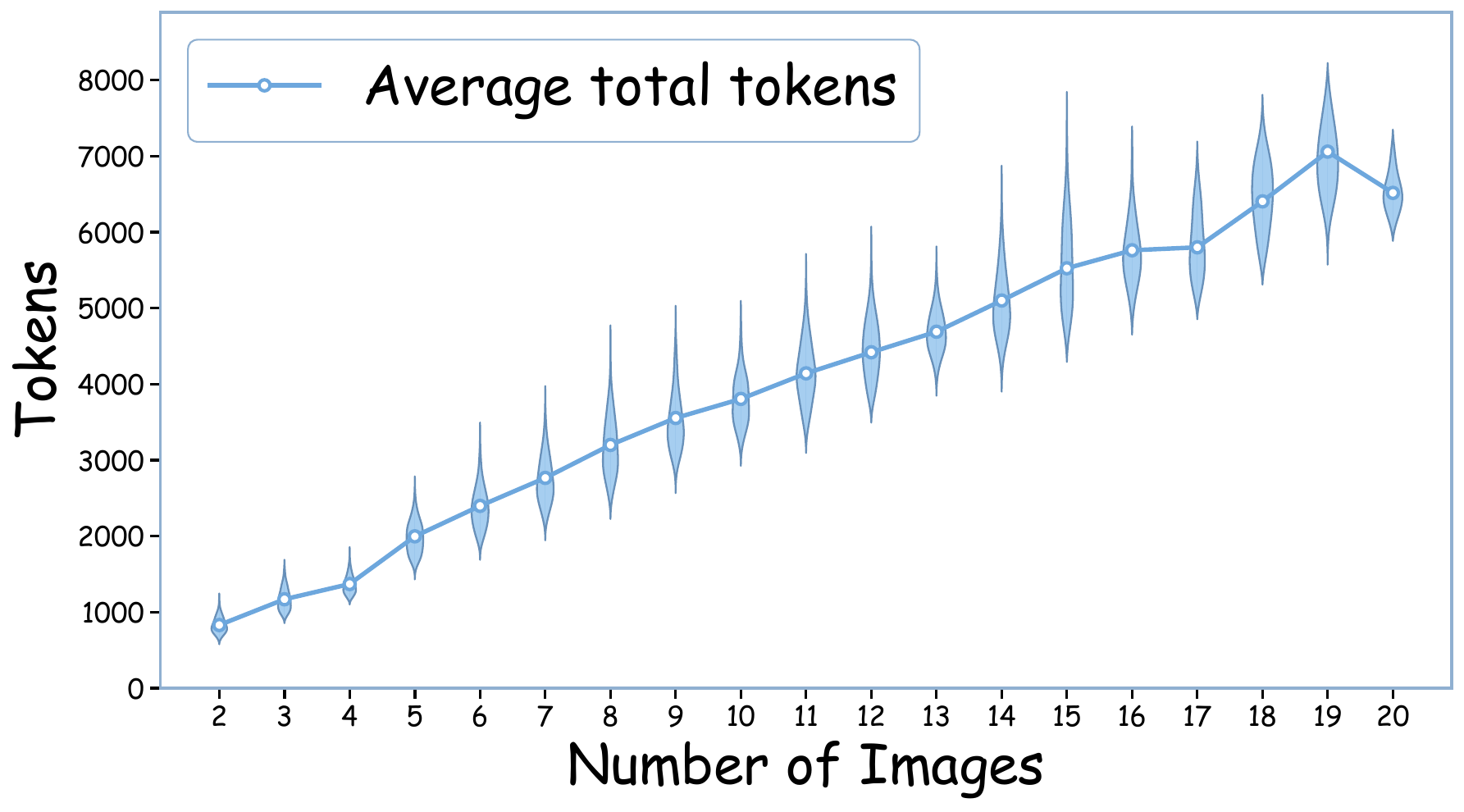}
        \caption{Token count distribution by number of images.}
        \label{fig:token_by_image_count}
    \end{subfigure}
    \hfill
    \begin{subfigure}[b]{0.48\linewidth}
        \centering
        \includegraphics[width=\linewidth]{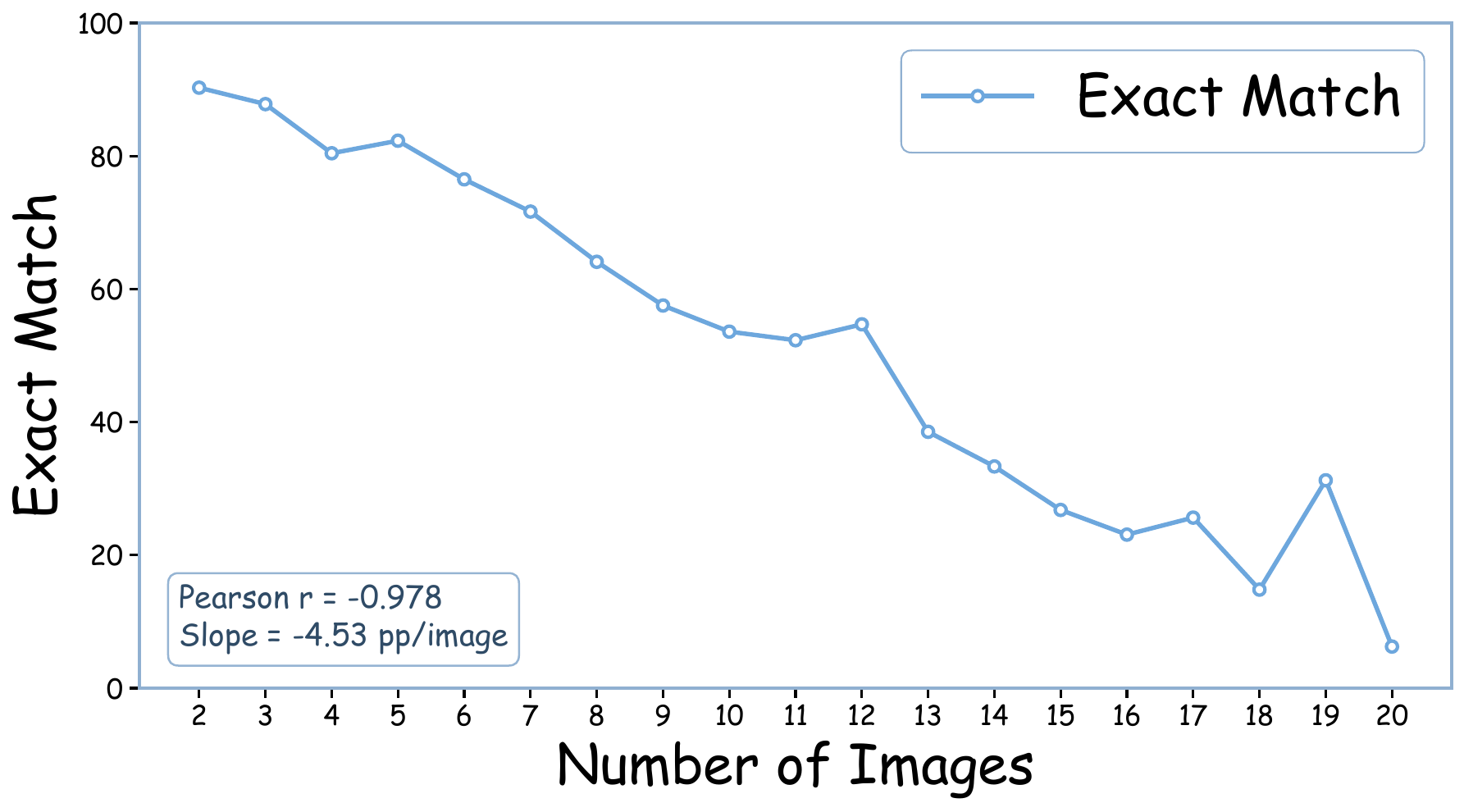}
        \caption{Exact-match accuracy distribution by number of images.}
        \label{fig:difficulty_by_image_count}
    \end{subfigure}
    \caption{\textbf{The effect of context length on task difficulty.} Exact-match accuracy
  decreases as image count increases (Pearson r = -0.983; slope = -4.51 pp/
  image), indicating higher task difficulty with more images.}
    \label{fig:context_length_analysis}
\end{figure}
Since context length is hard to define directly in interleaved multimodal documents, we use image count as a proxy. Figure~\ref{fig:token_by_image_count} shows that overall context length increases consistently with the number of images\footnote{We estimate the total multimodal token count using the official Qwen3-VL AutoProcessor.}, validating image count as an effective indirect indicator. 

Based on this proxy, we study how context length affects task difficulty by evaluating Gemini-3.1-pro-preview-thinking on samples with different numbers of images. As shown in Figure~\ref{fig:difficulty_by_image_count}, Exact Match accuracy decreases steadily as the number of images increases, indicating that the task becomes progressively harder with longer context~\cite{Lost-in-the-Middle,LongContextTransferFromLanguageToVision,LongLLaVA}. Motivated by this trend, we divide the benchmark into three subsets---\emph{easy}, \emph{medium}, and \emph{hard}---according to empirical difficulty. Results are reported in Table~\ref{tab:coherence-difficulty-results}.

\begin{table*}[t]
    \centering
    \caption{Performance of different models under different difficulty levels on COHERENCE. \textbf{Exact} denotes exact-match accuracy, and \textbf{Kendall} denotes Kendall-based partial score. Within each group, \textcolor{green!70!black}{\textbf{green}} cells indicate the best performance in each column, while \textcolor{blue!70!black}{\textbf{blue}} cells indicate the second-best performance.}
    \label{tab:coherence-difficulty-results}
    \resizebox{0.98\linewidth}{!}{
    \begin{tabular}{lcccccccc}
        \toprule
        \multirow{2}{*}{\textbf{Model}}
        & \multicolumn{2}{c}{\textbf{Easy}}
        & \multicolumn{2}{c}{\textbf{Medium}}
        & \multicolumn{2}{c}{\textbf{Hard}}
        & \multicolumn{2}{c}{\textbf{Overall}} \\
        \cmidrule(lr){2-3} \cmidrule(lr){4-5} \cmidrule(lr){6-7} \cmidrule(lr){8-9}
        & \textbf{Exact} & \textbf{Kendall}
        & \textbf{Exact} & \textbf{Kendall}
        & \textbf{Exact} & \textbf{Kendall}
        & \textbf{Exact} & \textbf{Kendall} \\
        \midrule
        \rowcolor{gray!15}
        \multicolumn{9}{c}{\textbf{Open-Source Models}} \\
        \midrule
        Qwen3-VL-4B & 44.46 & 76.18 & 20.83 & 71.62 & 3.32 & 45.81 & 34.28 & 72.90 \\
        Qwen3-VL-8B & 44.62 & 77.00 & 23.64 & 76.24 & 5.82 & 61.76 & 35.45 & 75.86 \\
        Qwen3-VL-30B-A3B & 55.78 & 81.56 & 34.60 & 78.47 & 11.91 & 63.69 & 46.24 & 79.50 \\
        Qwen3-VL-235B-A22B & 55.83 & 82.00 & 34.80 & 81.53 & 11.63 & 77.94 & 46.32 & 81.61 \\
        Step-VL-10B & 50.82 & 78.16 & 26.80 & 76.20 & 4.16 & 65.10 & 40.19 & 76.75 \\
        GLM-4.6V & 52.57 & 80.23 & 31.05 & 79.33 & 11.63 & 69.15 & 43.09 & 79.29 \\
        Intern-S1-Pro & 57.60 & 82.48 & 34.90 & 80.43 & 8.31 & 61.19 & 47.25 & 80.56 \\
        Kimi-K2.5 & 68.15 & 87.20 & \second{51.58} & \second{86.47} & \second{25.21} & \second{83.52} & \second{60.19} & \second{86.74} \\
        Qwen3.5-4B & 54.66 & 81.83 & 34.60 & 80.75 & 13.30 & 76.37 & 45.64 & 81.16 \\
        Qwen3.5-35B-A3B & 64.31 & 85.68 & 46.00 & 84.73 & 22.16 & 80.37 & 55.82 & 85.06 \\
        Qwen3.5-122B-A10B & \second{68.84} & \second{87.46} & 49.65 & 85.67 & 18.28 & 81.46 & 59.57 & 86.52 \\
        Qwen3.5-397B-A17B & \best{71.57} & \best{88.84} & \best{58.19} & \best{88.36} & \best{31.30} & \best{83.61} & \best{64.81} & \best{88.37} \\
        \midrule
        \rowcolor{gray!15}
        \multicolumn{9}{c}{\textbf{Closed-Source Models}} \\
        \midrule
        doubao-seed-2-0-mini-260215 & 59.72 & 83.75 & 38.80 & 82.93 & 15.24 & \second{80.09} & 50.24 & 83.27 \\
        doubao-seed-2-0-lite-260215 & 64.47 & 85.71 & 46.00 & 84.13 & 19.94 & 78.87 & 55.79 & 84.79 \\
        doubao-seed-2-0-pro-260215 & 70.83 & 86.95 & 52.12 & 83.35 & 26.87 & 69.71 & 62.10 & 84.76 \\
        Claude-sonnet-4-6-thinking & 74.06 & \second{89.64} & 57.11 & \best{87.98} & 26.87 & \best{84.01} & 65.72 & \second{88.76} \\
        GPT-5.2 & 62.53 & 77.05 & 40.23 & 73.51 & 15.79 & 59.15 & 52.46 & 74.84 \\
        GPT-5.4-high & \second{78.88} & 88.45 & \best{63.67} & 85.49 & \best{34.63} & 72.56 & \second{71.29} & 86.54 \\
        Gemini-3.1-pro-preview-thinking & \best{81.27} & \best{92.44} & \second{61.85} & \second{87.68} & \second{29.09} & 79.45 & \best{71.82} & \best{90.11} \\
        \bottomrule
    \end{tabular}
    }
\end{table*}

\subsection{Error Analysis}
To better understand model failures on \textsc{COHERENCE}, we define and quantify six representative error types. These categories are not only descriptive of \emph{where} models fail, but also diagnostic of \emph{which underlying capabilities} are lacking.

\begin{itemize}
    \item \textbf{Global Assignment Drift}: the model captures local correspondences but fails to maintain global consistency across the interleaved context, indicating weaknesses in long-context alignment and global planning.

    \item \textbf{Step-State Confusion}: the model understands the overall content but confuses adjacent or visually similar steps or states, indicating limited fine-grained step-state discrimination.

    \item \textbf{Fine-Detail Miss}: the model overlooks subtle but important local cues, reflecting deficiencies in detail perception and fine-grained visual grounding.

    \item \textbf{Semantic Over-Interpretation}: the model forces image--text alignment beyond what the evidence supports, revealing weak evidence-constrained semantic calibration.

    \item \textbf{Visual Hallucination}: the model hallucinates nonexistent objects, attributes, or relations and reasons from them, reflecting poor visual faithfulness.

    \item \textbf{Instruction Violation}: the output fails to satisfy task requirements, such as repeated image use or invalid formats, indicating weaknesses in instruction following and structured output control.
\end{itemize}

\begin{figure}[h]
    \centering
    \begin{subfigure}[b]{0.48\linewidth}
        \centering
        \includegraphics[width=\linewidth]{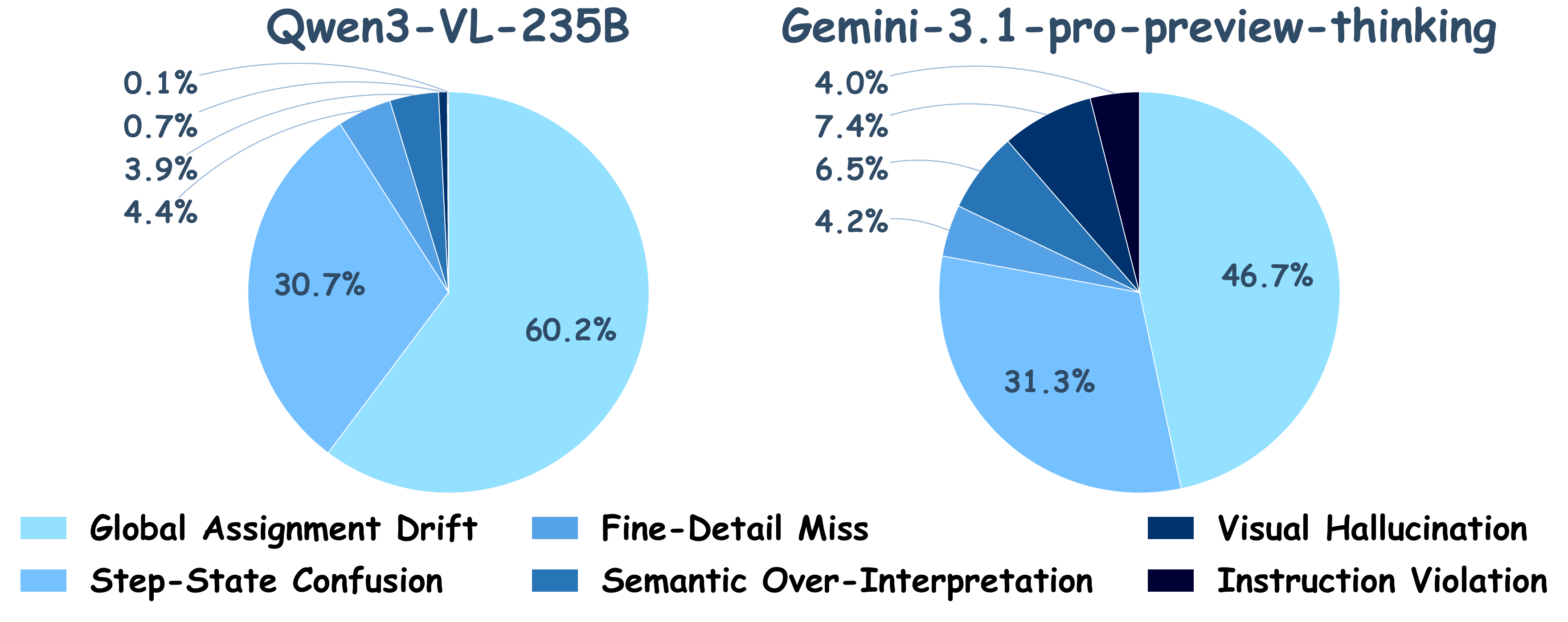}
        \caption{Error Type Distribution}
        \label{fig:error_analysis_qwen}
    \end{subfigure}
    \hfill
    \begin{subfigure}[b]{0.48\linewidth}
        \centering
        \includegraphics[width=\linewidth]{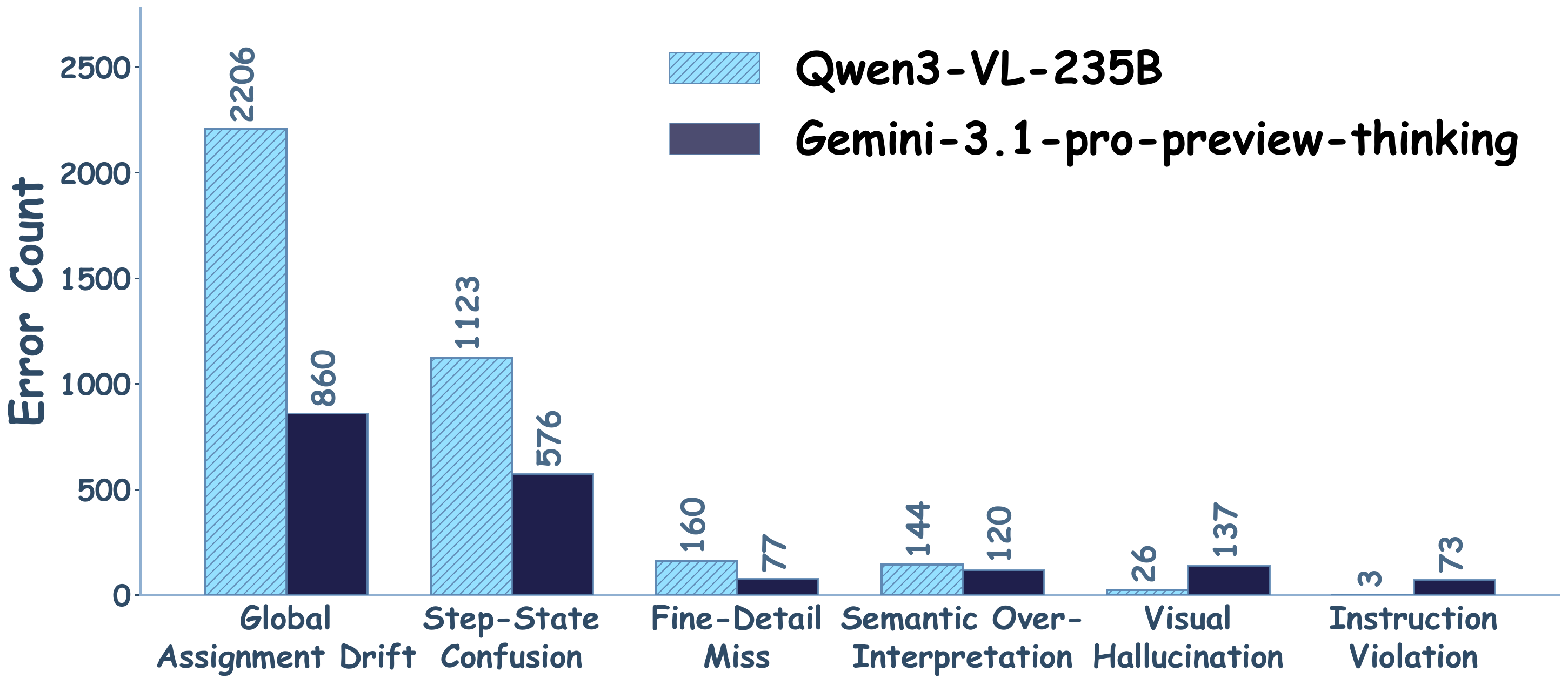}
        \caption{Error Count Comparison by Type}
        \label{fig:error_analysis_gemini}
    \end{subfigure}
    \caption{Error analysis of Qwen3-VL-235B and Gemini-3.1-pro-preview-thinking on COHERENCE.}
    \label{fig:error_analysis}
\end{figure}

For a more fine-grained error analysis, we compare Qwen3-VL-235B, a strong open-source modular MLLM, with Gemini-3.1-pro-preview-thinking, a strong closed-source native MLLM. This comparison provides a more mechanistic view of model failures and highlights the limitations of current systems. The results show that Gemini-3.1-pro-preview-thinking makes substantially fewer \textbf{Global Assignment Drift} errors than Qwen3-VL-235B, suggesting stronger global alignment. Its lower rates of \textbf{Step-State Confusion} \textbf{Fine-Detail Miss} and \textbf{Semantic Over-Interpretation} further indicate better local image-text alignment. However, as a strong reasoning model, Gemini-3.1-pro-preview-thinking is also more prone to losing faithfulness to the original context during long reasoning, which can lead to hallucinated visual interpretations. Consequently, it exhibits higher rates of \textbf{Visual Hallucination} and \textbf{Instruction Violation} than Qwen3-VL-235B. Examples of these error types are provided in Appendix~\ref{appdix:error_cases}.

\section{Discussion}

\subsection{Interleaved Image-Text Understanding is Inherently a Context-Centric Task}
Recent interest in context learning has grown alongside the rise of long-context and agentic systems, where models are increasingly expected to solve tasks by reading, tracking, and integrating information from extended interaction histories, external documents, and tool outputs, rather than relying solely on static parametric knowledge~\cite{ChainOfAgents,AgenticLU,DocAgent,CL-Bench}. 

In this sense, interleaved image-text understanding is naturally a context learning problem, models must locate relevant evidence from the surrounding context, associate visual and textual information across segments, and maintain a coherent interpretation at the document level. This perspective also helps clarify the relationship between COHERENCE and recent benchmarks such as CL-bench~\cite{CL-Bench}. While CL-bench emphasizes whether language models can learn and use new knowledge from context beyond pre-training, COHERENCE extends this paradigm to the multimodal setting and focuses on a complementary challenge: whether models can accurately ground, locate, and associate image-text evidence within interleaved multimodel contexts. We therefore view COHERENCE not only as a benchmark for interleaved multimodal understanding, but also as a step toward evaluating context learning in multimodal environments.

\subsection{COHERENCE as a Controlled Diagnostic Task for Interleaved Understanding}
Open-ended VQA or free-form generation better reflect real-world scenarios for interleaved image-text understanding. However, their unconstrained outputs make model errors difficult to attribute, as failures may arise from multiple sources such as local misalignment, cross-segment reasoning errors, or insufficient global context integration.

COHERENCE instead formulates the task as a structured prediction problem over multimodal elements, converting implicit understanding into explicit and verifiable outputs. This design enables fine-grained and decomposable evaluation across multiple dimensions, including local alignment, cross-segment association, and global consistency. Importantly, this controlled formulation allows systematic isolation and attribution of failure modes in interleaved multimodal understanding. We position COHERENCE as a diagnostic benchmark by providing more interpretable and fine-grained insights into model behavior.

\subsection{Broader Implications for Interleaved Multimodal Understanding}

Interleaved image-text content is increasingly common in real-world applications such as document understanding, web browsing, and multimodal agents~\cite{OBELICS,MINT-1T,OmniCorpus,MM1}, where models must operate over long, structured contexts and continuously integrate information across modalities. In such settings, maintaining fine-grained and context-faithful image-text alignment is critical, as errors in cross-modal grounding can accumulate and propagate through multi-step reasoning.

COHERENCE provides a controlled and quantifiable way to evaluate this capability. Our results show that even state-of-the-art models that perform strongly on traditional single-image VQA and general multimodal benchmarks still have substantial room for improvement in interleaved image-text contexts. In particular, while many models exhibit strong local alignment ability, they often struggle to maintain global consistency across long contexts, highlighting a key limitation. We  hope COHERENCE can facilitate more fine-grained analysis of cross-modal grounding, long-context reasoning, and document-level consistency.

\section{Conclusion}
We introduce COHERENCE, a new benchmark for evaluating fine-grained understanding in complex interleaved image-text contexts. Unlike traditional single-image VQA settings, COHERENCE transforms the otherwise hard-to-diagnose ability of interleaved multimodal understanding into a controlled evaluation task based on image-text correspondence recovery, making it quantifiable, low-noise, and amenable to fine-grained analysis. Systematic experiments on this benchmark show that, although current MLLMs are already capable of processing interleaved inputs, they still exhibit substantial limitations in maintaining global image-text coherence and performing fine-grained cross-modal grounding over complex contexts. We hope COHERENCE will encourage the community to place greater emphasis on the quality of interleaved multimodal understanding, and to view it as one of the key capabilities that next-generation MLLMs must further advance.

\section{Acknowledgments}

This research project was supported by Shanghai Artificial Intelligence Laboratory.

\bibliographystyle{plain}
\bibliography{refs}

\clearpage
\appendix
\section{Benchmark Statistics} \label{app:statistics}
COHERENCE contains 6,161 high-quality interleaved image-text instances with 39,963 images in total, averaging 6.49 images per instance. The benchmark covers four representative domains, including 2076 instances from WikiHow, 1,398 from StoryBird, 1,326 from Cooking, and 1,361 from Science. In terms of empirical difficulty, it consists of 3,774 easy instances, 2,026 medium instances, and 361 hard instances. Table~\ref{tab:dataset-statistics} summarizes the overall benchmark and its key statistics.

\begin{table}[h]
  \centering
  \caption{Dataset statistics of COHERENCE.}
  \label{tab:dataset-statistics}
  \begin{tabular}{lc}
      \toprule
      \textbf{Statistic} & \textbf{Value} \\
      \midrule
      Total instances & 6,161 \\
      Total images & 39,963 \\
      Avg. images per instance & 6.49 \\
      \midrule
      WikiHow & 2,076 \\
      StoryBird & 1,398 \\
      Cooking & 1,326 \\
      Science & 1,361 \\
      \midrule
      Easy & 3,774 \\
      Medium & 2,026 \\
      Hard & 361 \\
      \bottomrule
  \end{tabular}
\end{table}

\section{Domain-Wise Analysis} \label{app:domain-analysis}
COHERENCE exhibits substantial performance variation across domains, suggesting that interleaved image-text understanding is not a single unified ability. WikiHow is relatively easier because image--text correspondence is often supported by explicit local cues. By contrast, Cooking and Science are more challenging because they impose stronger procedural constraints. In these domains, models must not only recognize the local content of an image, but also infer intermediate states and ground fine-grained visual evidence within the broader process. StoryBird presents a different type of difficulty. Although its images are visually simpler and more stylized, successful matching often depends on interpreting visual metaphor and distinguishing highly subtle differences across otherwise similar scenes. As a result, the task relies more heavily on fine-grained cross-modal grounding than on explicit local cues alone.

Taken together, these results suggest that COHERENCE evaluates a spectrum of cross-modal abilities, ranging from local explicit alignment to process-level state tracking and fine-grained contextual grounding.

\section{The Necessity of Multimodal Integration}
COHERENCE is designed to evaluate the comprehensive understanding ability of MLLMs in interleaved image-text settings. A key question, therefore, is whether the task can be solved using only text or only images. To answer this question, we further analyze the performance of Qwen3-VL-235B under unimodal settings. As shown in Table~\ref{tab:ablation-input-settings}, this analysis helps verify whether COHERENCE truly measures interleaved multimodal understanding, rather than degenerating into a single-modality inference problem.

Specifically, we evaluate two degraded settings: text-only, where all images are removed, and image-only, where all text is removed. The text-only setting can be regarded as approximate random guessing, since without visual evidence the model is largely unable to recover the correct image-text correspondence. The image-only setting, in contrast, is used to compare against the full interleaved image-text setting, in order to test whether the model can solve the task merely by relying on temporal or visual continuity among images. We use Exact Match as the evaluation metric throughout. The results show that performance under both text-only and image-only settings is substantially worse than under the full interleaved image-text setting, indicating that correct global matching requires jointly integrating both textual and visual information, rather than relying on a single modality alone.

\begin{table}[h]
  \centering
  \caption{Ablation results of Qwen3-VL-235B under different input settings on COHERENCE (Exact Match).}
  \label{tab:ablation-input-settings}
  \resizebox{0.7\linewidth}{!}{
  \begin{tabular}{lccccc}
      \toprule
      \textbf{Input Setting} & \textbf{WikiHow} & \textbf{StoryBird} & \textbf{Cooking} & \textbf{Science} & \textbf{Overall} \\
      \midrule
      Text-only (random guess) & 4.58 & 3.15 & 2.56 & 1.32 & 3.10 \\
      Image-only & 14.40 & 16.17 & 4.45 & 3.01 & 10.14 \\
      Interleaved-image-text & 64.45 & 32.12 & 46.83 & 32.77 & 46.32 \\
      \bottomrule
  \end{tabular}
  }
\end{table}

\section{Prompt Template} \label{app:prompt}
\subsection{Evaluation Prompt} \label{sec:evaluation-prompt}
\begin{promptbox}{Evaluation Prompt}

\#\# Task: Interleaved-Image-Text Matching

You are given an article about "\{title\}" with \{num\_placeholders\} image placeholders marked as [IMAGE\_PLACEHOLDER]. You are also given \{len(image\_sequence)\} candidate images (Image 0, Image 1, ..., Image \{len(image\_sequence) - 1\}) shown below.

Your task is to determine which image should be placed at each placeholder position based on the surrounding text context.

\#\# Article Text (with placeholders):
\vspace{-0.5em}
\begin{center}
\textbf{\{content\}}
\end{center}
\vspace{-0.5em}

\#\# Candidate Images (Image 0 to Image \{len(image\_sequence) - 1\}):

\vspace{-0.5em}
\begin{center}
\textbf{\{Image list\}}
\end{center}
\vspace{-0.5em}
\#\# Instructions:

1. **Read the text carefully**: Each [IMAGE\_PLACEHOLDER] appears within a specific context. The surrounding text describes what should be shown in that image.

2. **Analyze each placeholder**: For each placeholder (in order from first to last), identify what the nearby text is describing - this tells you what the image should show.

3. **Match images to placeholders**: Look at the \{len(image\_sequence)\} candidate images provided and determine which image best matches the context around each placeholder.

4. **Important**: The same image index can only be used once. Each placeholder needs a different image.

\#\# Output Format:

First reason step by step, then output your final answer on the LAST line as a Python list:

- Format: [\{", ".join(["index" + str(i) for i in range(num\_placeholders)])\}]

- The list position corresponds to the placeholder order (first placeholder is index 0).

- Each value is the image index to place at that placeholder.

- Example: [2, 0, 1, 3, 4] means placeholder 1 uses Image 2, placeholder 2 uses Image 0, etc.

- Do NOT output the inverse mapping (i.e., image -> placeholder).

- The list must have exactly \{num\_placeholders\} integers, each between 0 and \{len(image\_sequence) - 1\}.

Now analyze the text and images, then provide your answer.

\end{promptbox}

\subsection{Error Analysis Prompt}
\label{appendix:error_analysis_prompt}

\begin{promptbox}{Error Analysis Prompt} 
You are an error analyst for interleaved image-text matching. Your task is to perform systematic root-cause analysis for an incorrect prediction, then provide: 

1) one paragraph explaining why the prediction is wrong, 

2) one primary error type, 

3) two secondary error types. 

This is a long-context interleaved image-text assignment task. You must jointly evaluate text evidence, image evidence, and structural constraints. Do not judge from local fragments only. Use exactly one label for primary\_error\_type and exactly two labels for secondary\_error\_types. 

Allowed labels (exact string match) and interpretation criteria: 

\textbf{1. Global Assignment Drift}: Local image-text pairings may look reasonable, but the final mapping is globally inconsistent across the full article (e.g., systematic shift, wrong overall alignment, boundary drift). 

\textbf{2. Step-State Confusion}: The mismatch mainly comes from mixing up nearby steps/states that are semantically close or visually similar. 

\textbf{3. Fine-Detail Miss}: The mismatch is caused by missing decisive fine-grained cues (small objects, subtle state changes, local attributes, tool/action details). 

\textbf{4. Semantic Over-Interpretation}: The mismatch is driven by reading more meaning into an image than the visible evidence supports, then forcing that interpretation into alignment. 

\textbf{5. Visual Hallucination}: The reasoning relies on visual elements that are not actually present in the image. 

\textbf{6. Instruction Violation}: The output breaks task constraints or format requirements (e.g., invalid list format, duplicate image index use, missing/extra assignments, illegal indices). 

[Original Prompt] 
\vspace{-0.5em}
\begin{center}
\textbf{\{prompt\_text (including text\_with\_placeholders and image\_list)\} }
\end{center}
\vspace{-0.5em}

[Model Output] 
\vspace{-0.5em}
\begin{center}
\textbf{\{raw\_output\} }
\end{center}
\vspace{-0.5em}

[Prediction] 
\vspace{-0.5em}
\begin{center}
\textbf{\{prediction\} }
\end{center}
\vspace{-0.5em}

[Gold Answer] 
\vspace{-0.5em}
\begin{center}
\textbf{\{answer\} }
\end{center}
\vspace{-0.5em}

Return exactly one JSON object (no markdown, no explanation text):\\
\{\\
\hspace*{0.7cm}"error\_reason": "one concise paragraph explaining why the prediction is wrong",\\
\hspace*{0.7cm}"primary\_error\_type": "one label from the six",\\
\hspace*{0.7cm}"secondary\_error\_types": ["one label", "one label"]\\
\}
\end{promptbox}

\section{An Extended Setting with Extra Candidate Images}

To further broaden the evaluation scope of \textsc{COHERENCE}, we consider an extended setting in which the number of candidate images is larger than the number of image placeholders. In the standard formulation, each instance contains a one-to-one correspondence between placeholders and candidate images, and the model is only required to recover the correct placeholder--image assignment. While this setting already evaluates fine-grained alignment under interleaved image--text context, it does not test whether the model can distinguish sequence-relevant images from additional visually plausible but irrelevant candidates. The extended setting is introduced to evaluate this additional aspect of robustness.

Concretely, for an instance with $m$ image placeholders, we augment the original candidate pool with $n$ extra images, where $n \in \{1,2,3\}$. These extra images do not belong to the target sequence, but are presented together with the original candidates during inference. The model is therefore required to solve a broader assignment problem: it must identify which images are actually relevant to the interleaved context, assign those relevant images to the correct placeholder positions, and avoid using the irrelevant ones. Compared with the standard formulation, this variant places stronger demands on both local image--text discrimination and global consistency over the full candidate pool.

To evaluate model behavior under this setting, we report two complementary metrics. \textbf{Exact} measures whether the model recovers the full placeholder assignment correctly, and therefore reflects end-to-end success on the original task. In addition, for settings with $n>0$, we report \textbf{Reject}, which measures whether the model correctly identifies all extra candidate images as irrelevant. This metric is introduced to explicitly capture the model's ability to filter out distractor images, which is not observable in the original one-to-one formulation. Together, these two metrics allow us to separate failures in \emph{placement} from failures in \emph{rejection}.

Table~\ref{tab:coherence-extra-image-results} reports model performance under different values of $n$. This comparison allows us to examine how robust current models remain when the candidate pool becomes increasingly over-complete. In particular, it reveals not only whether models can still recover the correct placeholder assignment, but also whether they can reliably reject irrelevant visual candidates under long interleaved image--text context. As such, this extended setting provides a broader view of model behavior beyond exact one-to-one correspondence recovery.

\begin{table*}[t]
    \centering
    \caption{Performance of models on \textsc{COHERENCE} under different numbers of extra candidate images. Here, $n$ denotes the number of extra candidate images added beyond the number of placeholders. \textbf{Exact} denotes exact-match accuracy, and \textbf{Reject} denotes the accuracy of correctly identifying all irrelevant candidate images. Within each column, \textcolor{green!70!black}{\textbf{green}} cells indicate the best performance, while \textcolor{blue!70!black}{\textbf{blue}} cells indicate the second-best performance.}
    \label{tab:coherence-extra-image-results}
    \resizebox{0.8\linewidth}{!}{
    \begin{tabular}{lccccccc}
        \toprule
        \multirow{2}{*}{\textbf{Model}}
        & \multicolumn{1}{c}{$n=0$}
        & \multicolumn{2}{c}{$n=1$}
        & \multicolumn{2}{c}{$n=2$}
        & \multicolumn{2}{c}{$n=3$} \\
        \cmidrule(lr){2-2} \cmidrule(lr){3-4} \cmidrule(lr){5-6} \cmidrule(lr){7-8}
        & \textbf{Exact}
        & \textbf{Exact} & \textbf{Reject}
        & \textbf{Exact} & \textbf{Reject}
        & \textbf{Exact} & \textbf{Reject} \\
        \midrule
        Qwen3 VL 4B      & 34.28 & 24.99 & 48.41 & 24.07 & 45.09 & 22.79 & 42.36 \\
        Qwen3 VL 8B      & 35.45 & 30.12 & 69.70 & 29.91 & 63.16 & 27.84 & 59.00 \\
        Qwen3 VL 30B     & 46.24 & 39.12 & 71.55 & 38.71 & 66.60 & 36.83 & 62.75 \\
        Qwen3 VL 235B    & 46.32 & 45.61 & \best{91.14} & 43.99 & \best{85.64} & 42.41 & \best{81.01} \\
        Qwen3.5 4B       & 45.64 & 34.86 & 52.80 & 33.74 & 50.92 & 32.79 & 46.27 \\
        Qwen3.5 35B-A3   & 55.82 & 46.88 & 67.08 & 46.36 & 64.79 & 44.86 & 61.89 \\
        Qwen3.5 122B-A10 & \second{59.57} & \second{52.98} & 74.61 & \second{51.23} & 71.22 & \second{50.62} & 69.11 \\
        Qwen3.5 397B-A17 & \best{64.81} & \best{54.36} & \second{75.59} & \best{53.89} & \second{73.69} & \best{52.46} & \second{71.11} \\
        \bottomrule
    \end{tabular}
    }
\end{table*}

\section{Case Study} 
Due to the length of the original prompt, we only show the article text with image placeholders and candidate image list in this section, we also provide ground truth for reference. The detailed evaluation instructions we used can be found in Section~\ref{sec:evaluation-prompt}.

\subsection{Dataset Samples}
\label{appdix:dataset cases}
\clearpage
\thispagestyle{empty}

\begin{observationbox}{Cooking Case 1}
\#\# Article Text (with placeholders):

Pinwheels Samosa are great twist to most popular and most wanted Indian snack *SAMOSA*. They make mouthwatering appetizer. Pinwheel samosa looks delicious and are a great appetizer for party. They are so delicious you may eat them just as they are or serve them with coriander chutney and tomato ketchup.
\vspace{-0.5em}
\begin{center}
[IMAGE\_PLACEHOLDER]
\end{center}
\vspace{-0.5em}
Ingredients FOR CRUST : 1 cup all purpose flour 2 tablespoons fine semolina flour 1/2 teaspoon salt 3 tablespoons oil Approximately 1/3 cup cold water FOR FILLING : 4 medium size potatoes (boiled peeled and roughly mashed) 1/2 cup green peas (boiled) 1 teaspoon cumin seeds 1 tablespoon green chillies (finely chopped) 2 teaspoons coriander powder 1/2 teaspoon red chilly powder 1/2 teaspoon garam masala 1 teaspoon dry mango powder (amchur) 1-1/4 teaspoons salt 2 tablespoons fresh coriander (finely chopped) ALSO NEEDED : 3 tablespoons all purpose flour 1/4  cup water Oil to fry

\vspace{-0.5em}
\begin{center}
[IMAGE\_PLACEHOLDER]
\end{center}
\vspace{-0.5em}

MethodCRUST : Mix flour, salt and oil add water to make soft dough.
add water as needed.
cover it with a damp cloth.
Let the dough rest for at least 15 minutes.
FILLING : Squeeze the water from green peas and mix all the ingredients for filling; potatoes, cumin seeds, coriander powder, green chilly, mango powder, salt, and fresh coriander together.
Divide the filling in 2 parts and set aside.
MAKING PINWHEEL : Mix the all purpose flour and water to make thin batter.
Set aside.
Knead the dough for a minute and divide into 2 equal parts and roll them into balls.
Roll the ball into about 11 inch diameter.
Spread one part of the filling evenly over rolled dough, slightly pressing Roll the sheet gently, but firmly until you have a nice firm log.
Use a little water to seal the end of the sheet firmly,roll the log gently 6-7 times this helps keeping together.
With a sharp knife slice the both ends about 1 inch long then slice the log into ½ inch thick, this should make about 14 -16 pinwheels.
Press each pinwheel lightly.

\vspace{-0.5em}
\begin{center}
[IMAGE\_PLACEHOLDER]
\end{center}
\vspace{-0.5em}

FryingHeat the oil in a frying pan over medium high heat. The frying pan should have at least 1 inch of oil. To check if the oil is ready, drop a small piece of dough. The piece of dough should come up but not change color. Dip the pinwheel in the batter one at a time and slowly drop in the oil in frying pan. Fry the pinwheels in small batches. Pinwheels will take about 3 to 4 minutes to cook. Turn them occasionally. Fry the pinwheels until both sides are golden-brown. Repeat this process. The crispy, delicious pinwheels are now ready to serve.

\vspace{-0.5em}
\begin{center}
[IMAGE\_PLACEHOLDER]
\end{center}
\vspace{-0.5em}

\#\# Candidate Images (Image 0 to Image 3):

\begin{center}
\begin{minipage}[t]{0.23\linewidth}
\centering
Image 0\\[2pt]
\includegraphics[width=\linewidth]{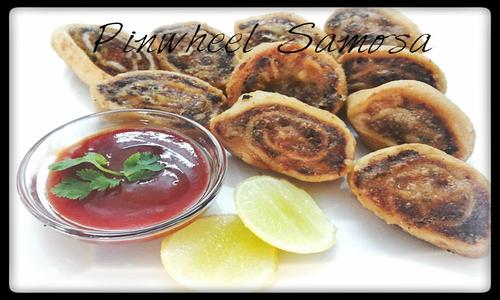}
\end{minipage}
\hfill
\begin{minipage}[t]{0.23\linewidth}
\centering
Image 1\\[2pt]
\includegraphics[width=\linewidth]{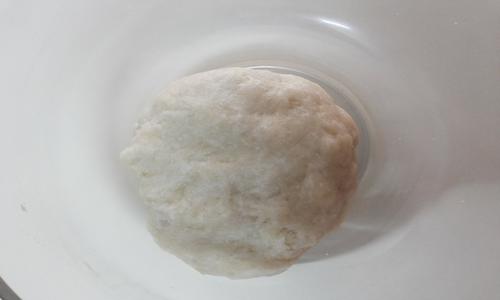}
\end{minipage}
\hfill
\begin{minipage}[t]{0.23\linewidth}
\centering
Image 2\\[2pt]
\includegraphics[width=\linewidth]{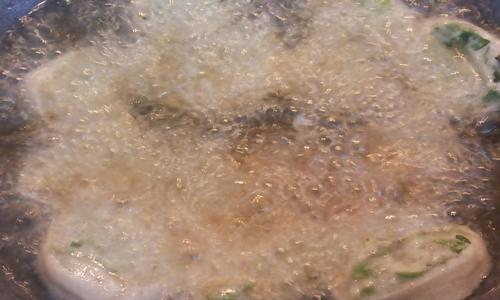}
\end{minipage}
\hfill
\begin{minipage}[t]{0.23\linewidth}
\centering
Image 3\\[2pt]
\includegraphics[width=\linewidth]{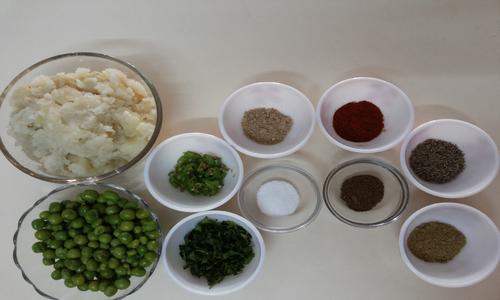}
\end{minipage}
\end{center}

\#\# Answer: [0, 3, 1, 2]
\end{observationbox}

\begin{observationbox}{Cooking Case 2}
\#\# Article Text (with placeholders):

Grandma Hawkins' Poor Man's CakeBack Story: Grandma Hawkins had 13 kids (11 boys) so everything she cooked made a big batch.  This one uses three regular size loaf pans and three little  ones or 5 regular size loaf pans.  Grandma also never really measured  anything.  My Dad really liked this bread so one time followed her  around while she was making it and measured everything and recorded the  recipe.  Dad made this around the Holidays and I have been for many  years now too.  Takes me back in time every time I make it. I always  send some to Mom and my siblings.  Good stuff.
\vspace{-0.5em}
\begin{center}
[IMAGE\_PLACEHOLDER]
\end{center}
\vspace{-0.5em}
Ingredients1 lb raisins cooked and cooled.  Raisins in a pan and add water just to cover the raisins, simmer stirring occasionally for a few minutes, remove from heat cover the pan and set aside to cool.  The raisins will absorb much of the water.  Don't drain, it will all be used. 6 Cups flour (more or less) 2 Cups Sugar 1 Cup shortening or 1/2 cup shortening + 1/2 cup butter 1 Cup Buttermilk or sour milk.  Grandma would make sour milk by adding one Tablespoon vinegar to 1/4 cup of canned milk (NOT sweetened condensed milk) then adding enough water to make 1 Cup. 3 eggs 1/2 teaspoon salt 2 teaspoons Baking soda 2 teaspoons Cinnamon Nuts, cherries, dates to suit

\vspace{-0.5em}
\begin{center}
[IMAGE\_PLACEHOLDER]
\end{center}
\vspace{-0.5em}

MixingMix together the shortening and sugar.
I'm using a stand mixer but it could be done by hand.
The batter gets pretty stiff so it's a bit of a workout.
Once the shortening and sugar are mixed, add in the baking soda, cinnamon and salt and mix to combine Once it's all combined, add the eggs and mix again.
Next add the raisins.
Put them all in there including any liquid that wasn't absorbed, it'll add flavor and keep it moist.
Now add 1/2 the flour and 1/2 the buttermilk/sour milk.
Keep adding a cup or so of flour and a little buttermilk at a time until all the flour and buttermilk are incorporated.
You're going for texture here, stiff batter is what you want.
After all that, throw in a couple hand fulls of chopped nuts.
I'm using walnuts but as a kid we had a hickory tree next door that would drop a lot on our side so we'd use those and they were great too.
I also added green and red candied cherries at this point, 1/2 an 8oz container each.
One could leave out the cherries if you like.
Mix it all up.

\vspace{-0.5em}
\begin{center}
[IMAGE\_PLACEHOLDER]
\end{center}
\vspace{-0.5em}

Prepare for BakingThis is pretty important: Grease the loaf pans REALLY WELL make sure you get down in all the corners or it will stick.  I use the wrapper from the shortening and smear around what stuck to the wrapper, that's usually enough, use butter if you need more. Divide between all your pans and lightly press in just to even it out, like I said, it's a pretty stiff batter.

\vspace{-0.5em}
\begin{center}
[IMAGE\_PLACEHOLDER]
\end{center}
\vspace{-0.5em}

BakeBake at 350F for 50 minutes.  Grandma's original instructions said to check it with a broom straw, I just use a knife stuck in the middle to see if the batter is set or not. Usually the small loaf pans are done at 50 minutes, the large pans use another 10 minutes. This bread freezes well and is great at the holidays or whenever. If you make this, let me know.  If you like it, I'd appreciate a vote in the Baking Contest. Cheers,

\vspace{-0.5em}
\begin{center}
[IMAGE\_PLACEHOLDER]
\end{center}
\vspace{-0.5em}

\#\# Candidate Images (Image 0 to Image 4):

\begin{center}
\begin{minipage}[t]{0.18\linewidth}
\centering
Image 0\\[2pt]
\includegraphics[width=\linewidth]{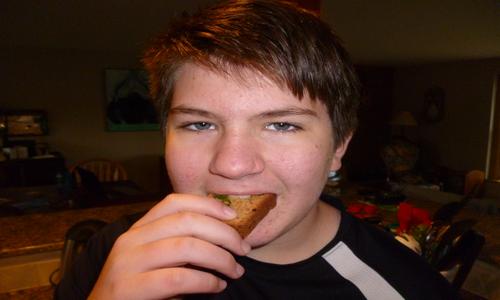}
\end{minipage}
\hfill
\begin{minipage}[t]{0.18\linewidth}
\centering
Image 1\\[2pt]
\includegraphics[width=\linewidth]{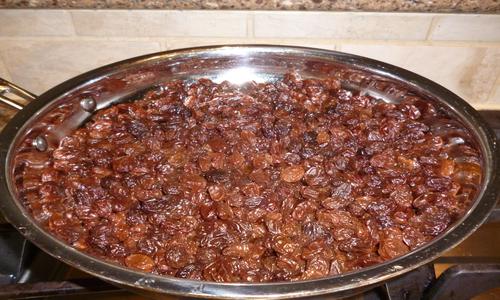}
\end{minipage}
\hfill
\begin{minipage}[t]{0.18\linewidth}
\centering
Image 2\\[2pt]
\includegraphics[width=\linewidth]{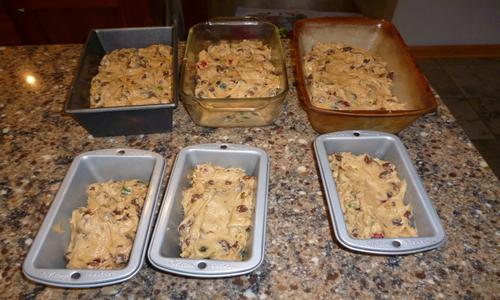}
\end{minipage}
\hfill
\begin{minipage}[t]{0.18\linewidth}
\centering
Image 3\\[2pt]
\includegraphics[width=\linewidth]{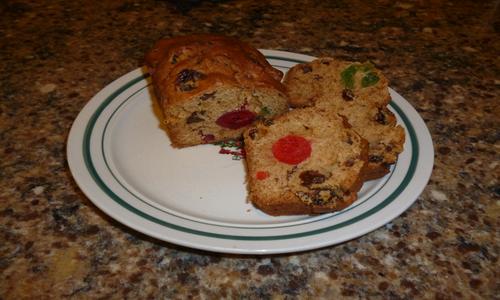}
\end{minipage}
\hfill
\begin{minipage}[t]{0.18\linewidth}
\centering
Image 4\\[2pt]
\includegraphics[width=\linewidth]{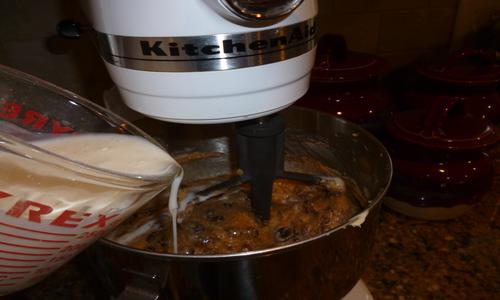}
\end{minipage}
\end{center}

\#\# Answer: [3, 1, 4, 2, 0]
\end{observationbox}

\begin{observationbox}{Cooking Case 3}
\#\# Article Text (with placeholders):

Ginger TeaIt is very easy to make and a healthy drink which gives relief to your tiredness. Benefits  :  Drinking a cup of ginger tea before travelling can help prevent the nausea and vomiting associated with motion sickness. Useful in improving digestion and increasing absorption of food, ginger tea can bloating after eating too much. Ginger contains anti-inflammatory properties that make it an ideal home remedy for muscle and joint problems. In addition to drinking ginger tea, you can also use it to soak inflamed joints. Ginger tea can help relieve congestion associated with the common cold.

\vspace{-0.5em}
\begin{center}
[IMAGE\_PLACEHOLDER]
\end{center}
\vspace{-0.5em}

Ingredients Required 

Ginger Tea leaves Milk Utensils Tea mesh / leaves filter Knife Sugar (if required) Crusher Spoon

\vspace{-0.5em}
\begin{center}
[IMAGE\_PLACEHOLDER]
\end{center}
\vspace{-0.5em}

How to Make

Then take the ginger and peel the skin and cut the ginger into small pieces as shown in pic. Then take the crusher and crush the ginger. Note : It is important to crush the ginger because it gives more flavour/ taste to tea.

\vspace{-0.5em}
\begin{center}
[IMAGE\_PLACEHOLDER]
\end{center}
\vspace{-0.5em}

After crushing the ginger take milk as per your requirement and  put it on flame then put tea leaves \& ginger into the milk After that stir well with the spoon till it boils well.

\vspace{-0.5em}
\begin{center}
[IMAGE\_PLACEHOLDER]
\end{center}
\vspace{-0.5em}

After it is boiled well and becomes dark brown in colour as shown in pic, take it  from the flame . Filter the tea with the mesh and take it in cup, add sugar (as per your requirement), stir well and it is ready to drink.

\vspace{-0.5em}
\begin{center}
[IMAGE\_PLACEHOLDER]
\end{center}
\vspace{-0.5em}

\#\# Candidate Images (Image 0 to Image 4):

\begin{center}
\begin{minipage}[t]{0.18\linewidth}
\centering
Image 0\\[2pt]
\includegraphics[width=\linewidth]{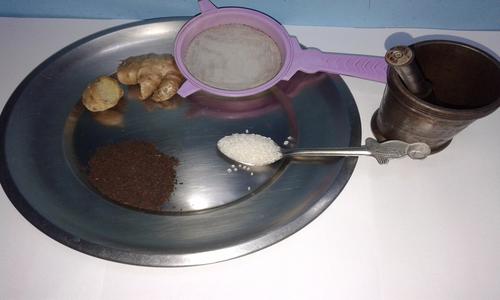}
\end{minipage}
\hfill
\begin{minipage}[t]{0.18\linewidth}
\centering
Image 1\\[2pt]
\includegraphics[width=\linewidth]{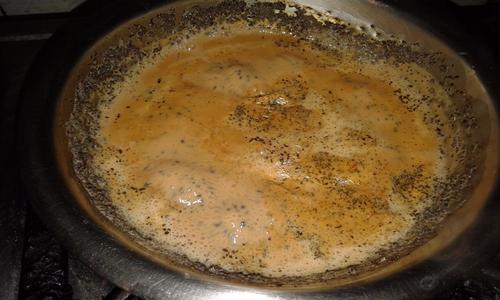}
\end{minipage}
\hfill
\begin{minipage}[t]{0.18\linewidth}
\centering
Image 2\\[2pt]
\includegraphics[width=\linewidth]{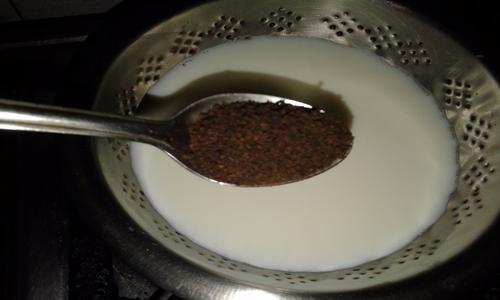}
\end{minipage}
\hfill
\begin{minipage}[t]{0.18\linewidth}
\centering
Image 3\\[2pt]
\includegraphics[width=\linewidth]{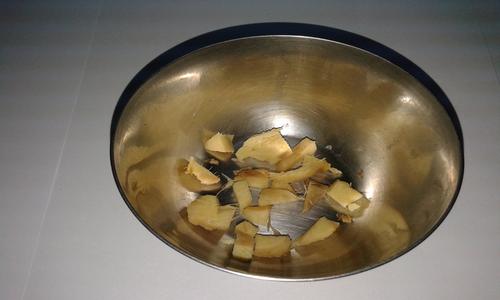}
\end{minipage}
\hfill
\begin{minipage}[t]{0.18\linewidth}
\centering
Image 4\\[2pt]
\includegraphics[width=\linewidth]{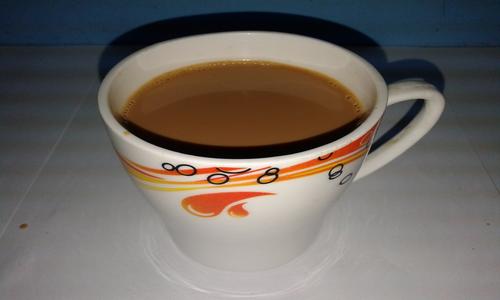}
\end{minipage}
\end{center}

\#\# Answer: [4, 0, 3, 2, 1]
\end{observationbox}

\begin{observationbox}{Cooking Case 4}
\#\# Article Text (with placeholders):

5 Minute Breakfast Quesadilla

I usually snooze longer than I should have and by the time I wake up, I don't have time to make breakfast. Breakfast is an important meal of the day, so I try to make something substantial to power me through until lunch. This recipe is one of the quickest breakfast foods to cook with few ingredient, yet, is still delicious to eat.

\vspace{-0.5em}
\begin{center}
[IMAGE\_PLACEHOLDER]
\end{center}
\vspace{-0.5em}

Ingredients* 

Tortilla or any Wrap  -- I used Pita since that's what I had in my fridge * Egg * Cheese Slice * Salt * Leftover Veggies Kitchen Tools * Pan * Small Bowl * Wooden Spatula

\vspace{-0.5em}
\begin{center}
[IMAGE\_PLACEHOLDER]
\end{center}
\vspace{-0.5em}

Heat Up the Tortilla

* Heat the pan on low heat * When warm place the tortilla on the pan * Cover it * Break the tortilla in half (only half will be used to make one Quesadilla, but you can always make more)

\vspace{-0.5em}
\begin{center}
[IMAGE\_PLACEHOLDER]
\end{center}
\vspace{-0.5em}

Scramble the Egg

* Crack the egg into a small bowl * Scramble it using a whisk or a fork * Pour it to a pan in Medium heat * When the egg begins to cook, use your cooking spatula to break the eggs to smaller pieces * Add salt

\vspace{-0.5em}
\begin{center}
[IMAGE\_PLACEHOLDER]
\end{center}
\vspace{-0.5em}

Adding Cheese

* Break up the cheese slices to smaller pieces by folding it

\vspace{-0.5em}
\begin{center}
[IMAGE\_PLACEHOLDER]
\end{center}
\vspace{-0.5em}

Assembling the Quesadilla

* Add the cooked scrambled eggs to the tortilla or wrap * Add the cheese slices * Heat it in the pan on low heat and cover it for a few second to melt the cheese * Fold the tortilla or wrap * 

[Optional] You can also add vegetables like bell peppers and tomatoes to the quesadilla or any leftover veggies in your fridge. Enjoy your quick 5 minute breakfast meal.

\vspace{-0.5em}
\begin{center}
[IMAGE\_PLACEHOLDER]
\end{center}
\vspace{-0.5em}

\#\# Candidate Images (Image 0 to Image 5):

\begin{center}
\begin{minipage}[t]{0.15\linewidth}
\centering
Image 0\\[2pt]
\includegraphics[width=\linewidth]{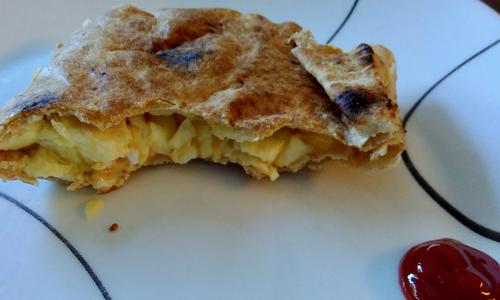}
\end{minipage}
\hfill
\begin{minipage}[t]{0.15\linewidth}
\centering
Image 1\\[2pt]
\includegraphics[width=\linewidth]{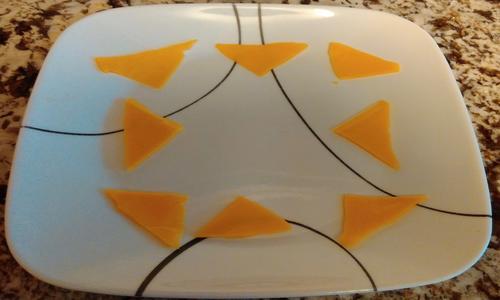}
\end{minipage}
\hfill
\begin{minipage}[t]{0.15\linewidth}
\centering
Image 2\\[2pt]
\includegraphics[width=\linewidth]{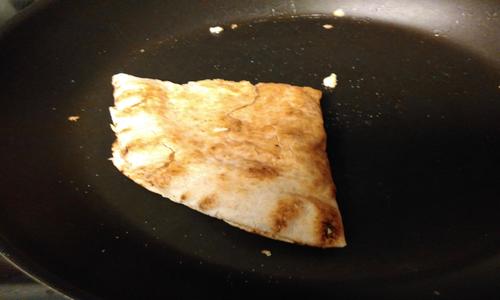}
\end{minipage}
\hfill
\begin{minipage}[t]{0.15\linewidth}
\centering
Image 3\\[2pt]
\includegraphics[width=\linewidth]{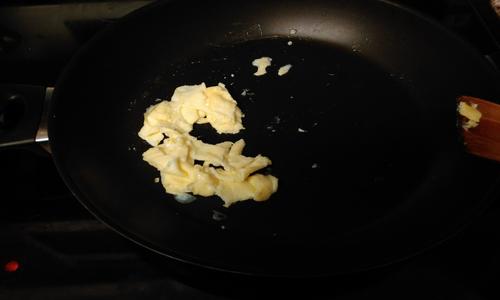}
\end{minipage}
\hfill
\begin{minipage}[t]{0.15\linewidth}
\centering
Image 4\\[2pt]
\includegraphics[width=\linewidth]{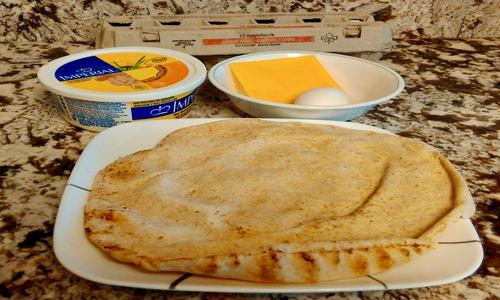}
\end{minipage}
\begin{minipage}[t]{0.15\linewidth}
\centering
Image 5\\[2pt]
\includegraphics[width=\linewidth]{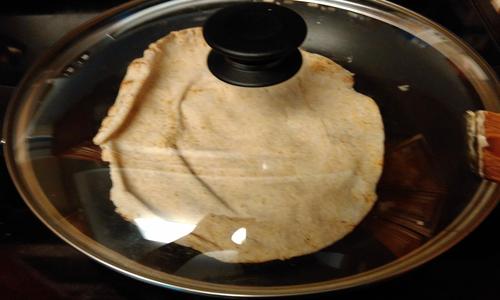}
\end{minipage}
\end{center}

\#\# Answer: [0, 4, 5, 3, 1, 2]
\end{observationbox}

\begin{observationbox}{Cooking Case 5}
\#\# Article Text (with placeholders):

Steamed Salmon on the GrillGrilled salmon is so tasty with just the minimum of adornment, and I've come to rely on a method that is simple, keeps the moisture in, fairly foolproof, and easy to clean up. My mouth waters just thinking about it!

\vspace{-0.5em}
\begin{center}
[IMAGE\_PLACEHOLDER]
\end{center}
\vspace{-0.5em}

Preparing Your Salmon for GrillingCut your pieces into sections or steaks that are about a nice medium thickness and lay them out on a generous sized piece of foil. I like cooking them with some slices of lemon, but it isn't necessary. I have a lemon tree, so that's easy enough.

\vspace{-0.5em}
\begin{center}
[IMAGE\_PLACEHOLDER]
\end{center}
\vspace{-0.5em}

Sealing in the SalmonI like to fold the foil so it overlaps and seals the salmon securely, so it won't drip all over the place and also so it cooks evenly.

Leave the seam near the top so you can check on it when it is nearly ready.

\vspace{-0.5em}
\begin{center}
[IMAGE\_PLACEHOLDER]
\end{center}
\vspace{-0.5em}

Preparing the GrillI like to cook the salmon at about 375 degrees for about 20-25 minutes, depending on the thickness.

When you check it  to see if it's done, be careful not to scald your hand from the steam when you unfold a corner of it.

As it steams in its own juices, you will be overcome by the delicious smells! If it needs more time, simply re-fold and close the lid for a few more minutes.

\vspace{-0.5em}
\begin{center}
[IMAGE\_PLACEHOLDER]
\end{center}
\vspace{-0.5em}

Removing the Salmon From the GrillWhen they are just so, they will be flaky and want to fall apart.
This is where you want to use your spatula (or maybe two) with concentration and dexterity.
Slide it under the pieces and have a tray close at hand for a quick and safe transfer.

Sometimes the fillets need a little separating with a knife, as they tend to bond together a bit as they cook.

By the way, no grill to clean off!

\vspace{-0.5em}
\begin{center}
[IMAGE\_PLACEHOLDER]
\end{center}
\vspace{-0.5em}

The PresentationLay your beautiful grilled/steamed salmon on your nicest platter to give it the respect it deserves. The result is a miracle, and you didn't even work hard! I like to put out a dish of fresh lemon wedges for folks to spritz on the salmon if they prefer. This will be the gorgeous centerpiece of any meal. Happy grilling!

\vspace{-0.5em}
\begin{center}
[IMAGE\_PLACEHOLDER]
\end{center}
\vspace{-0.5em}

\#\# Candidate Images (Image 0 to Image 5):

\begin{center}
% 第一排
\begin{minipage}[t]{0.22\linewidth}
\centering
Image 0\\[2pt]
\includegraphics[width=\linewidth]{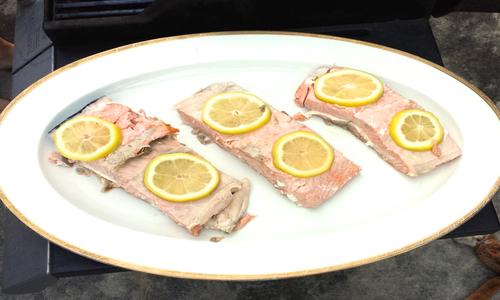}
\end{minipage}
\hfill
\begin{minipage}[t]{0.22\linewidth}
\centering
Image 1\\[2pt]
\includegraphics[width=\linewidth]{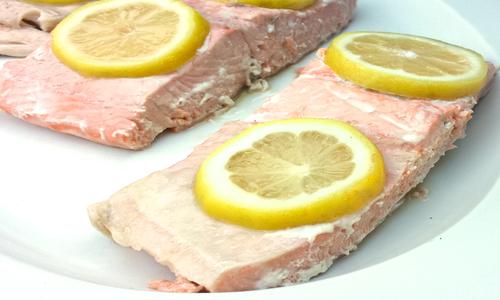}
\end{minipage}
\hfill
\begin{minipage}[t]{0.22\linewidth}
\centering
Image 2\\[2pt]
\includegraphics[width=\linewidth]{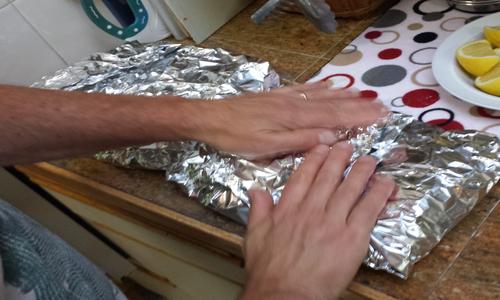}
\end{minipage}

\vspace{0.5em}

% 第二排
\begin{minipage}[t]{0.22\linewidth}
\centering
Image 3\\[2pt]
\includegraphics[width=\linewidth]{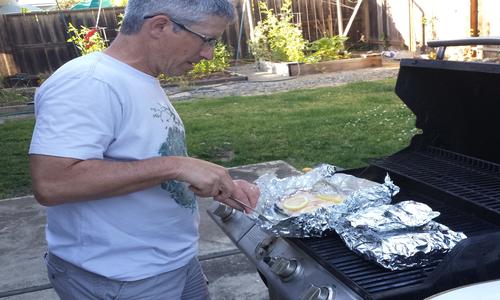}
\end{minipage}
\hfill
\begin{minipage}[t]{0.22\linewidth}
\centering
Image 4\\[2pt]
\includegraphics[width=\linewidth]{sections/TempFigure4/image_04.jpg}
\end{minipage}
\hfill
\begin{minipage}[t]{0.22\linewidth}
\centering
Image 5\\[2pt]
\includegraphics[width=\linewidth]{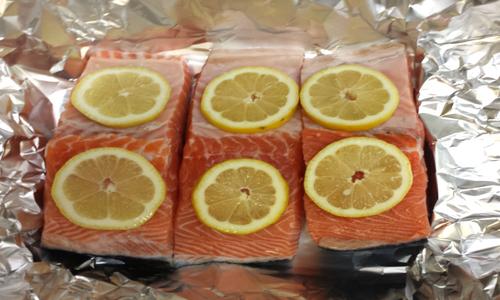}
\end{minipage}
\end{center}

\#\# Answer: [1, 5, 2, 4, 3, 0]
\end{observationbox}

\begin{observationbox}{WikiHow Case 1}
\#\# Article Text (with placeholders):

Play each note slowly when you first start sight reading. Take your time and play each note in the sheet music. Play deliberately and get used to playing off the sheet music. Once you're more comfortable, you can pick up the speed in which you play.

\vspace{-0.5em}
\begin{center}
[IMAGE\_PLACEHOLDER]
\end{center}
\vspace{-0.5em}

Read 2 notes ahead as you play the music. Once you get the notes down, you can start concentrating on rhythm. Always try to read ahead so that you know what comes next, even if you miss a note here or there. This is especially important if you have to flip to the next page. As you get better at sight reading, try to read several notes or beats ahead of what you're playing.  You'll know that you're ready to read ahead when you can comfortably sit in front of a new piece of music and follow along with the notes as you play.

\vspace{-0.5em}
\begin{center}
[IMAGE\_PLACEHOLDER]
\end{center}
\vspace{-0.5em}

Play the piece all the way through without stopping. Skip over the notes that you missed and try to get back on the rhythm. Continue to follow the sheet music with your eyes as you play.

\vspace{-0.5em}
\begin{center}
[IMAGE\_PLACEHOLDER]
\end{center}
\vspace{-0.5em}

Keep your eyes on the music and don’t look down at your hands. Feel for the keys to ensure that your hands are in the right position. Use your hearing to recognize if you're off-key rather than looking down at your hands.  It will take time to learn to keep your eyes on the music, but this gets easier as you gain experience with sight reading.

\vspace{-0.5em}
\begin{center}
[IMAGE\_PLACEHOLDER]
\end{center}
\vspace{-0.5em}

Ignore more complex note commands until you feel comfortable. The slower you go, the more time you’ll have to hit each key and the more accurate you'll be as you sight read. On the music, there may be tempo markings telling you to drastically change your rhythm. Ignore these markings until you're proficient at sight reading. In addition to these markings, there may also be a notation above each note called an articulation. You should also ignore these.   Tempo markings will often be found on the top and to the left of the notes.  Some examples of tempo markings include allegro (brisk), presto (very fast), moderato (moderately fast), grave (slow/solemn), and lento (slow).  A small dot above the note is called a staccato and signifies that the note should be shorter in duration. This is an example of articulation.  A slur is an articulation that looks like a curved line written above the notes. When you see this articulation it means that you shouldn't put any beats or spaces in between the notes.

\vspace{-0.5em}
\begin{center}
[IMAGE\_PLACEHOLDER]
\end{center}
\vspace{-0.5em}

Play another piece of sheet music. Once you're done playing the first piece of music, switch to a different piece and start the process over. Don't go back and try to play the first piece perfectly because that builds muscle memory, not sight reading skills.

\vspace{-0.5em}
\begin{center}
[IMAGE\_PLACEHOLDER]
\end{center}
\vspace{-0.5em}

\#\# Candidate Images (Image 0 to Image 5):

\begin{center}
% 第一排
\begin{minipage}[t]{0.15\linewidth}
\centering
Image 0\\[2pt]
\includegraphics[width=\linewidth]{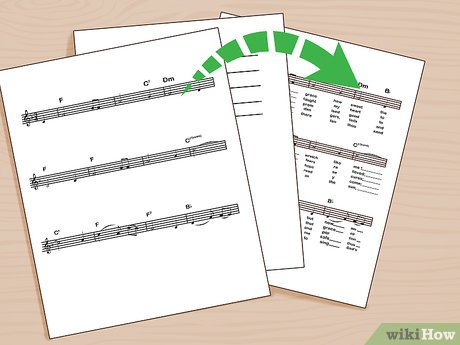}
\end{minipage}
\hfill
\begin{minipage}[t]{0.15\linewidth}
\centering
Image 1\\[2pt]
\includegraphics[width=\linewidth]{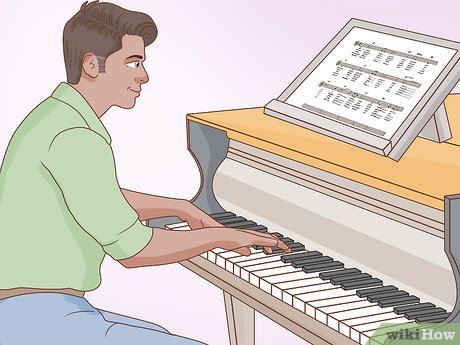}
\end{minipage}
\hfill
\begin{minipage}[t]{0.15\linewidth}
\centering
Image 2\\[2pt]
\includegraphics[width=\linewidth]{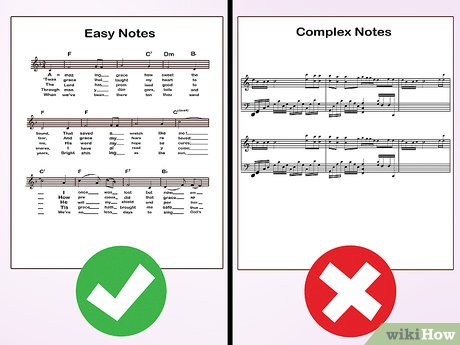}
\end{minipage}
\hfill
\begin{minipage}[t]{0.15\linewidth}
\centering
Image 3\\[2pt]
\includegraphics[width=\linewidth]{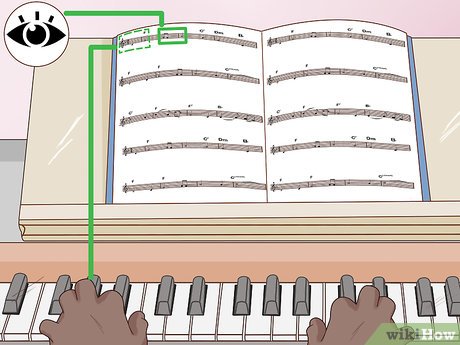}
\end{minipage}
\hfill
\begin{minipage}[t]{0.15\linewidth}
\centering
Image 4\\[2pt]
\includegraphics[width=\linewidth]{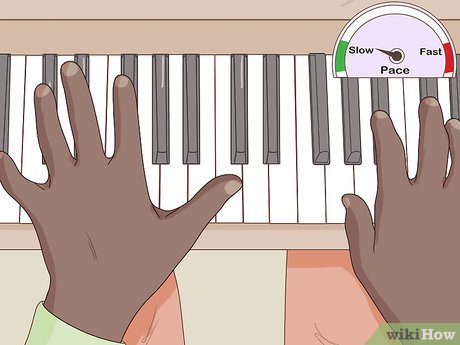}
\end{minipage}
\hfill
\begin{minipage}[t]{0.15\linewidth}
\centering
Image 5\\[2pt]
\includegraphics[width=\linewidth]{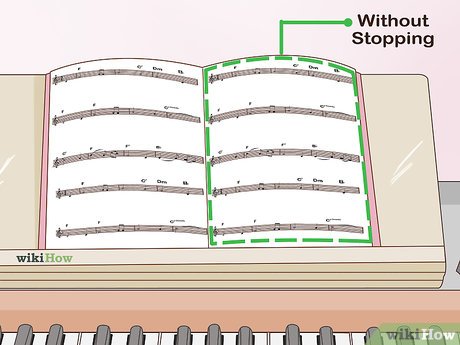}
\end{minipage}
\end{center}

\#\# Answer: [4, 3, 5, 1, 2, 0]
\end{observationbox}

\begin{observationbox}{WikiHow Case 2}
\#\# Article Text (with placeholders):

Begin with clean sheet metal. Use soap and water to clean off any debris or dust on the sheet metal. Rinse the item with clean water and dry it with a soft cloth.

\vspace{-0.5em}
\begin{center}
[IMAGE\_PLACEHOLDER]
\end{center}
\vspace{-0.5em}
Apply protective eye-wear and a mask. You should always protect your eyes and the rest of your face from the machines you're using. These measures are also necessary to keep dust and polish out of your eyes, nose, and mouth.

\vspace{-0.5em}
\begin{center}
[IMAGE\_PLACEHOLDER]
\end{center}
\vspace{-0.5em}

Sand the sheet metal. In order to get a mirror finish on your car, boat, or aluminum panels, you're going to need to do some sanding. Begin sanding with a medium grit sandpaper and work your way up to finer sandpaper. While it's possible to sand the aluminum by hand, using a sanding machine will make your job incredibly easier.  For a quick polish, start with 400-grit sandpaper and give the entire area a good sanding. Then use an 800-grit sandpaper and give the area another good once-over.   For a thorough polish, start with 120 grit and move up to 240, 320, 400, and finally, 600-grit.

\vspace{-0.5em}
\begin{center}
[IMAGE\_PLACEHOLDER]
\end{center}
\vspace{-0.5em}

Apply a cutting compound to your buffing tool. Before buffing, apply a cutting compound to the buffing tool. The cutting compound protects the metal and gives it a nice shine. Read the instructions on the package to determine which compound to use for your project.  In general, you can start with a firm wheel and a brown compound to do your initial polishing, then choose a softer wheel and a rouge (red) compound to get a high shine and smooth finish.

\vspace{-0.5em}
\begin{center}
[IMAGE\_PLACEHOLDER]
\end{center}
\vspace{-0.5em}

Use a rotating buffing tool to buff the aluminum. A cotton buffing tool works well for aluminum. Use circular motions to buff the sheet metal. Follow the instructions in the manual and exercise caution when using the buffing tool.

\vspace{-0.5em}
\begin{center}
[IMAGE\_PLACEHOLDER]
\end{center}
\vspace{-0.5em}

Wipe off any traces of compound. Use a soft, clean cloth to remove the compound residue from the aluminum. Wipe until a mirror finish is achieved.

\vspace{-0.5em}
\begin{center}
[IMAGE\_PLACEHOLDER]
\end{center}
\vspace{-0.5em}

\#\# Candidate Images (Image 0 to Image 5):

\begin{center}
% 第一排
\begin{minipage}[t]{0.3\linewidth}
\centering
Image 0\\[2pt]
\includegraphics[width=\linewidth]{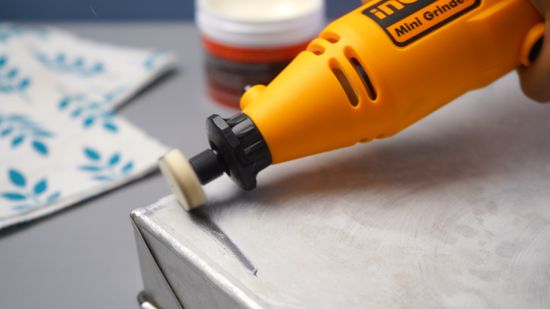}
\end{minipage}
\hfill
\begin{minipage}[t]{0.3\linewidth}
\centering
Image 1\\[2pt]
\includegraphics[width=\linewidth]{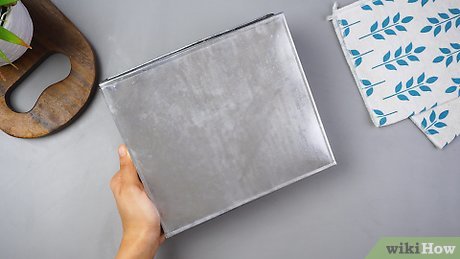}
\end{minipage}
\hfill
\begin{minipage}[t]{0.3\linewidth}
\centering
Image 2\\[2pt]
\includegraphics[width=\linewidth]{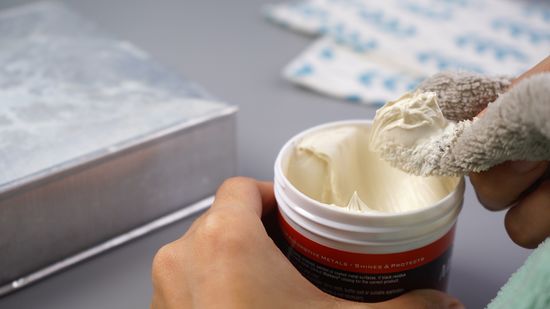}
\end{minipage}

\vspace{0.5em}

% 第二排
\begin{minipage}[t]{0.3\linewidth}
\centering
Image 3\\[2pt]
\includegraphics[width=\linewidth]{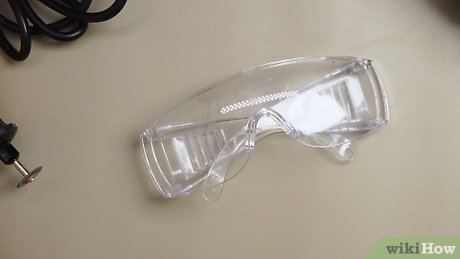}
\end{minipage}
\hfill
\begin{minipage}[t]{0.3\linewidth}
\centering
Image 4\\[2pt]
\includegraphics[width=\linewidth]{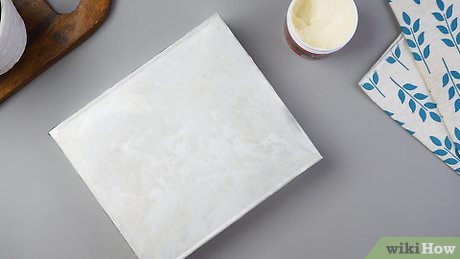}
\end{minipage}
\hfill
\begin{minipage}[t]{0.3\linewidth}
\centering
Image 5\\[2pt]
\includegraphics[width=\linewidth]{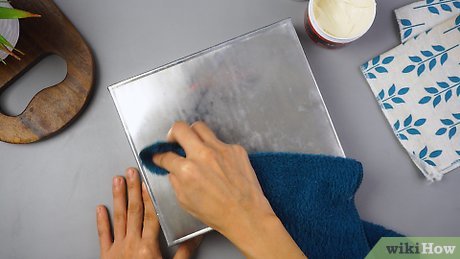}
\end{minipage}
\end{center}

\#\# Answer: [4, 3, 1, 2, 0, 5]
\end{observationbox}

\begin{observationbox}{WikiHow Case 3}
\#\# Article Text (with placeholders):

Cover your rafter trusses with plywood sheathing. Put down your first sheet of plywood at the corner of one end of the roof. Make sure it’s lying horizontally across the exposed rafters, and that the edges are flush with the edges of the end rafters. Drive a nail into each corner of the plywood to hold it in place temporarily.   Most constructions experts recommend using ⁄ 16 in (1.1 cm) oriented strand board (OSB) for small-scale roofing projects.   Plywood sheathing will provide structural support for your new roof, as well as give you a flat, stable surface to attach your other roofing materials too.

\vspace{-0.5em}
\begin{center}
[IMAGE\_PLACEHOLDER]
\end{center}
\vspace{-0.5em}

to fill in any gaps in the sheathing. Plywood is sold in large sheets, which means that you’ll most likely need to use multiple sheets and cut them to fit. Try to cover the remaining space using as few pieces as possible, starting from the lower portion of the roof.   It’s important to cut your plywood so that the end of each section covers half the width of the rafter it’s resting on. That way, the neighboring section will fit in easily beside it, and you’ll have a nice solid surface to nail into.   Make all your cuts with your plywood oriented the same way to ensure that the strand grain is running in a single direction. A consistent grain pattern will increase the strength of your roof sheathing.

\vspace{-0.5em}
\begin{center}
[IMAGE\_PLACEHOLDER]
\end{center}
\vspace{-0.5em}

Fasten your plywood sheathing to the rafters using 8D finishing nails. Drive nails every 6 inches (15 cm) through the face of the plywood and down into the rafter below. Work your way up the length of each rafter from the bottom edge. When you’re done, look for any loose sections of plywood that may require additional nails.   For the sake of caution, do your fastening from your ladder, reaching as far as you can safely with your hammer or roofing nailer.  The combined strength of the OSB and supporting rafters will be capable of withstanding weights of up to several hundred pounds.

\vspace{-0.5em}
\begin{center}
[IMAGE\_PLACEHOLDER]
\end{center}
\vspace{-0.5em}

Cut fascia boards to finish the edges of your roof. Once you have the sheathing in place, your final task will be to mount fascia boards to cover the exposed ends of your rafters. Cut your 2 in (5.1 cm) x 4 in (10 cm) or 2 in (5.1 cm) x 6 in (15 cm) to match the length of the shed. Fasten the fascia boards by nailing them to the end face of every other rafter using 8D finishing nails.   You’ll need to put up 2 fascia boards for gable, gambrel, skillion, and saltbox and other slanted roof styles—1 for each sloped edge. For flat roofs, it will look best to install a fascia board on every side.  When cutting your fascia boards, be sure to use the same size lumber as you did for your rafters to guarantee an exact fit.
\vspace{-0.5em}
\begin{center}
[IMAGE\_PLACEHOLDER]
\end{center}
\vspace{-0.5em}

\#\# Candidate Images (Image 0 to Image 3):

\begin{center}
\begin{minipage}[t]{0.23\linewidth}
\centering
Image 0\\[2pt]
\includegraphics[width=\linewidth]{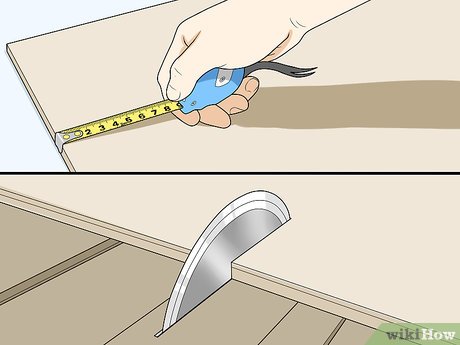}
\end{minipage}
\hfill
\begin{minipage}[t]{0.23\linewidth}
\centering
Image 1\\[2pt]
\includegraphics[width=\linewidth]{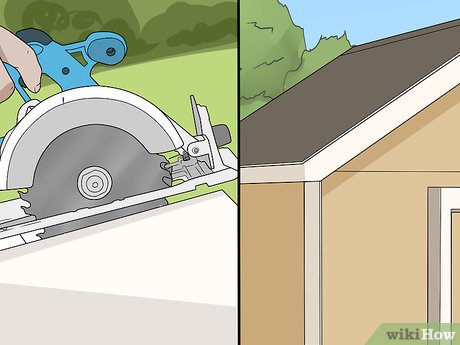}
\end{minipage}
\hfill
\begin{minipage}[t]{0.23\linewidth}
\centering
Image 2\\[2pt]
\includegraphics[width=\linewidth]{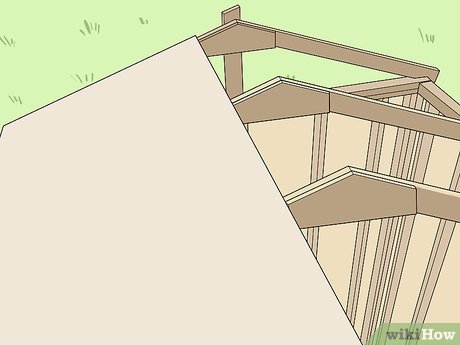}
\end{minipage}
\hfill
\begin{minipage}[t]{0.23\linewidth}
\centering
Image 3\\[2pt]
\includegraphics[width=\linewidth]{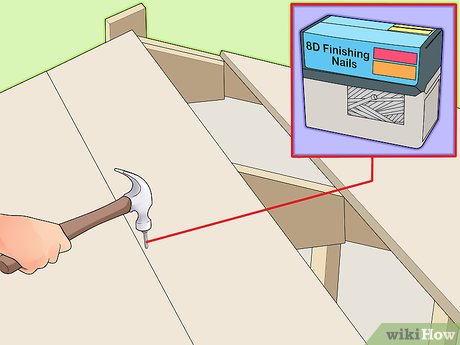}
\end{minipage}
\hfill
\end{center}

\#\# Answer: [2, 0, 3, 1]
\end{observationbox}

\begin{observationbox}{WikiHow Case 4}
\#\# Article Text (with placeholders):

Right-click the file you want to open. You can change the default media player for any audio or video file on your Mac. Use your mouse or touchpad to move your cursor on the file, and right-click on it to expand a drop-down menu of options.

\vspace{-0.5em}
\begin{center}
[IMAGE\_PLACEHOLDER]
\end{center}
\vspace{-0.5em}

Click Get Info . This option should be at the top of the third section on the right-click menu. It will open a new window with the file and format details of this video.

\vspace{-0.5em}
\begin{center}
[IMAGE\_PLACEHOLDER]
\end{center}
\vspace{-0.5em}

Click the arrow next to Open with (to expand the menu). The default media player for this file type will display. If the "Open with" panel is already open, you can skip this step to expand the menu since clicking this arrow will only close the panel if it's open.

[\vspace{-0.5em}
\begin{center}
[IMAGE\_PLACEHOLDER]
\end{center}
\vspace{-0.5em}

Click the currently listed default player. This will prompt a list of apps compatible with your file type to drop-down.   Click Other if you don't see your favorite media player displayed in the list.  Alternatively, click App Store at the bottom of the menu to see a list of available software for download. It will open the Mac App Store, and list all media players that will play, edit, or convert this file format.

\vspace{-0.5em}
\begin{center}
[IMAGE\_PLACEHOLDER]
\end{center}
\vspace{-0.5em}

Select a media player from the list. Click on the media software you want to set as your new default player for this file format.

\vspace{-0.5em}
\begin{center}
[IMAGE\_PLACEHOLDER]
\end{center}
\vspace{-0.5em}

Click Change All . This will change your default media player selection for all files with the same file format extension. You will have to confirm your action in a pop-up box by clicking Continue . Changing the default player for an audio or video format will not apply your changes to all file formats. For example, if you change the default video player for MOV files, you will still have to change it for AVI files manually.

\vspace{-0.5em}
\begin{center}
[IMAGE\_PLACEHOLDER]
\end{center}
\vspace{-0.5em}

\#\# Candidate Images (Image 0 to Image 5):

\begin{center}
% 第一排
\begin{minipage}[t]{0.3\linewidth}
\centering
Image 0\\[2pt]
\includegraphics[width=\linewidth]{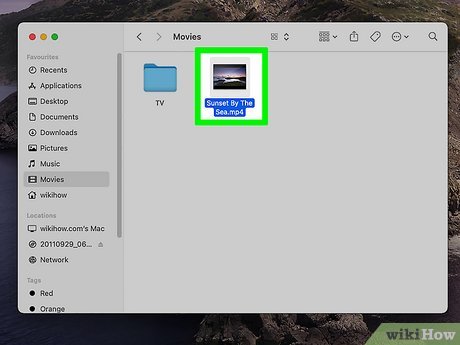}
\end{minipage}
\hfill
\begin{minipage}[t]{0.3\linewidth}
\centering
Image 1\\[2pt]
\includegraphics[width=\linewidth]{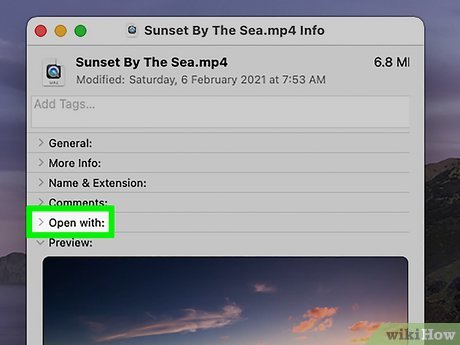}
\end{minipage}
\hfill
\begin{minipage}[t]{0.3\linewidth}
\centering
Image 2\\[2pt]
\includegraphics[width=\linewidth]{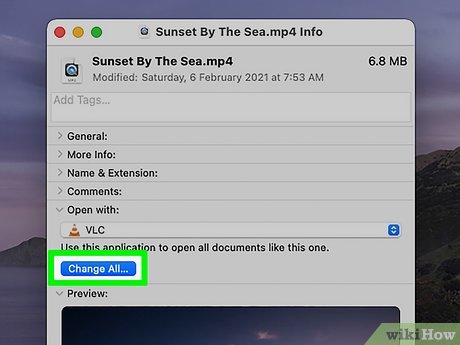}
\end{minipage}

\vspace{0.5em}

% 第二排
\begin{minipage}[t]{0.3\linewidth}
\centering
Image 3\\[2pt]
\includegraphics[width=\linewidth]{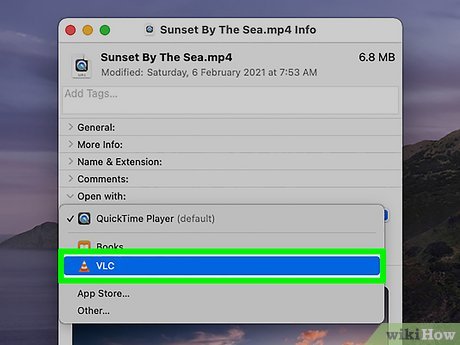}
\end{minipage}
\hfill
\begin{minipage}[t]{0.3\linewidth}
\centering
Image 4\\[2pt]
\includegraphics[width=\linewidth]{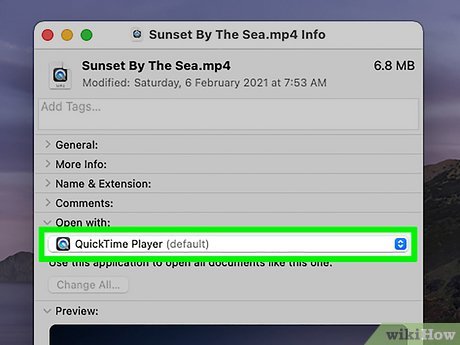}
\end{minipage}
\hfill
\begin{minipage}[t]{0.3\linewidth}
\centering
Image 5\\[2pt]
\includegraphics[width=\linewidth]{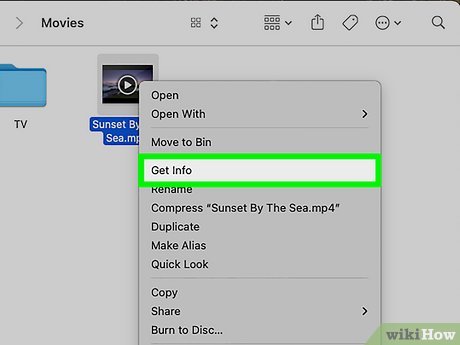}
\end{minipage}
\end{center}

\#\# Answer: [0, 5, 1, 4, 3, 2],
\end{observationbox}

\begin{observationbox}{WikiHow Case 5}
\#\# Article Text (with placeholders):

Collect your materials. Painting your face to look like a skull is quite easy, but you will need a few special paint brushes and face paints to accomplish this look. Before you get started, you will need:   face paints in white, grey, and black  a small paintbrush for outlining and detailing  a thick paintbrush for filling in large areas  a fine paintbrush for precise details  a cup of water for thinning out face paint

\vspace{-0.5em}
\begin{center}
[IMAGE\_PLACEHOLDER]
\end{center}
\vspace{-0.5em}

Outline your eyes. Use your small paintbrush and white paint to paint white lines all around your eye sockets. This is the area just below your eyebrows and about an inch or two below your lower eyelid. Try to trace along the bones around your eyes to create this outline.  If the face paint seems too thick, then you can just dip your brush into the water before dipping it into the paint.

\vspace{-0.5em}
\begin{center}
[IMAGE\_PLACEHOLDER]
\end{center}
\vspace{-0.5em}

Create the mask outline around your forehead and cheeks. Next, use the small paintbrush and white face paint to create an outline that extends over your forehead and around your cheekbones as well. The outline should then curve down towards your lips and end at the corner of your lips.  Your outline should resemble the shape of a skull when you are done.

\vspace{-0.5em}
\begin{center}
[IMAGE\_PLACEHOLDER]
\end{center}
\vspace{-0.5em}

Draw a triangle above each nostril. Use the white face paint and your thin paintbrush to create two triangles right above each of your nostrils. There should be one triangle above each nostril when you are done.

\vspace{-0.5em}
\begin{center}
[IMAGE\_PLACEHOLDER]
\end{center}
\vspace{-0.5em}

Fill in the outline. Next, switch to a larger brush to fill in the large area that you just outlined with white face paint. Do not fill in the triangles above your nostrils or the areas around your eyes. Leave these areas unpainted for now.  Allow this first coat of face paint to dry completely before you start shading or adding details.

\vspace{-0.5em}
\begin{center}
[IMAGE\_PLACEHOLDER]
\end{center}
\vspace{-0.5em}

\#\# Candidate Images (Image 0 to Image 5):

\begin{center}
\begin{minipage}[t]{0.18\linewidth}
\centering
Image 0\\[2pt]
\includegraphics[width=\linewidth]{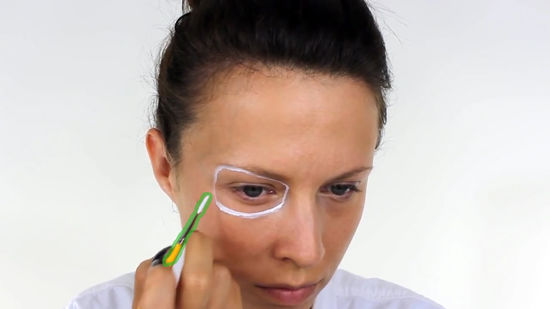}
\end{minipage}
\hfill
\begin{minipage}[t]{0.18\linewidth}
\centering
Image 1\\[2pt]
\includegraphics[width=\linewidth]{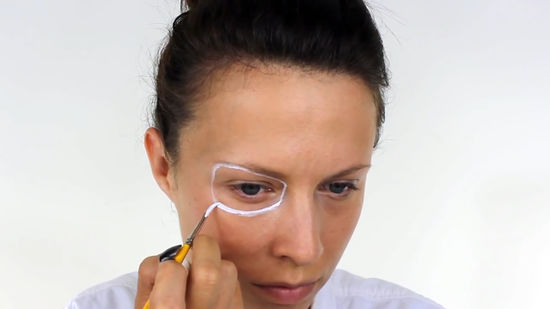}
\end{minipage}
\hfill
\begin{minipage}[t]{0.18\linewidth}
\centering
Image 2\\[2pt]
\includegraphics[width=\linewidth]{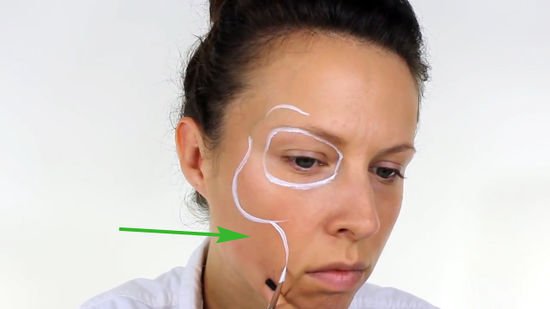}
\end{minipage}
\hfill
\begin{minipage}[t]{0.18\linewidth}
\centering
Image 3\\[2pt]
\includegraphics[width=\linewidth]{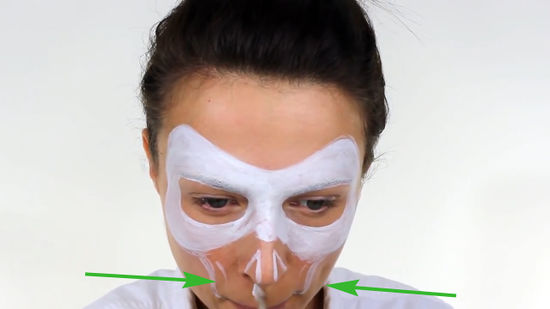}
\end{minipage}
\hfill
\begin{minipage}[t]{0.18\linewidth}
\centering
Image 4\\[2pt]
\includegraphics[width=\linewidth]{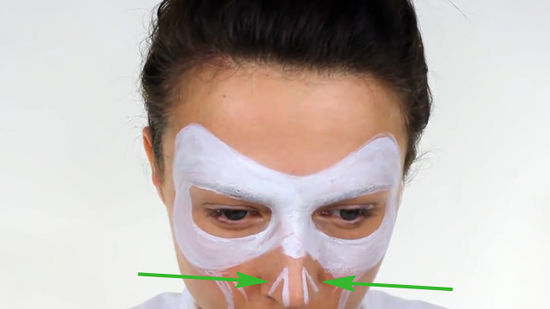}
\end{minipage}
\end{center}

\#\# Answer: [0, 1, 2, 4, 3],
\end{observationbox}

\begin{observationbox}{StoryBird Case 1}
\#\# Article Text (with placeholders):

And this is my story...

\vspace{-0.5em}
\begin{center}
[IMAGE\_PLACEHOLDER]
\end{center}
\vspace{-0.5em}

I was a strange kid but I lived a happy life. I grew up in a foster home with no real family, and that was all I wanted.

\vspace{-0.5em}
\begin{center}
[IMAGE\_PLACEHOLDER]
\end{center}
\vspace{-0.5em}

I would always play by myself. No one there liked me I felt like I didn't belong. I knew I was different, I felt like I was meant to do something great.

\vspace{-0.5em}
\begin{center}
[IMAGE\_PLACEHOLDER]
\end{center}
\vspace{-0.5em}

I made a few friends but they all left me when I would play outside and get messy. But when I was outside I felt like I was home like something was waiting for me.

\vspace{-0.5em}
\begin{center}
[IMAGE\_PLACEHOLDER]
\end{center}
\vspace{-0.5em}

I loved playing around in the pool, I once bathed in the River. I got in BIG trouble but I felt clean when I did it in the river and dirty when in the bath

\vspace{-0.5em}
\begin{center}
[IMAGE\_PLACEHOLDER]
\end{center}
\vspace{-0.5em}

I was right about something waiting for me. My real family was and I found them I am 11 now I found them when I was 9. I love my people and they love me! This is where I was meant to be in the wild...in my true home.

\vspace{-0.5em}
\begin{center}
[IMAGE\_PLACEHOLDER]
\end{center}
\vspace{-0.5em}

\#\# Candidate Images (Image 0 to Image 5):

\begin{center}
\begin{minipage}[t]{0.23\linewidth}
\centering
Image 0\\[2pt]
\includegraphics[width=\linewidth]{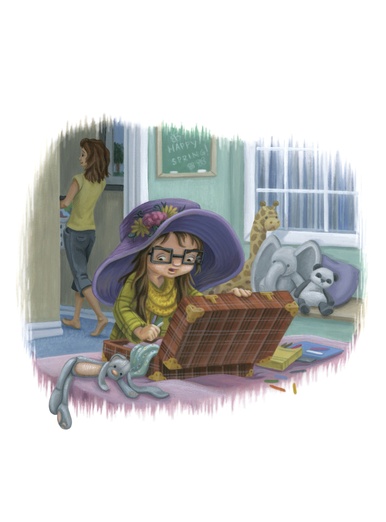}
\end{minipage}
\hfill
\begin{minipage}[t]{0.23\linewidth}
\centering
Image 1\\[2pt]
\includegraphics[width=\linewidth]{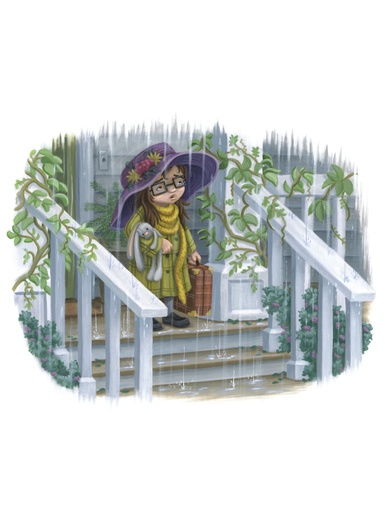}
\end{minipage}
\hfill
\begin{minipage}[t]{0.23\linewidth}
\centering
Image 2\\[2pt]
\includegraphics[width=\linewidth]{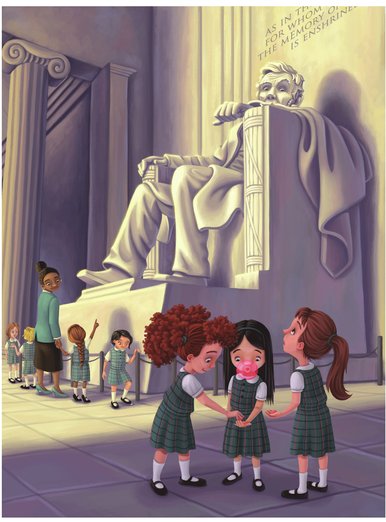}
\end{minipage}

\vspace{0.5em}

\begin{minipage}[t]{0.23\linewidth}
\centering
Image 3\\[2pt]
\includegraphics[width=\linewidth]{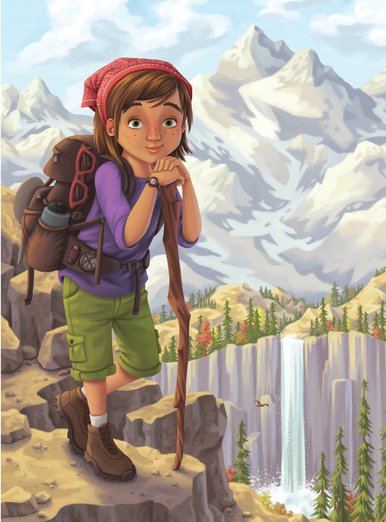}
\end{minipage}
\hfill
\begin{minipage}[t]{0.23\linewidth}
\centering
Image 4\\[2pt]
\includegraphics[width=\linewidth]{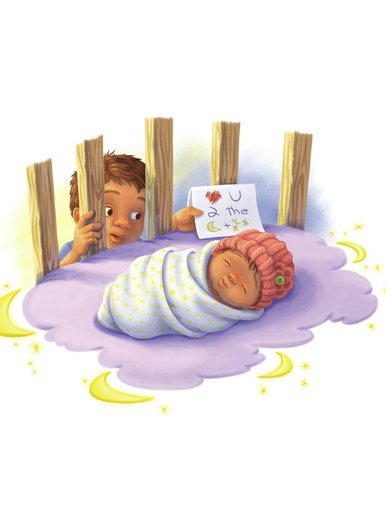}
\end{minipage}
\hfill
\begin{minipage}[t]{0.23\linewidth}
\centering
Image 5\\[2pt]
\includegraphics[width=\linewidth]{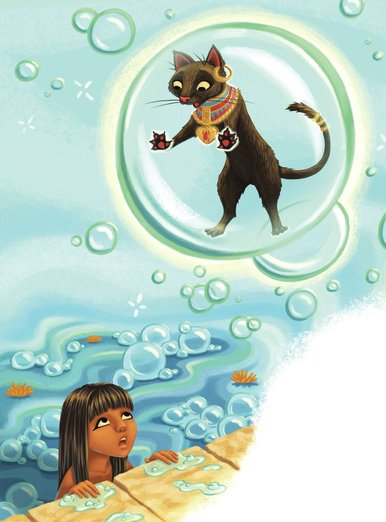}
\end{minipage}
\end{center}

\#\# Answer: [4, 1, 0, 2, 5, 3]
\end{observationbox}

\begin{observationbox}{StoryBird Case 2}
\#\# Article Text (with placeholders):

“Forgive a soul that breathes on your existence. Speak of no forlorn, for I am forever here for you darling,”he saith. She picks the pearls that runs from her eyes with a handkerchief and rushes looking him in the eyes, asking: “Do you love me Charles.” He heaves and tight he hugs her again with his sleeves and responds: “What would a swan look like without a lake? an empty space, a shallow existence, a morbid being

\vspace{-0.5em}
\begin{center}
[IMAGE\_PLACEHOLDER]
\end{center}
\vspace{-0.5em}

We wandered cities under peaceful rain. Time had no definition. We only felt our skins touching. We only heard our laughter and words. We only seen the faces of each other. For us that is what love has meant to be .. And so we believed in it.

\vspace{-0.5em}
\begin{center}
[IMAGE\_PLACEHOLDER]
\end{center}
\vspace{-0.5em}

I see him in all faces “ My guardian angel “..

\vspace{-0.5em}
\begin{center}
[IMAGE\_PLACEHOLDER]
\end{center}
\vspace{-0.5em}

He finds me in all spaces “his guardian Angel” ..

\vspace{-0.5em}
\begin{center}
[IMAGE\_PLACEHOLDER]
\end{center}
\vspace{-0.5em}

\#\# Candidate Images (Image 0 to Image 3):

\begin{center}
\begin{minipage}[t]{0.22\linewidth}
\centering
Image 0\\[2pt]
\includegraphics[width=\linewidth]{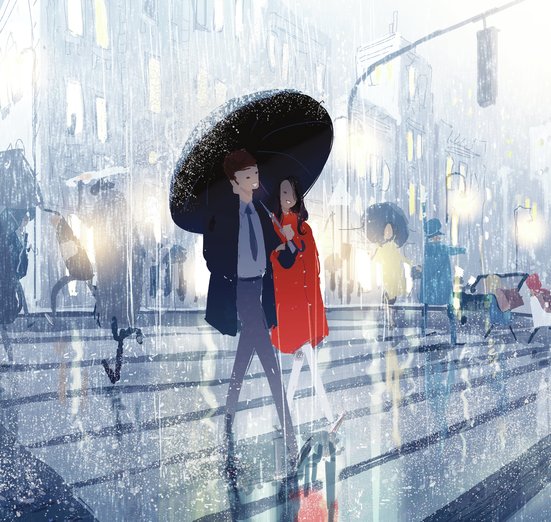}
\end{minipage}
\hfill
\begin{minipage}[t]{0.22\linewidth}
\centering
Image 1\\[2pt]
\includegraphics[width=\linewidth]{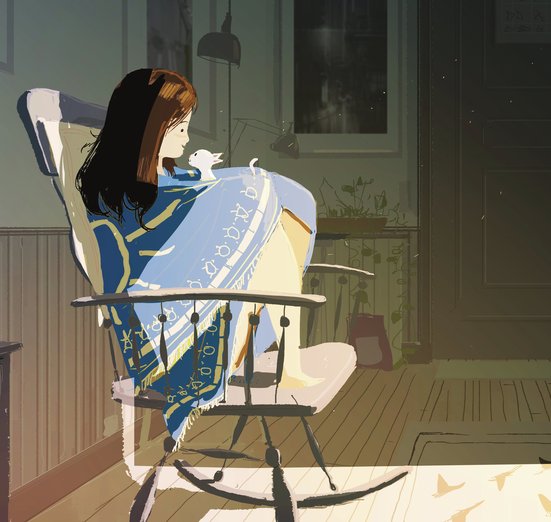}
\end{minipage}
\hfill
\begin{minipage}[t]{0.22\linewidth}
\centering
Image 2\\[2pt]
\includegraphics[width=\linewidth]{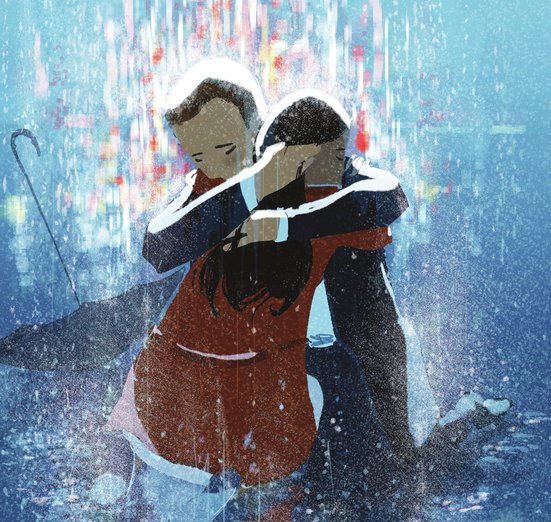}
\end{minipage}
\hfill
\begin{minipage}[t]{0.22\linewidth}
\centering
Image 3\\[2pt]
\includegraphics[width=\linewidth]{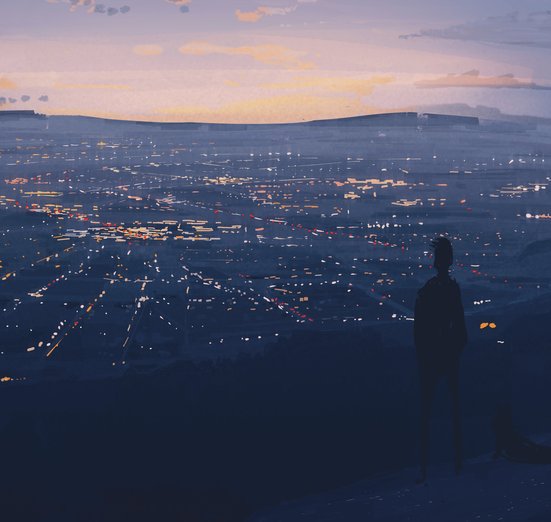}
\end{minipage}
\end{center}

\#\# Answer: [2, 0, 1, 3]
\end{observationbox}

\begin{observationbox}{StoryBird Case 3}
\#\# Article Text (with placeholders):

These will be good decorations. We can use some different colored paper. Now, have a seat so we can start!

\vspace{-0.5em}
\begin{center}
[IMAGE\_PLACEHOLDER]
\end{center}
\vspace{-0.5em}

Can I borrow the scissors? There is another pair in the left drawer.

\vspace{-0.5em}
\begin{center}
[IMAGE\_PLACEHOLDER]
\end{center}
\vspace{-0.5em}

I am making a star.What are you making Bird? Cool Blue! I love stars. I am making a flower inside of a circle for my little sister.

\vspace{-0.5em}
\begin{center}
[IMAGE\_PLACEHOLDER]
\end{center}
\vspace{-0.5em}

That is cool Bird! I am giving this to my little sister too! Rose will really like it! Look Blue! I am finished. My little sister Lilly is going to be so happy. I wonder when my mom is picking me up so I can give the flower to Lilly!

\vspace{-0.5em}
\begin{center}
[IMAGE\_PLACEHOLDER]
\end{center}
\vspace{-0.5em}

\#\# Candidate Images (Image 0 to Image 3):

\begin{center}
\begin{minipage}[t]{0.23\linewidth}
\centering
Image 0\\[2pt]
\includegraphics[width=\linewidth]{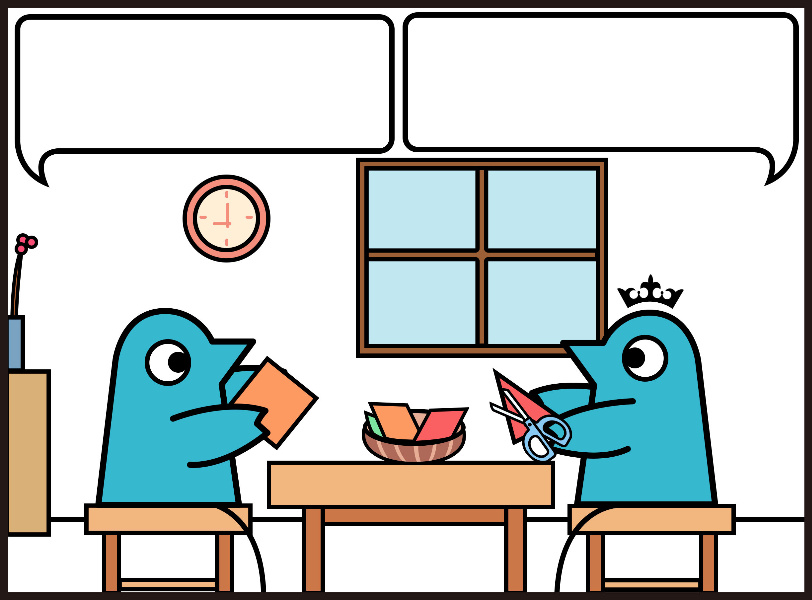}
\end{minipage}
\hfill
\begin{minipage}[t]{0.23\linewidth}
\centering
Image 1\\[2pt]
\includegraphics[width=\linewidth]{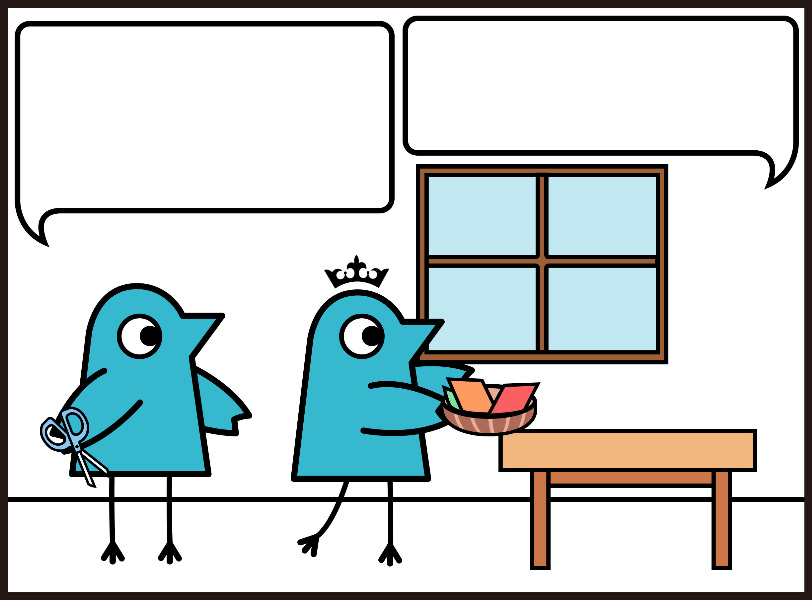}
\end{minipage}
\hfill
\begin{minipage}[t]{0.23\linewidth}
\centering
Image 2\\[2pt]
\includegraphics[width=\linewidth]{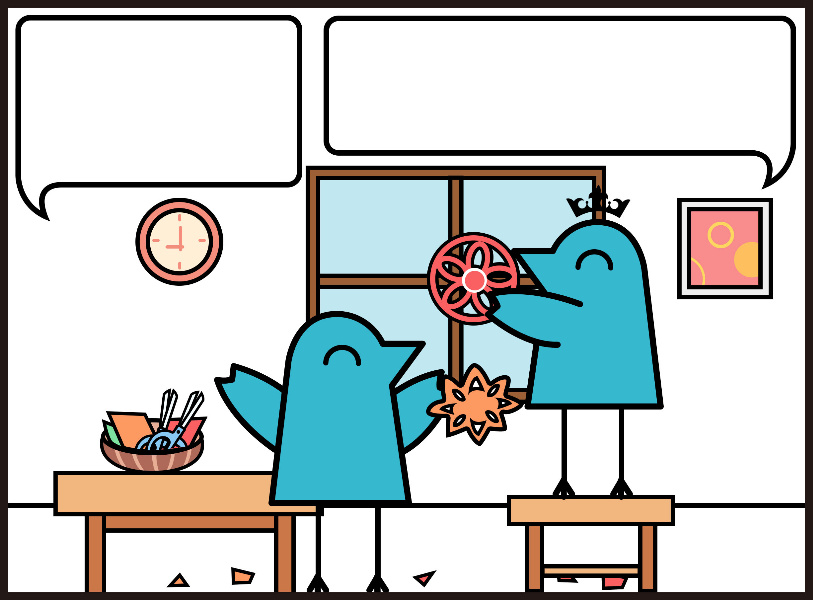}
\end{minipage}
\hfill
\begin{minipage}[t]{0.23\linewidth}
\centering
Image 3\\[2pt]
\includegraphics[width=\linewidth]{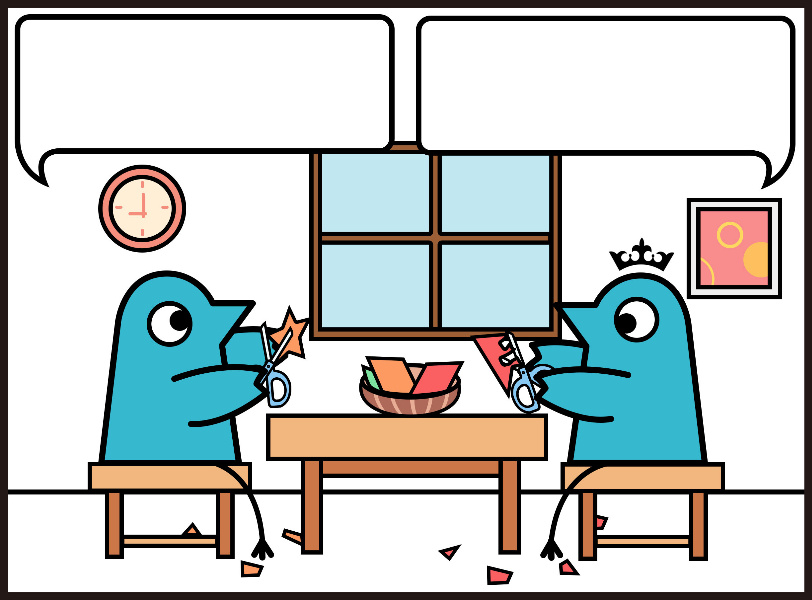}
\end{minipage}
\end{center}

\#\# Answer: [1, 0, 3, 2]
\end{observationbox}

\begin{observationbox}{StoryBird Case 4}
\#\# Article Text (with placeholders):

I am a super hero! I.AM.YOUR. ROYAL. SERVANT.

\vspace{-0.5em}
\begin{center}
[IMAGE\_PLACEHOLDER]
\end{center}
\vspace{-0.5em}

I’m bored of playing super heroes. Me too.

\vspace{-0.5em}
\begin{center}
[IMAGE\_PLACEHOLDER]
\end{center}
\vspace{-0.5em}

I will defeat all the bad guys in the world! Yes!!! Just saying... I am a secret bad guy in disguise.. do not capture me because my watch is very expensive. I will take off my watch and then you can defeat me.

\vspace{-0.5em}
\begin{center}
[IMAGE\_PLACEHOLDER]
\end{center}
\vspace{-0.5em}

WHOEVER IS WRITING THIS BOOK STOP MY LEGS FROM WIGGLING, BECAUSE I HAVE A VERY EXPENSIVE WATCH ON AND IF YOU BREAK IT I WILL BE VERY ANGRY.

\vspace{-0.5em}
\begin{center}
[IMAGE\_PLACEHOLDER]
\end{center}
\vspace{-0.5em}

\#\# Candidate Images (Image 0 to Image 3):

\begin{center}
\begin{minipage}[t]{0.23\linewidth}
\centering
Image 0\\[2pt]
\includegraphics[width=\linewidth]{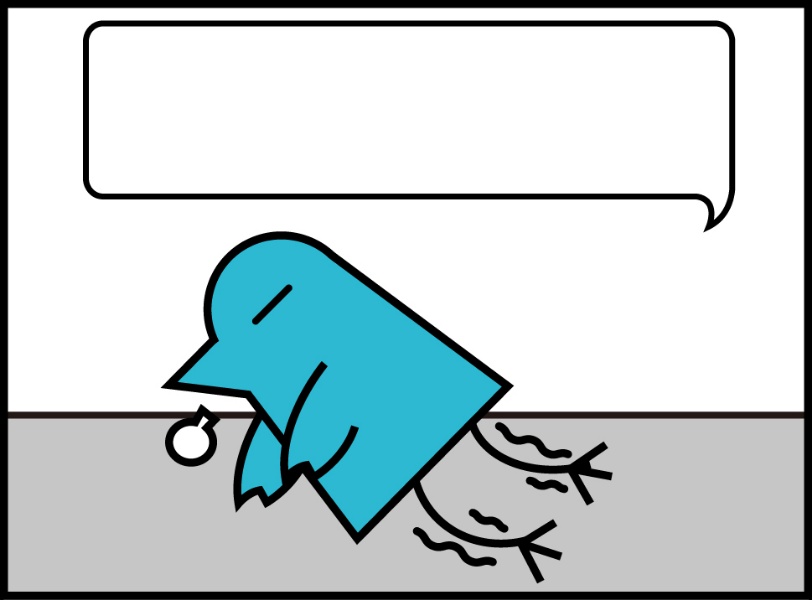}
\end{minipage}
\hfill
\begin{minipage}[t]{0.23\linewidth}
\centering
Image 1\\[2pt]
\includegraphics[width=\linewidth]{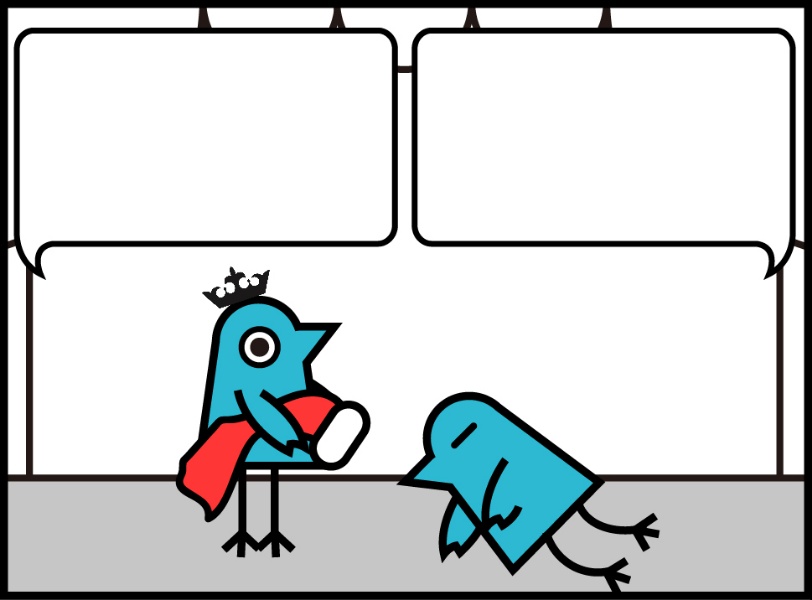}
\end{minipage}
\hfill
\begin{minipage}[t]{0.23\linewidth}
\centering
Image 2\\[2pt]
\includegraphics[width=\linewidth]{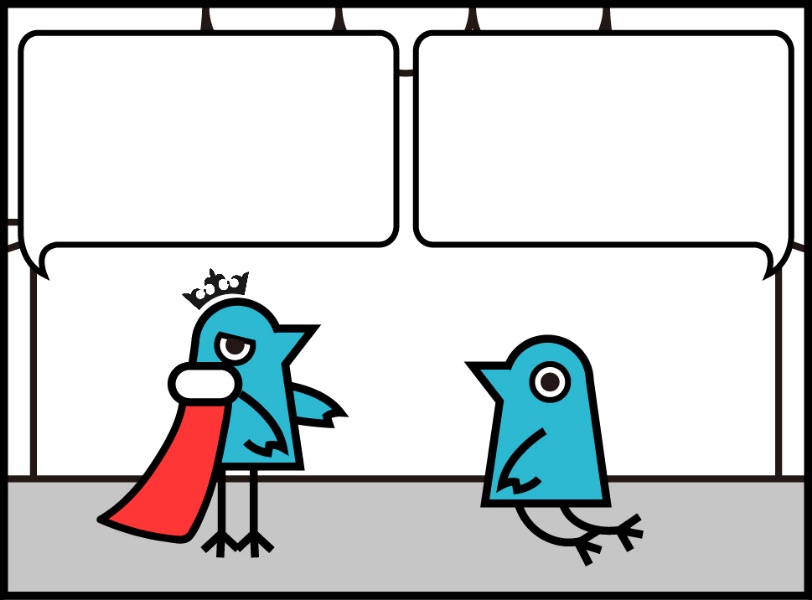}
\end{minipage}
\hfill
\begin{minipage}[t]{0.23\linewidth}
\centering
Image 3\\[2pt]
\includegraphics[width=\linewidth]{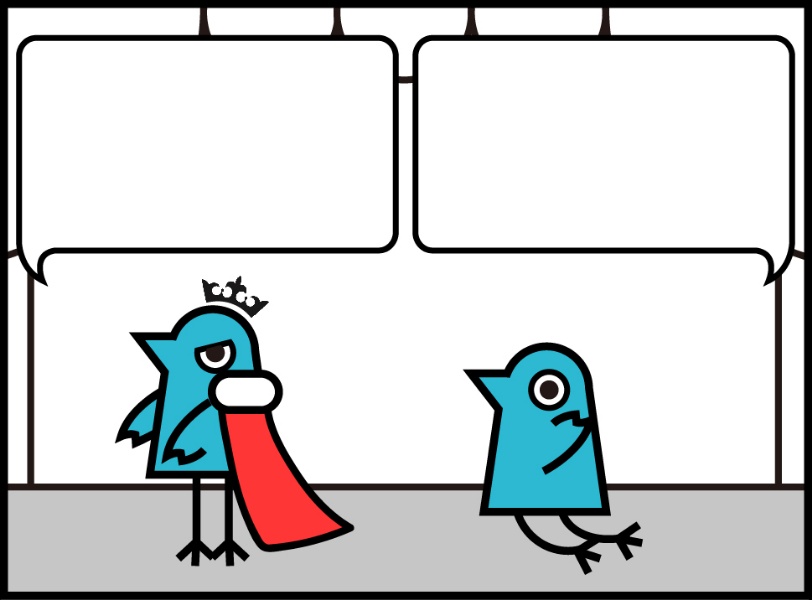}
\end{minipage}
\end{center}

\#\# Answer: [2, 1, 3, 0]
\end{observationbox}

\begin{observationbox}{StoryBird Case 5}
\#\# Article Text (with placeholders):

Every Sunday Grandma Ann´s children went to her job for lunch

\vspace{-0.5em}
\begin{center}
[IMAGE\_PLACEHOLDER]
\end{center}
\vspace{-0.5em}

Grandmother Ann told stories to her grandchildren about her studies in plants, but no one paid attention, only the little Sofía while the others played on their cell phones

\vspace{-0.5em}
\begin{center}
[IMAGE\_PLACEHOLDER]
\end{center}
\vspace{-0.5em}

One day the children had gone out to play in the park and suddenly Tommy started to cry very hard and his stomach hurt a lot, apparenly he had eaten something that looked like blackberries on a tree, nobody knew what to do

\vspace{-0.5em}
\begin{center}
[IMAGE\_PLACEHOLDER]
\end{center}
\vspace{-0.5em}

Sofia had run to the tree for some leaves for Tommy to eat, at that moment Tommy stopped feeling pain

\vspace{-0.5em}
\begin{center}
[IMAGE\_PLACEHOLDER]
\end{center}
\vspace{-0.5em}

Sofia knew what to do because she had heard her grandmother talk about a fruit similar to blackberries but it was poisonous and how to eliminate the poison from the body

\vspace{-0.5em}
\begin{center}
[IMAGE\_PLACEHOLDER]
\end{center}
\vspace{-0.5em}

Everyone congradulated Sofia and they had learned the importance of listening to the elderly.

\vspace{-0.5em}
\begin{center}
[IMAGE\_PLACEHOLDER]
\end{center}
\vspace{-0.5em}

\#\# Candidate Images (Image 0 to Image 5):

\begin{center}
\begin{minipage}[t]{0.25\linewidth}
\centering
Image 0\\[2pt]
\includegraphics[width=\linewidth]{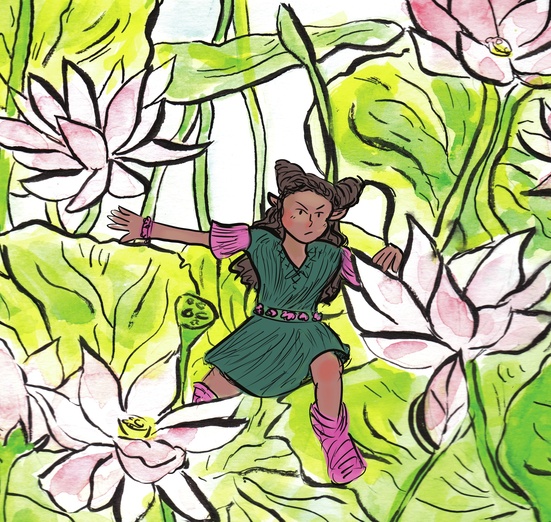}
\end{minipage}
\hfill
\begin{minipage}[t]{0.4\linewidth}
\centering
Image 1\\[2pt]
\includegraphics[width=\linewidth]{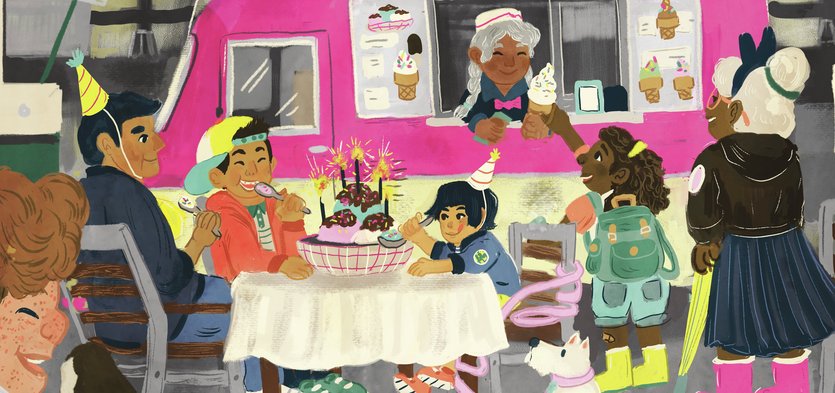}
\end{minipage}
\hfill
\begin{minipage}[t]{0.25\linewidth}
\centering
Image 2\\[2pt]
\includegraphics[width=\linewidth]{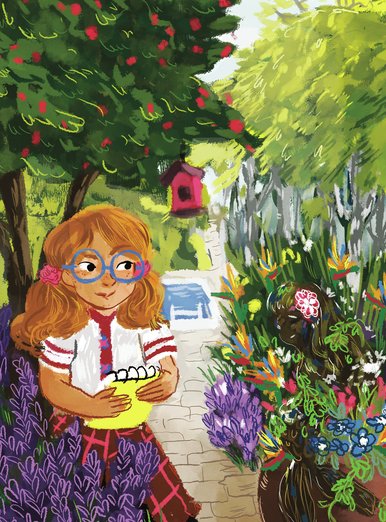}
\end{minipage}

\vspace{0.5em}

\begin{minipage}[t]{0.25\linewidth}
\centering
Image 3\\[2pt]
\includegraphics[width=\linewidth]{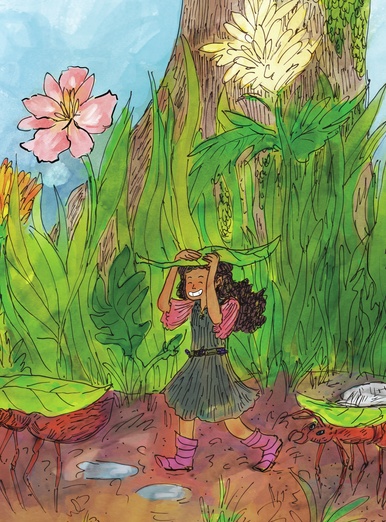}
\end{minipage}
\hfill
\begin{minipage}[t]{0.3\linewidth}
\centering
Image 4\\[2pt]
\includegraphics[width=\linewidth]{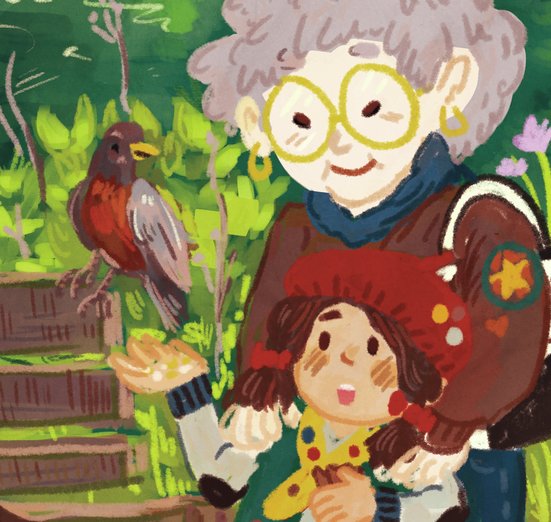}
\end{minipage}
\hfill
\begin{minipage}[t]{0.3\linewidth}
\centering
Image 5\\[2pt]
\includegraphics[width=\linewidth]{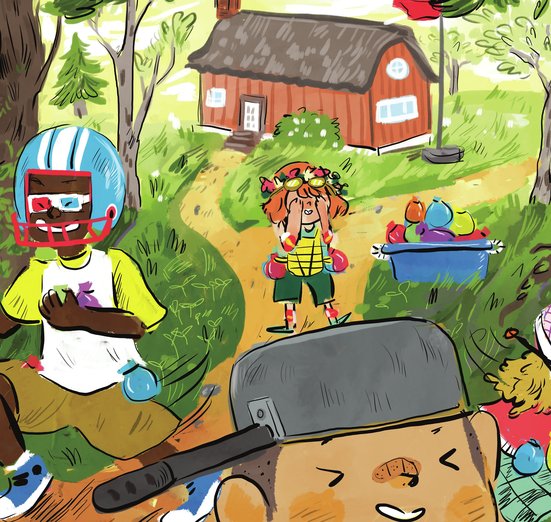}
\end{minipage}
\end{center}

\#\# Answer: [1, 2, 5, 3, 0, 4]
\end{observationbox}

\begin{observationbox}{Science Case 1}
\#\# Article Text (with placeholders):

Water Powered Flashlight - Micro MEDELIS BATTERYJust Add Water and This Tiny Emergency Flashlight Glows ... I named this MEDELIS BATTERY This battery generates electric power out of Tap Water (drinking water, river,lake,rain or any similar liquid)... its small, powerful and super basic MEDELIS BATTERY will generates electricity until the water dries up...for many hours (you can always refill water - make it work again)

\vspace{-0.5em}
\begin{center}
[IMAGE\_PLACEHOLDER]
\end{center}
\vspace{-0.5em}

To Build Water Powered Flashlight (MEDELIS BATTERY) You Will Need Some StuffCopper Plate, Magnesium Ribbon \& Thread...(common thread - normal)

\vspace{-0.5em}
\begin{center}
[IMAGE\_PLACEHOLDER]
\end{center}
\vspace{-0.5em}

Thread must be between Magnesium and Copper - Be sure you don’t let the Magnesium \& Copper touch each other. You must make 4 (MEDELIS BATTERY) cells each cell will produce 1,3 volt connect 4 (MEDELIS BATTERIES) so that the positive of one connects to the negative of the ... and you get 5,2 volts Copper (+) Magnesium (-) Now connect LED and your flashlight ready to go...

\vspace{-0.5em}
\begin{center}
[IMAGE\_PLACEHOLDER]
\end{center}
\vspace{-0.5em}

Add Water and This Tiny Emergency Flashlight Glows ...NOW Just Add Water and This Tiny Emergency Flashlight Glows .. Tap Water, drinking water, river,lake,rain or any similar liquid... MEDELIS BATTERY will generates electricity until the water dries up...for many hours (you can always refill water - make it work again) if you do not let water to dried (put it all in a plastic bag) battery will generates electric power for very long, long....long time more info >  MEDELIS BATTERY

\vspace{-0.5em}
\begin{center}
[IMAGE\_PLACEHOLDER]
\end{center}
\vspace{-0.5em}

\#\# Candidate Images (Image 0 to Image 3):

\begin{center}
\begin{minipage}[t]{0.23\linewidth}
\centering
Image 0\\[2pt]
\includegraphics[width=\linewidth]{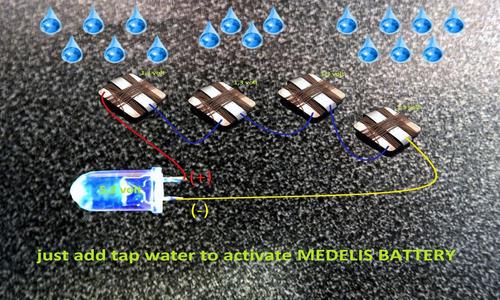}
\end{minipage}
\hfill
\begin{minipage}[t]{0.23\linewidth}
\centering
Image 1\\[2pt]
\includegraphics[width=\linewidth]{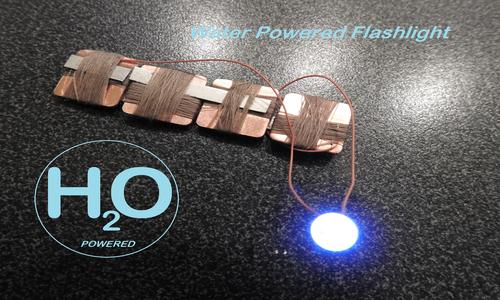}
\end{minipage}
\hfill
\begin{minipage}[t]{0.23\linewidth}
\centering
Image 2\\[2pt]
\includegraphics[width=\linewidth]{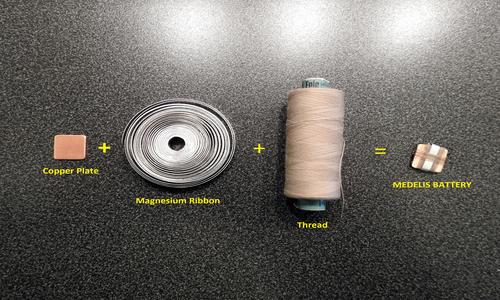}
\end{minipage}
\hfill
\begin{minipage}[t]{0.23\linewidth}
\centering
Image 3\\[2pt]
\includegraphics[width=\linewidth]{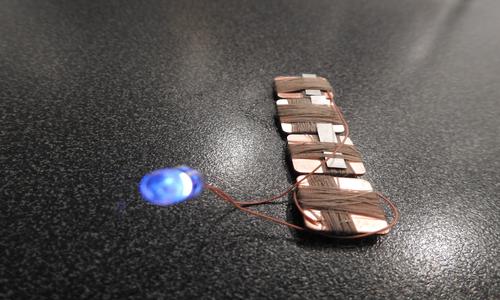}
\end{minipage}
\end{center}

\#\# Answer: [1, 2, 0, 3]
\end{observationbox}

\begin{observationbox}{Science Case 2}
\#\# Article Text (with placeholders):

How to Make a Secret Message With Baking SodaThis Instructable will show you how to create, and expose secret messages using water and baking soda!

\vspace{-0.5em}
\begin{center}
[IMAGE\_PLACEHOLDER]
\end{center}
\vspace{-0.5em}

Gather the Supplies!For this fun activity you'll need the following supplies: Baking Soda Printer paper Water (in a cup) Candle Lighter Paint Brush

\vspace{-0.5em}
\begin{center}
[IMAGE\_PLACEHOLDER]
\end{center}
\vspace{-0.5em}

Mix Baking Soda With WaterPour about a half cup of water into a cup.  With the water at room temperature add a teaspoon of baking soda and stir. If all the baking soda dissolves, add another teaspoon of baking soda until there is a thin amount of baking soda settled on the bottom of the cup.  Now you know your solution is super-saturated (yes, that's a technical term;)

\vspace{-0.5em}
\begin{center}
[IMAGE\_PLACEHOLDER]
\end{center}
\vspace{-0.5em}

Write the Secret MessageUsing a paint brush, paint any picture or text you'll like to be your "secret message".  Paint with a liberal amount of the solution so that you can see the message in wet solution.  Once the message is dry, allow it to dry for 15 about minutes.  once it dries the paper will look like a normal, unused sheet of paper.

\vspace{-0.5em}
\begin{center}
[IMAGE\_PLACEHOLDER]
\end{center}
\vspace{-0.5em}

Reveal the Message!!To reveal the message, expose the paper to a lit candle.  Move the paper back-and-forth slowly to heat up the paper.  The parts of the paper with the message will show dark coloration.  Be careful not to hold the paper in one place over the candle or the paper will catch fire.  Do this step near a sink just in-case it catches fire.  If the paper catches, toss it in the sink and run water over it.

\vspace{-0.5em}
\begin{center}
[IMAGE\_PLACEHOLDER]
\end{center}
\vspace{-0.5em}

You're Have Now Successfully Passed a Secret Message!Your government would be proud!  Write large and clearly, and get plenty of practice in.  It took me and my kids several tries and several kitchen fires to get this working right!  Have fun!

\vspace{-0.5em}
\begin{center}
[IMAGE\_PLACEHOLDER]
\end{center}
\vspace{-0.5em}

\#\# Candidate Images (Image 0 to Image 5):

\begin{center}
\begin{minipage}[t]{0.25\linewidth}
\centering
Image 0\\[2pt]
\includegraphics[width=\linewidth]{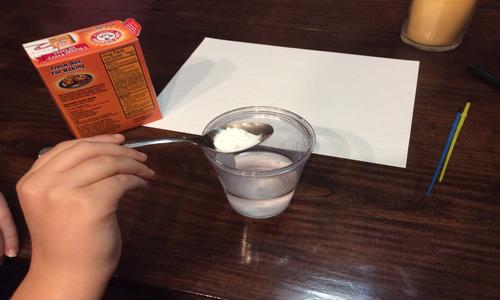}
\end{minipage}
\hfill
\begin{minipage}[t]{0.25\linewidth}
\centering
Image 1\\[2pt]
\includegraphics[width=\linewidth]{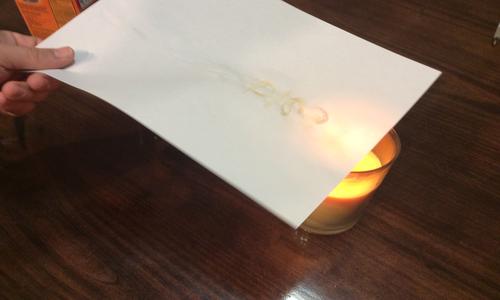}
\end{minipage}
\hfill
\begin{minipage}[t]{0.25\linewidth}
\centering
Image 2\\[2pt]
\includegraphics[width=\linewidth]{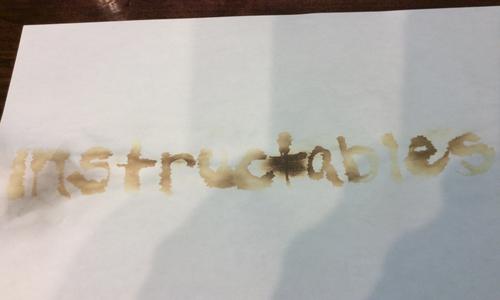}
\end{minipage}

\vspace{0.5em}

\begin{minipage}[t]{0.25\linewidth}
\centering
Image 3\\[2pt]
\includegraphics[width=\linewidth]{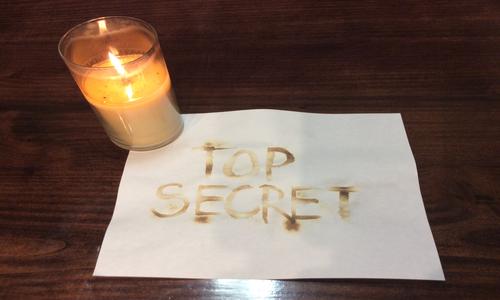}
\end{minipage}
\hfill
\begin{minipage}[t]{0.25\linewidth}
\centering
Image 4\\[2pt]
\includegraphics[width=\linewidth]{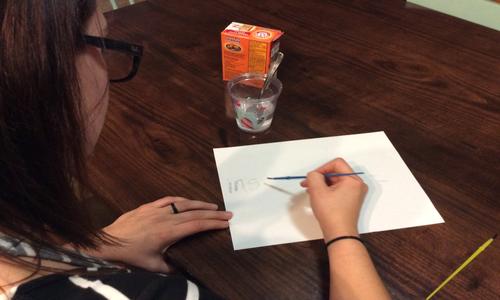}
\end{minipage}
\hfill
\begin{minipage}[t]{0.25\linewidth}
\centering
Image 5\\[2pt]
\includegraphics[width=\linewidth]{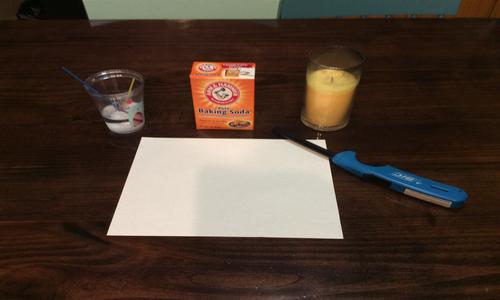}
\end{minipage}
\end{center}

\#\# Answer: [3, 5, 0, 4, 1, 2]
\end{observationbox}

\begin{observationbox}{Science Case 3}
\#\# Article Text (with placeholders):

Phase Inversion Process for Membrane FormationTo use these instructions your are expected to have a knowledge of basic lab safety and equipment. These instructions will explain how to make a membrane using the phase inversion technique, meaning to create a solution and remove the solvent quickly by changing the phase of the solution leaving a porous, solid membrane behind. These instructions are intended for lab researchers that want to try a new technique. In this example I will be making a polysulfone membrane.

\vspace{-0.5em}
\begin{center}
[IMAGE\_PLACEHOLDER]
\end{center}
\vspace{-0.5em}

MaterialsVial with lid large enough to hold solution Materials to create solution Stir bar Stir plate Doctor blade Glass plate to use Doctor Blade on Large tub of water

\vspace{-0.5em}
\begin{center}
[IMAGE\_PLACEHOLDER]
\end{center}
\vspace{-0.5em}

Prepare Your MaterialsWARNING : make sure you know the possible hazards of the chemicals you are using and wear appropriate personal protection equipment. Decide what ratio of materials to use. Calculate how much of each material you will use using the molar mass and density for each substance. Measure out the correct amounts for each material.

\vspace{-0.5em}
\begin{center}
[IMAGE\_PLACEHOLDER]
\end{center}
\vspace{-0.5em}

Create the Solution If you are using a volatile solvent make sure to cover it while you measure out the other materials or measure it last and add it directly to the container holding the solution. Add the measured materials to the vial with lid. Add a stir bar to the solution and cap the vial. Place the vial on the stir plate and stir until fully dissolved. Make sure the stir bar can move fully but is not splashing the solution.

\vspace{-0.5em}
\begin{center}
[IMAGE\_PLACEHOLDER]
\end{center}
\vspace{-0.5em}

Create the MembraneMake sure the doctor blade is ready before uncapping the vial. This Step must be performed quickly to prevent skinning. Performing this step under a fume hood will prevent dangerous gaseous substances from being inhaled and help the membrane dry faster. Set the doctor blade to the desired membrane thickness. Uncap the vial and quickly pour the solution onto the glass plate. The solution can vary dramatically in viscosity depending on what material proportions are used. SLOWLY drag the doctor blade over the solution making sure no pooling occurs at the edge of the blade. Do not drag the blade back over the membrane because it could lead to a nonuniform membrane. Dunk the area of the glass plate with the membrane on it in the water. The membrane should distinctly change phase. For my PSF membrane it immediately changes white. Repeat steps 2-4 until the rest of the solution is used up.

\vspace{-0.5em}
\begin{center}
[IMAGE\_PLACEHOLDER]
\end{center}
\vspace{-0.5em}

Removing the MembraneRemember every solution is different so this Instructable does not guarantee that the membrane will remove cleanly. Try to remove the membrane with a GLOVED hand. If it easily slides off place it on a paper towel. If the membrane does not slide off easily use a sharp object like a razor blade to separate the membrane from the glass plate. Note: it may be necessary to leave the membrane in water to prevent the pores from closing.

\vspace{-0.5em}
\begin{center}
[IMAGE\_PLACEHOLDER]
\end{center}
\vspace{-0.5em}

\#\# Candidate Images (Image 0 to Image 5):

\begin{center}
\begin{minipage}[t]{0.15\linewidth}
\centering
Image 0\\[2pt]
\includegraphics[width=\linewidth]{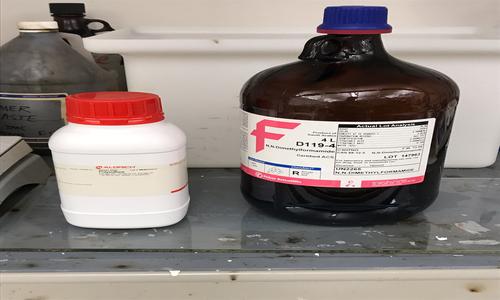}
\end{minipage}
\hfill
\begin{minipage}[t]{0.15\linewidth}
\centering
Image 1\\[2pt]
\includegraphics[width=\linewidth]{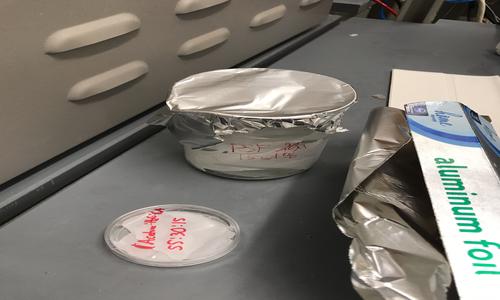}
\end{minipage}
\hfill
\begin{minipage}[t]{0.15\linewidth}
\centering
Image 2\\[2pt]
\includegraphics[width=\linewidth]{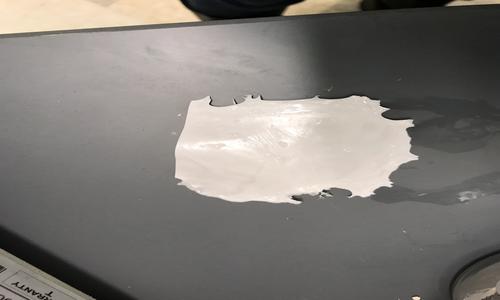}
\end{minipage}
\hfill
\begin{minipage}[t]{0.15\linewidth}
\centering
Image 3\\[2pt]
\includegraphics[width=\linewidth]{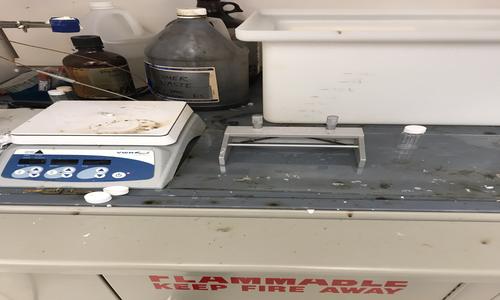}
\end{minipage}
\hfill
\begin{minipage}[t]{0.15\linewidth}
\centering
Image 4\\[2pt]
\includegraphics[width=\linewidth]{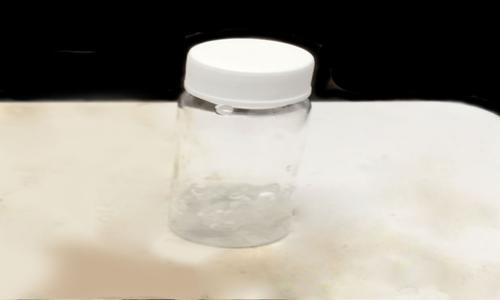}
\end{minipage}
\hfill
\begin{minipage}[t]{0.15\linewidth}
\centering
Image 5\\[2pt]
\includegraphics[width=\linewidth]{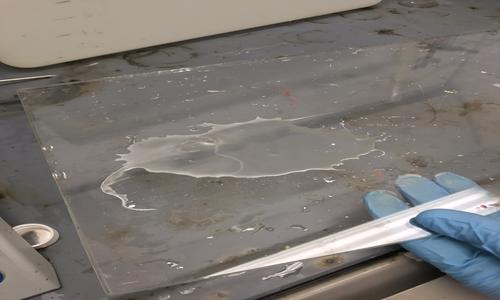}
\end{minipage}
\end{center}

\#\# Answer: [2, 3, 0, 4, 5, 1]
\end{observationbox}

\begin{observationbox}{Science Case 4}
\#\# Article Text (with placeholders):

Turn Grape Juice Into Delicious WineHow to make wine from concentrated grape juice.  This is the recipe we've used to make delicious wine very cheaply. If you are able to get the juice on sale, it works out to less than a dollar a bottle!  The finished wine is fairly sweet (more like a dessert wine) and about 9-10\% alcohol. If you prefer a drier wine, you could adjust the fermentation time and sugar to better suit your tastes.

\vspace{-0.5em}
\begin{center}
[IMAGE\_PLACEHOLDER]
\end{center}
\vspace{-0.5em}

What You'll NeedTo make wine from juice, you will need some basic wine making supplies. All of the equipment should be available at your local home brew and wine making store.   Equipment: -5 gallon carboy (glass container) -Primary fermenter (large food safe plastic bucket) -Airlock -Siphon hose -Corking machine -Wine bottles -Corks   Ingredients: -12 cans of white grape juice -Campden tablets -3kg sugar -Champagne yeast -Yeast nutrient -Sparkolloid powder  Makes 5 gallons

\vspace{-0.5em}
\begin{center}
[IMAGE\_PLACEHOLDER]
\end{center}
\vspace{-0.5em}

Primary FermentationSanitize the primary fermenter and then add the following:  12 cans of white grape juice 3 kg of sugar 48 cans of water 1 package of champagne yeast  Leave for 5 days

\vspace{-0.5em}
\begin{center}
[IMAGE\_PLACEHOLDER]
\end{center}
\vspace{-0.5em}

Racking5 days later, siphon from the primary fermenter into the sanitized carboy (this is called racking). When racking, try to leave all the sludge at the bottom and then discard it.  Rack again 1 month later (2nd racking) and add 1/2 tsp of yeast nutrient.  Rack again 1 month later (3rd racking) and add 2 tsp of yeast nutrient.  5 days after the 3rd racking, add 15g of sparkolloid powder dissolved in 2 1/4 cups of water. Stir vigorously. The sparkolloid powder is a clearing agent that will turn the contents of your carboy from cloudy to crystal clear.

\vspace{-0.5em}
\begin{center}
[IMAGE\_PLACEHOLDER]
\end{center}
\vspace{-0.5em}

Bottle2 weeks after adding the sparkolloid powder, add 5 campden tablets dissolved in boiling water. The campden tablets will stop the fermentation and help keep the wine from going bad.  Sanitize 30 wine bottles (you will probably only need 28 but it's better to have a few extra clean bottles on hand).  Siphon the wine into the bottles and use the corking machine to cork them.  Print labels onto regular printer paper and use a glue stick to attach them to the bottles. While adhesive labels are easy to find they can be difficult to remove from the bottles when it is time to make your next batch.  You can enjoy the wine immediately, but it is even better after a month.

\vspace{-0.5em}
\begin{center}
[IMAGE\_PLACEHOLDER]
\end{center}
\vspace{-0.5em}

\#\# Candidate Images (Image 0 to Image 4):

\begin{center}
\begin{minipage}[t]{0.18\linewidth}
\centering
Image 0\\[2pt]
\includegraphics[width=\linewidth]{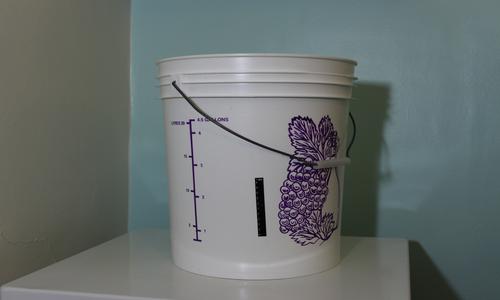}
\end{minipage}
\hfill
\begin{minipage}[t]{0.18\linewidth}
\centering
Image 1\\[2pt]
\includegraphics[width=\linewidth]{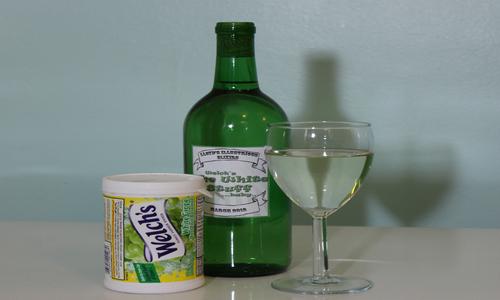}
\end{minipage}
\hfill
\begin{minipage}[t]{0.18\linewidth}
\centering
Image 2\\[2pt]
\includegraphics[width=\linewidth]{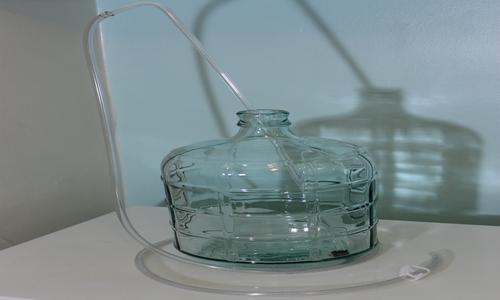}
\end{minipage}
\hfill
\begin{minipage}[t]{0.18\linewidth}
\centering
Image 3\\[2pt]
\includegraphics[width=\linewidth]{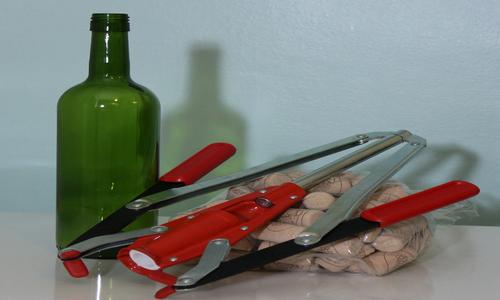}
\end{minipage}
\hfill
\begin{minipage}[t]{0.18\linewidth}
\centering
Image 4\\[2pt]
\includegraphics[width=\linewidth]{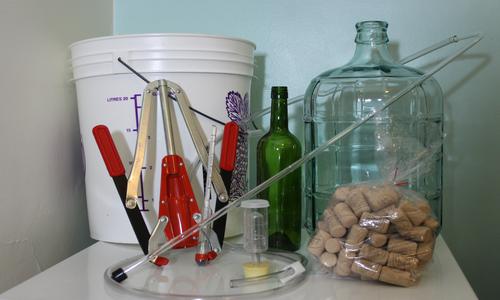}
\end{minipage}
\end{center}

\#\# Answer: [1, 4, 0, 2, 3]
\end{observationbox}

\begin{observationbox}{Science Case 5}
\#\# Article Text (with placeholders):

Chemical Propulsion: Unleashing Flight With Vinegar and Baking SodaA Vinegar and Baking Soda Rocket: Unlocking Flight Through Chemistry In the world of science and innovation, we're exploring how physics and chemistry come together. Imagine using everyday kitchen items to challenge gravity, sending a familiar mix into the sky, sparking curiosity. This is chemical propulsion, where vinegar and baking soda combine, defying expectations and showing us the basics of flight. As we dig into this fascinating vinegar and baking soda experiment, we uncover a simple yet remarkable way to make things fly. Presenting our vinegar and baking soda rocket—a clear example of how science helps us take to the skies. This was designed using Fusion360. I attend Excel Online High School.

\vspace{-0.5em}
\begin{center}
[IMAGE\_PLACEHOLDER]
\end{center}
\vspace{-0.5em}

Preparing the StopperPrepare Your Stopper: Before diving into the experiment, it's important to get the stopper ready. Create a small hole at the bottom of the stopper. This hole is where the baking soda will go. Making this hole will help the baking soda easily fall into the container when it's time to launch the rocket. I was easily able to achieve this with the cork by poking it with a screwdriver until I got a hole of suitable size. Now, let's proceed to the next stages of the experiment, where we'll load the container with the ingredients needed for our exciting chemical reaction and get ready to launch our vinegar and baking soda rocket!

\vspace{-0.5em}
\begin{center}
[IMAGE\_PLACEHOLDER]
\end{center}
\vspace{-0.5em}

Loading the Container1: Prepare Baking Soda: Measure an appropriate amount of baking soda using your measuring spoons or scoops. Start off using small amounts and see how well the rocket performs 2: Insert Baking Soda: Using the funnel, carefully insert the measured baking soda into the hole in the stopper. Make sure it goes all the way into the container. 3: Prepare Vinegar: Measure the vinegar using your measuring spoons or a pouring container. Start off using small amounts and see how well the rocket performs 4: Pour Vinegar into Container: Pour the measured vinegar into the container over the baking soda. Quickly seal the container with the stopper, ensuring a tight fit to create a seal. Experiment with Measurements (Optional): Remember, science is about exploration and discovery! Feel free to experiment with the amounts of baking soda and vinegar. It's more fun to observe how different quantities can affect the rocket's launch. Try varying the amounts slightly and observe the changes in the rocket's flight. This way, you can enjoy the process of discovery and uncover the perfect mixture for an exhilarating launch!

\vspace{-0.5em}
\begin{center}
[IMAGE\_PLACEHOLDER]
\end{center}
\vspace{-0.5em}

Launching the Rocket1: Position for Launch: Place the container on a flat surface in your safety area, with the stopper end facing down. 2: Move to a Safe Distance: Move to a safe distance, keeping your eyes protected with the safety goggles. 3: Observe and Launch: As the vinegar reacts with the baking soda, building pressure inside the container, observe as the chemical reaction unfolds. After a brief delay, the stopper will launch into the air. 4: Analyze and Learn: Pay close attention to the flight of the stopper. Note its height, trajectory, and any other observations. This can provide valuable insights into the physics of chemical propulsion and the basics of flight.

\vspace{-0.5em}
\begin{center}
[IMAGE\_PLACEHOLDER]
\end{center}
\vspace{-0.5em}

\#\# Candidate Images (Image 0 to Image 3):

\begin{center}
\begin{minipage}[t]{0.23\linewidth}
\centering
Image 0\\[2pt]
\includegraphics[width=\linewidth]{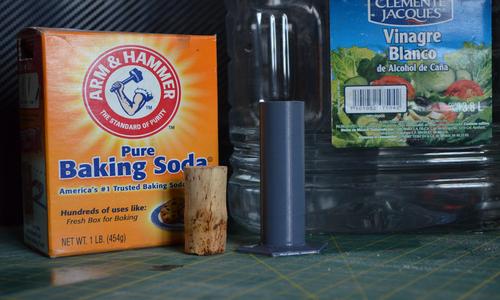}
\end{minipage}
\hfill
\begin{minipage}[t]{0.23\linewidth}
\centering
Image 1\\[2pt]
\includegraphics[width=\linewidth]{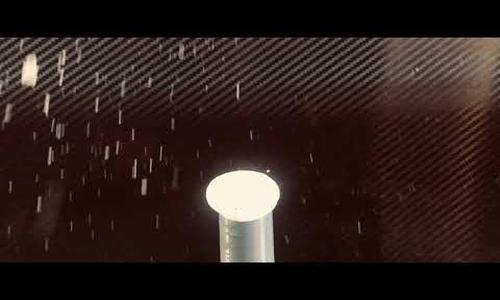}
\end{minipage}
\hfill
\begin{minipage}[t]{0.23\linewidth}
\centering
Image 2\\[2pt]
\includegraphics[width=\linewidth]{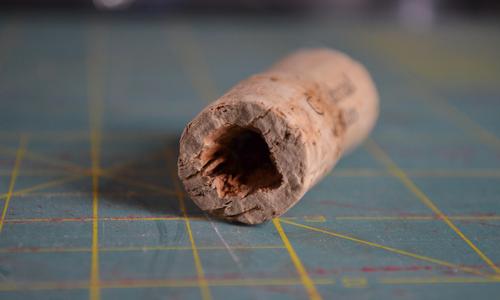}
\end{minipage}
\hfill
\begin{minipage}[t]{0.23\linewidth}
\centering
Image 3\\[2pt]
\includegraphics[width=\linewidth]{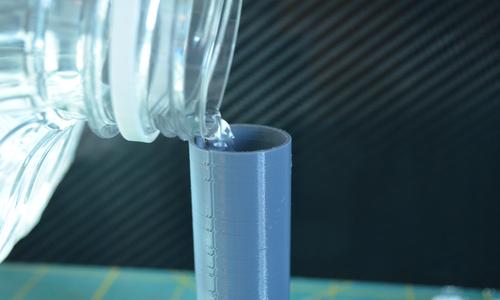}
\end{minipage}
\end{center}
\#\# Answer: [0, 2, 3, 1]
\end{observationbox}

\clearpage
\subsection{Error Case Studies}
\label{appdix:error_cases}
We use Qwen3.5-397B-A17B as the judge model, and provide the corresponding prompt in Appendix~\ref{appendix:error_analysis_prompt}. For readability, in this section we present only the original context, the model output, and the judge model output.
\begin{takeawaybox}{Error Case 1: Fine-Detail Miss}

\textbf{Article Text (with placeholders):}

You are given an article about "Broken TV" with 4 image placeholders marked as [IMAGE\_PLACEHOLDER]. You are also given 4 candidate images (Image 0, Image 1, ..., Image 3) shown below.

Your task is to determine which image should be placed at each placeholder position based on the surrounding text context.

\#\# Article Text (with placeholders):

Ah. Time to watch Birdflix.
\vspace{-0.5em}
\begin{center}
[IMAGE\_PLACEHOLDER]
\end{center}
\vspace{-0.5em}
Go team go!
\vspace{-0.5em}
\begin{center}
[IMAGE\_PLACEHOLDER]
\end{center}
\vspace{-0.5em}
Huh? No, no what happened!?
\vspace{-0.5em}
\begin{center}
[IMAGE\_PLACEHOLDER]
\end{center}
\vspace{-0.5em}
NOOO!!
\vspace{-0.5em}
\begin{center}
[IMAGE\_PLACEHOLDER]
\end{center}
\vspace{-0.5em}

\vspace{0.5em}
\textbf{Candidate Images (Image 0 to Image 3):}

\begin{center}
\begin{minipage}[t]{0.22\linewidth}
\centering
Image 0\\[2pt]
\includegraphics[width=\linewidth]{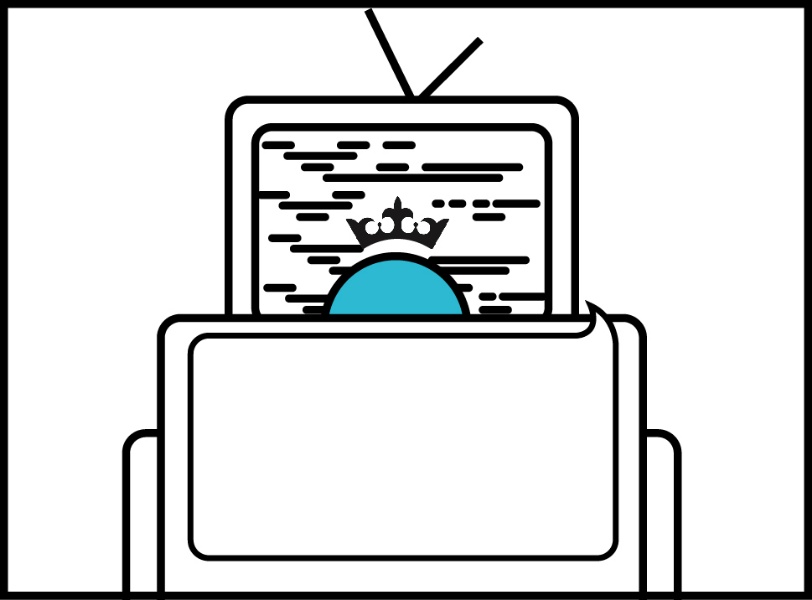}
\end{minipage}
\hfill
\begin{minipage}[t]{0.22\linewidth}
\centering
Image 1\\[2pt]
\includegraphics[width=\linewidth]{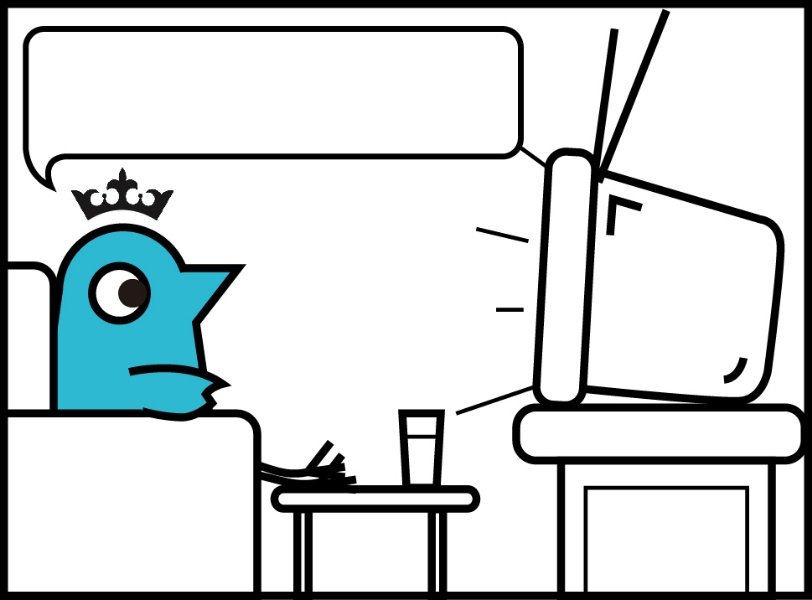}
\end{minipage}
\hfill
\begin{minipage}[t]{0.22\linewidth}
\centering
Image 2\\[2pt]
\includegraphics[width=\linewidth]{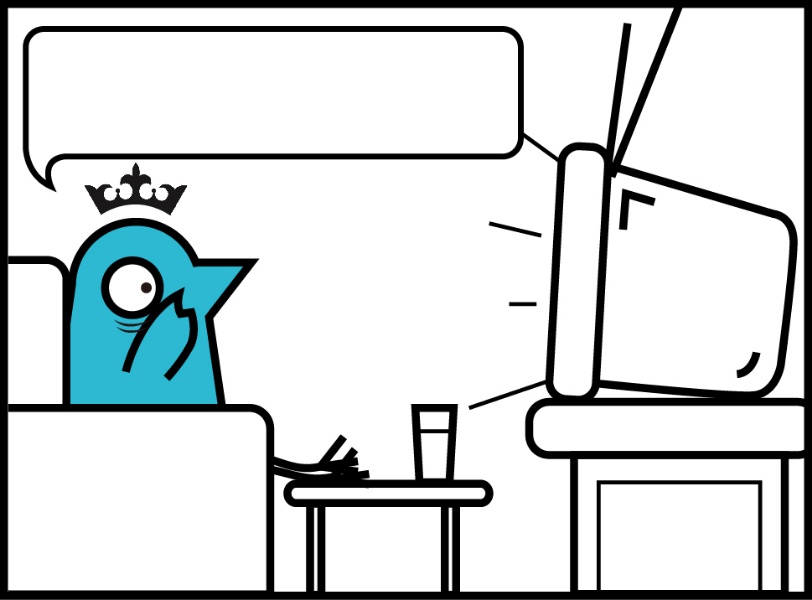}
\end{minipage}
\hfill
\begin{minipage}[t]{0.22\linewidth}
\centering
Image 3\\[2pt]
\includegraphics[width=\linewidth]{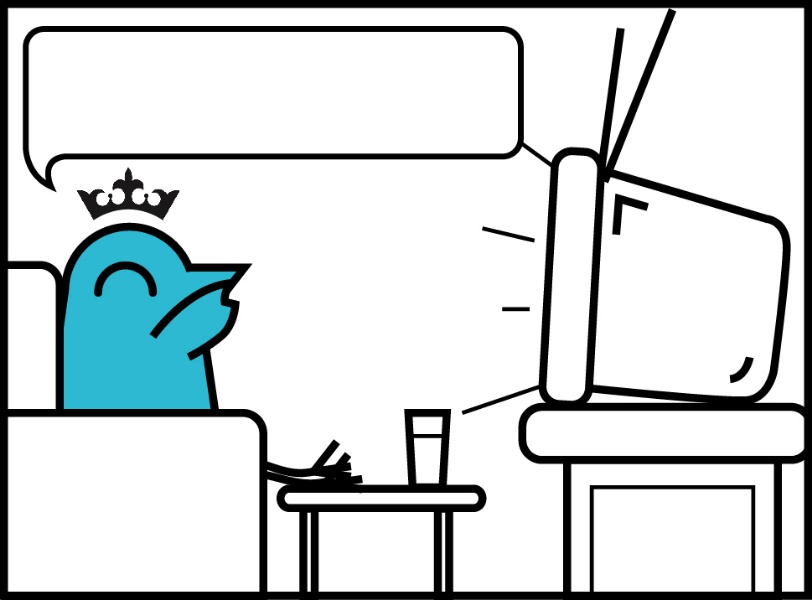}
\end{minipage}
\end{center}
\textbf{Ground Truth}

[1, 3, 2, 0]

\noindent\rule{\linewidth}{0.1pt}

\textbf{Model output} 

\#\#\# Step-by-Step Reasoning

**First placeholder: "Ah. Time to watch Birdflix."**
→ This is the beginning. The bird is about to start watching. Image 1 shows the bird sitting down, TV on, with a neutral expression — perfect for the start. Image 0 shows the TV screen with the logo, which might be what’s on screen when the show starts — but since the text is from the bird’s perspective ("Ah. Time to watch..."), the image should show the bird’s reaction. Image 1 is more appropriate here.

Wait — let’s reconsider. The text is "Ah. Time to watch Birdflix." — this could be the moment the show starts, and Image 0 shows the TV screen with the Birdflix logo. That might be the visual cue for the start of the show. But the other images show the bird watching, which is the narrative perspective.

Looking again, Image 0 is the only one that doesn’t show the bird — it’s a close-up of the TV. The other three (1, 2, 3) show the bird watching TV from the same angle.

Given that, perhaps Image 0 is meant to be the TV screen when the show starts — the logo appears.

But the text "Ah. Time to watch Birdflix." is likely spoken by the bird, so the image should show the bird’s perspective or reaction. Image 1 shows the bird sitting down, TV on, neutral — that fits "Ah. Time to watch..." as the bird settles in.

Alternatively, maybe Image 0 is the screen content, and the others are the bird’s reactions. That would make sense for interleaving.

Let’s try this mapping:

- **Placeholder 1: "Ah. Time to watch Birdflix."** → Image 0: TV screen with Birdflix logo. This is the first thing you see when you turn on the show.

- **Placeholder 2: "Go team go!"** → Image 3: Bird is smiling and happy — clearly enjoying the show, cheering.

- **Placeholder 3: "Huh? No, no what happened!?"** → Image 2: Bird is shocked, eyes wide — perfect for confusion.

- **Placeholder 4: "NOOO!!"** → Image 1: Bird has a sad or concerned expression — fits the despair of "NOOO!!"

This mapping uses all images once and matches the emotional progression:

1. Show starts → TV screen with logo (Image 0)
2. Excitement → Bird cheering (Image 3)
3. Shock → Bird surprised (Image 2)
4. Despair → Bird upset (Image 1)

This seems logical.

Let’s verify:

- Image 0: TV screen with logo — matches "Time to watch Birdflix" (the show is starting).
- Image 3: Bird smiling — matches "Go team go!" (happy, cheering).
- Image 2: Bird shocked — matches "Huh? No, no what happened!?" (surprised).
- Image 1: Bird with a downturned mouth — matches "NOOO!!" (sad, disappointed).

Yes, this is consistent.

So the final mapping:

Placeholder 1 → Image 0  
Placeholder 2 → Image 3  
Placeholder 3 → Image 2  
Placeholder 4 → Image 1  

Answer: [0, 3, 2, 1]

\noindent\rule{\linewidth}{0.1pt}

\textbf{Error Analysis} 

"error\_reason": "The model incorrectly swapped the first and last images, placing the broken TV screen (Image 0) at the beginning and the neutral bird (Image 1) at the climax. This error primarily stems from a Fine-Detail Miss, as the model failed to recognize the static lines on the TV in Image 0 as a decisive cue for a broken signal (fitting the 'NOOO!!' ending) and instead interpreted the logo as the show starting. Additionally, the model committed Visual Hallucination by describing the bird in Image 1 as having a 'sad or concerned expression' to justify its placement at the end, despite the image showing a neutral state. These perceptual failures resulted in a Global Assignment Drift where the narrative arc is structurally inverted.",

"primary\_error\_type": "Fine-Detail Miss",

"secondary\_error\_types": [
  "Visual Hallucination",
  "Global Assignment Drift"
]
    
\end{takeawaybox}

\clearpage

\begin{takeawaybox}{Error Case 2: Semantic Over-Interpretation}

\textbf{Article Text (with placeholders):}

Recognize that the brain doesn’t always store or recall information accurately. The human brain is an amazing thing, but it’s not perfect. While you might think that your memory is like a camera that records what happens, that’s actually not the case. Memories are constructive and they can change over time, especially as they are recalled, leading to memory distortion.   There are two types of memory: verbatim and gist. Verbatim memories are precise and detailed, while gist memories are fuzzier recollections. Both types of memory actually increase as you age.   “Fuzzy trace theory” refers to people’s false memories that come about from related ideas from a gist memory. For instance, if you’re given a list of words and asked to remember them, you might be certain that “hot” was on the list, when it actually contained similar words like scorching, bright, heat, sun, and so on.

\vspace{-0.5em}
\begin{center}
[IMAGE\_PLACEHOLDER]
\end{center}
\vspace{-0.5em}

Keep in mind that suggestions from others can contribute to false memories. If you’re trying to recall something and someone else is asking you questions about it, you may be more likely to misremember the details. This is especially true if the person is asking leading or closed questions, like “The car was blue, right?”  Unfortunately, this happens with police officers during interrogations and can lead to false accusations and even false confessions from innocent people. It’s also common during trials when prosecutors ask leading questions to eyewitness and jurors misremember eyewitness testimony containing those details, rather than an attorney implying them.

\vspace{-0.5em}
\begin{center}
[IMAGE\_PLACEHOLDER]
\end{center}
\vspace{-0.5em}

Remember that you might have more false memories if you have an active imagination. If you’re a very creative person with a strong imagination, you might have more false memories than the average person. This is because you’re able to add an emotional aspect, or even sensory details, to an imagined scenario, which makes it feel more realistic.   The same is true in people who tend to fantasize a lot. This could mean that some of your most emotional memories aren’t real memories.  Having a good imagination or a strong sense of fantasy isn’t a bad thing! Don’t let the possibility of false memories stifle your creativity. Nearly all people have false memories, and that’s not necessarily a bad thing.

\vspace{-0.5em}
\begin{center}
[IMAGE\_PLACEHOLDER]
\end{center}
\vspace{-0.5em}

Keep in mind that you’re more likely to create false memories if you’ve suffered from trauma. If you are depressed, have a history of trauma, or have post traumatic stress disorder (PTSD), you might be more likely to inaccurately remember things, especially if such memories are associated with emotional experiences.   Additionally, people with psychopathy like borderline personality disorder have more false memories.   There’s some evidence that people undergoing therapy have falsely recalled traumatic events, including childhood abuse and sexual abuse. This is more likely if the therapist uses hypnosis, visualization, or suggestion, which indicates that mental health professionals should be particularly cautious about the way they seek to help patients recover repressed memories.   People who misremember childhood events, particularly trauma and abuse, that actually didn’t occur, are said to have False Memory Syndrome.

\vspace{-0.5em}
\begin{center}
[IMAGE\_PLACEHOLDER]
\end{center}
\vspace{-0.5em}

\vspace{0.5em}
\textbf{Candidate Images (Image 0 to Image 3):}

\begin{center}
\begin{minipage}[t]{0.23\linewidth}
\centering
Image 0\\[2pt]
\includegraphics[width=\linewidth]{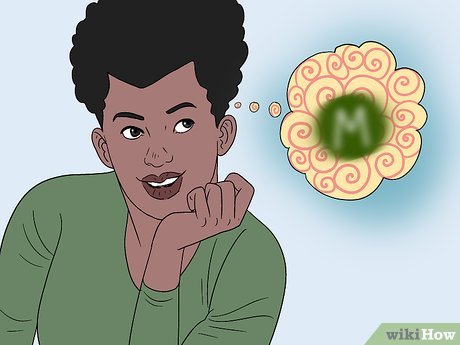}
\end{minipage}
\hfill
\begin{minipage}[t]{0.23\linewidth}
\centering
Image 1\\[2pt]
\includegraphics[width=\linewidth]{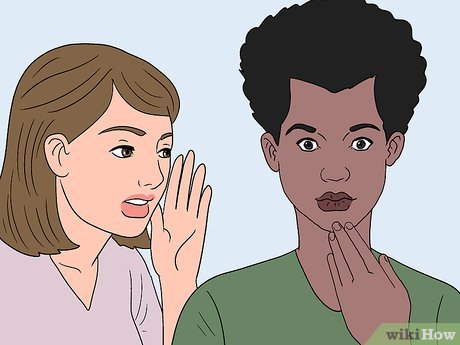}
\end{minipage}
\hfill
\begin{minipage}[t]{0.23\linewidth}
\centering
Image 2\\[2pt]
\includegraphics[width=\linewidth]{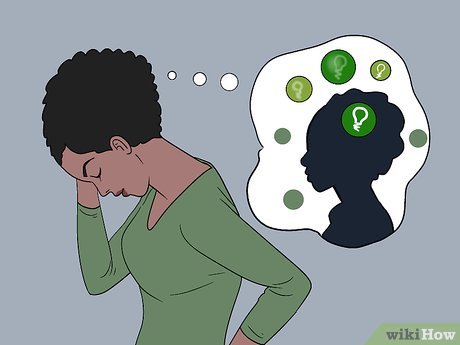}
\end{minipage}
\hfill
\begin{minipage}[t]{0.23\linewidth}
\centering
Image 3\\[2pt]
\includegraphics[width=\linewidth]{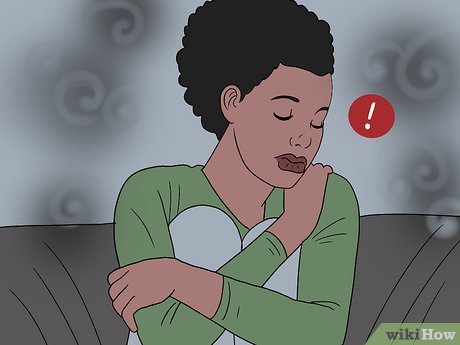}
\end{minipage}
\end{center}

\textbf{Ground Truth}

[2, 1, 0, 3]

\noindent\rule{\linewidth}{0.1pt}

\textbf{Model output} 

\#\#\# Step-by-Step Reasoning

1. **First placeholder context**: The section explains that memories are constructive, introduces verbatim/gist memory types, and fuzzy trace theory where fuzzy, meaning memories are often blurry and not perfectly precise. Image 0 matches this best: it shows a person with a thought bubble containing a blurry, fuzzy letter, visually representing fuzzy gist memory.

2. **Second placeholder context**: The section discusses how suggestions from other people (like leading questions from police/prosecutors) lead to false memories. Image 1 matches this: it shows one person whispering to another, directly representing another person sharing information/suggestions to someone else.

3. **Third placeholder context**: The section covers how active imagination and creativity lead to more false memories, as creative people add details to imagined scenarios. Image 2 matches this: it shows a person with a thought bubble containing lightbulbs (representing ideas/imagination) around a head silhouette, visually representing creative thought and imagination.

4. **Fourth placeholder context**: The section explains that trauma, PTSD, and related mental health conditions increase false memories and False Memory Syndrome. Image 3 matches this: it shows a distressed, sad person, visually representing the distress associated with trauma and related mental health struggles.

[0, 1, 2, 3]

\noindent\rule{\linewidth}{0.1pt}

\textbf{Error Analysis} 

"error\_reason": "The model incorrectly swapped Image 0 and Image 2 by prioritizing literal keyword matching over broader contextual fit. It associated the word 'fuzzy' in the first section with the visually 'fuzzy' letter in Image 0, overlooking that the daydreaming pose in Image 0 better suits the 'active imagination' context of the third section. This locally plausible but semantically shallow match forced a globally inconsistent mapping where the 'imagination' text was paired with a 'confused/stressed' image (Image 2) instead of the 'daydreaming' image.",

"primary\_error\_type": "Semantic Over-Interpretation",

"secondary\_error\_types": [
  "Fine-Detail Miss",
  "Global Assignment Drift"
]
    
\end{takeawaybox}

\clearpage

\begin{takeawaybox}{Error Case 3: Step-State Confusion}

\textbf{Article Text (with placeholders):}

House your guinea pigs in a solid enclosure. When a cat enters kill mode, a flimsy cage won't stop it. While there are steps to follow that will reduce the likelihood of your cat becoming unnecessarily aggressive, it is still worthwhile to have a cage for your guinea pigs that guarantees their safety from demise by tooth and claw. Recognize that your cat is also capable of seriously harming your guinea pig without intending to do so. As such, your guinea pig needs an area where they are completely safe, even if only from some overzealous rough-housing.

\vspace{-0.5em}
\begin{center}
[IMAGE\_PLACEHOLDER]
\end{center}
\vspace{-0.5em}

Establish specific areas where each pet is allowed to be. If possible, you'll want to keep your guinea pigs in a safe room where your cats are not regularly allowed. If you do not have sufficient space to allow this arrangement, choose a location for the cage that will prevent your cats from hanging out on or around the guinea pig enclosure.   Always keep all cat toys and guinea pig toys separated in your house.  If you have a guinea pig area, keep all of their worldly possessions in that area, and do not store or allow any cat stuff in that area.  Disallow your cat from sitting beside or resting on top of your guinea pig's enclosure at any time.

\vspace{-0.5em}
\begin{center}
[IMAGE\_PLACEHOLDER]
\end{center}
\vspace{-0.5em}

Allow both pets to become comfortable in their home environments independently. Especially if one arrives after you've already had the other for a while, allow the new arrival to acclimate to your home before introducing them to each other.  They will already be able to smell one another's presence. Let the suspense build a bit. Wait at least a week before introducing a cat and a guinea pig.  Technically, you just need to make sure they are not able to touch one another or make eye contact.  To be safe, don't allow them to see one another at all.

\vspace{-0.5em}
\begin{center}
[IMAGE\_PLACEHOLDER]
\end{center}
\vspace{-0.5em}

Emphasize the other pet's presence. Before actually introducing your pets, acclimate your cat to the guinea pig's scent.  There are several ways to do this.  The intent is to associate things your cat likes, such as food, with the guinea pig. This may seem counterintuitive, but the cat will not think of the guinea pig as food – instead, it will think the guinea pig smell indicates that other (and more appropriate) food is nearby.   Place the cat's food just outside the door of the room in which your guinea pig lives.  If your cat gets amped up outside the guinea pig's room, remove them.  Do so in a jovial, relaxed manner – you don't want to indicate punishment, you just want to distract the cat from its overzealousness.   Place a washcloth that you rubbed the guinea pig with near the cat's food bowl.  Once your cat is entirely comfortable with the washcloth (no longer nervously or excitedly sniffing it from time to time), use the washcloth to stroke the cat as well.  Use the same cloth to go back and forth between your pets, stroking each right after you stroked the other. As you do so, be relaxed and high spirited.  Send the vibe to each pet that everything is awesome!

\vspace{-0.5em}
\begin{center}
[IMAGE\_PLACEHOLDER]
\end{center}
\vspace{-0.5em}

\textbf{Candidate Images (Image 0 to Image 3):}

\begin{center}
\begin{minipage}[t]{0.23\linewidth}
\centering
Image 0\\[2pt]
\includegraphics[width=\linewidth]{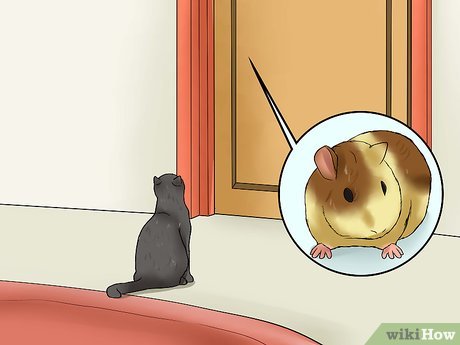}
\end{minipage}
\hfill
\begin{minipage}[t]{0.23\linewidth}
\centering
Image 1\\[2pt]
\includegraphics[width=\linewidth]{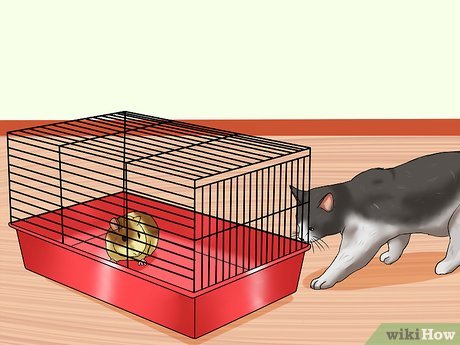}
\end{minipage}
\hfill
\begin{minipage}[t]{0.23\linewidth}
\centering
Image 2\\[2pt]
\includegraphics[width=\linewidth]{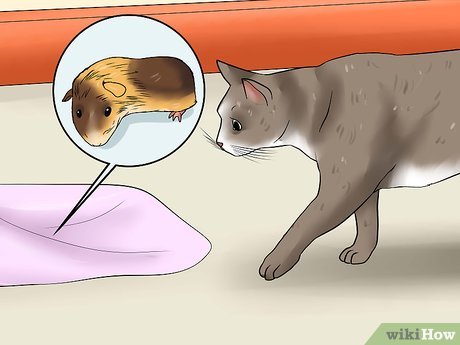}
\end{minipage}
\hfill
\begin{minipage}[t]{0.23\linewidth}
\centering
Image 3\\[2pt]
\includegraphics[width=\linewidth]{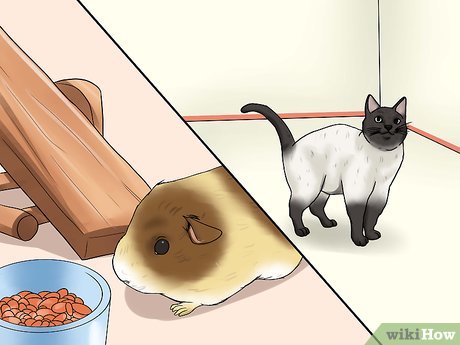}
\end{minipage}
\end{center}

\textbf{Ground Truth}

[1, 3, 0, 2]

\noindent\rule{\linewidth}{0.1pt}

\textbf{Model output} 

\#\#\# Step-by-Step Reasoning

1. **First placeholder context**: The section explains that memories are constructive, introduces verbatim/gist memory types, and fuzzy trace theory where fuzzy, meaning memories are often blurry and not perfectly precise. Image 0 matches this best: it shows a person with a thought bubble containing a blurry, fuzzy letter, visually representing fuzzy gist memory.

2. **Second placeholder context**: The section discusses how suggestions from other people (like leading questions from police/prosecutors) lead to false memories. Image 1 matches this: it shows one person whispering to another, directly representing another person sharing information/suggestions to someone else.

3. **Third placeholder context**: The section covers how active imagination and creativity lead to more false memories, as creative people add details to imagined scenarios. Image 2 matches this: it shows a person with a thought bubble containing lightbulbs (representing ideas/imagination) around a head silhouette, visually representing creative thought and imagination.

4. **Fourth placeholder context**: The section explains that trauma, PTSD, and related mental health conditions increase false memories and False Memory Syndrome. Image 3 matches this: it shows a distressed, sad person, visually representing the distress associated with trauma and related mental health struggles.

[1, 0, 3, 2]

\noindent\rule{\linewidth}{0.1pt}

\textbf{Error Analysis} 

"error\_reason": "The model correctly identified the first and last images but swapped the assignments for the second and third placeholders, confusing the sequence of establishing separate areas versus independent acclimation. Both sections discuss separation, leading the model to misalign Image 0 (cat at a closed door) and Image 3 (split-screen of pets) with the corresponding text. The model missed the specific textual cue 'safe room' in the second section which aligns with the closed door in Image 0, and instead over-interpreted the split-screen layout of Image 3 as representing 'specific areas' rather than independent environments.",

"primary\_error\_type": "Step-State Confusion",

"secondary\_error\_types": [
  "Fine-Detail Miss",
  "Semantic Over-Interpretation"
]
    
\end{takeawaybox}

\clearpage

\begin{takeawaybox}{Error Case 4: Visual Hallucination}

\textbf{Article Text (with placeholders):}

Draft up a concrete image of what you want to become. Making a major life change, from ending a long-term relationship to switching careers, is usually only terrifying because you don’t know what is coming next. That uncertainty can paralyze you if you don’t take the time to figure out where exactly you are going. You don’t need to know everything, no one ever could, but you do need a vision for how you are going to change.   What do you want to eliminate from your life?  What do you want to add?  Where do you see yourself 1 year after you’ve made your change?  What do you want, more than anything else, to spend your time doing?

\vspace{-0.5em}
\begin{center}
[IMAGE\_PLACEHOLDER]
\end{center}
\vspace{-0.5em}

Plan specific ways to change your lifestyle. Once you have a good idea of where you are going, you need to figure out how to get there. This is often the hardest part of changing, but it becomes much easier if you think of it in reverse. Say your goal is to become a famous author. To make this change a reality, think of the steps that would lead to becoming a famous author until you get to one you can work on:   Goal: To Be a Famous Author.  Get a book published.  Find a literary agent.  Write and edit a book.  Write every single day.  Draft up ideas for books. If you don’t have an idea yet, you would start here. If you do, it’s time to write every day!

\vspace{-0.5em}
\begin{center}
[IMAGE\_PLACEHOLDER]
\end{center}
\vspace{-0.5em}

Save up. Making a big change in life is a lot easier if you have a safety net to fall back on. You’re more likely to take a plunge when you know that failure doesn’t mean the end of the world, so save up a little extra money. This will allow you to focus on changing your life, not on paying the bills.  Open a savings account and start putting in a small percentage (5-10\%) of your paycheck in it.  Many financial advisors suggest having enough money to cover at least 6 months of living expenses before making a large change, like a move or career change.

\vspace{-0.5em}
\begin{center}
[IMAGE\_PLACEHOLDER]
\end{center}
\vspace{-0.5em}

Get educated. You never want to make a major lifestyle change without some knowledge of what you are in for. If you want to start a new career, going back to school is often the best way to get on track, as the specific knowledge will prepare you for a life in the field you desire. Even those looking for more “off-beat” changes, like traveling for a year or becoming an artist, need to study to get the most of their lifestyle change.  Look up autobiographies of similar people. While you don’t have to follow in their footsteps, they offer valuable advice about what to expect as you change.  Spend time researching your new change – what kind of equipment do you need? Do you need to move locations? What are the negatives of your new lifestyle, and do they make you want to change any less?

\vspace{-0.5em}
\begin{center}
[IMAGE\_PLACEHOLDER]
\end{center}
\vspace{-0.5em}

Get out of your old life quickly and respectfully. Once you’ve made the decision to make a change and you are confident it’s time to start, you need to cut your old ties. This does not mean that you will never see people from your “old life” again. Rather, it means that you need to take some time away from the routines, habits, and lifestyle to truly change things. Never burn bridges with cruel goodbyes or anger. Instead, let people know that you are ready for a change and would love their support as you go through with this.

\vspace{-0.5em}
\begin{center}
[IMAGE\_PLACEHOLDER]
\end{center}
\vspace{-0.5em}

Work on making your new change a reality every day. You need to commit to your new life fully if you ever expect to change. Sometimes this is easy – if you want to travel for a year you need to get on a plane and get out of the country. But sometimes this requires daily discipline. No matter how you cut it, for example, you need to write every single day if you want to be a famous author. Remember, change is about choice. Make the choices that bring about the change you desire.

\vspace{-0.5em}
\begin{center}
[IMAGE\_PLACEHOLDER]
\end{center}
\vspace{-0.5em}

\vspace{0.5em}
\textbf{Candidate Images (Image 0 to Image 5):}

\begin{center}
\begin{minipage}[t]{0.22\linewidth}
\centering
Image 0\\[2pt]
\includegraphics[width=\linewidth]{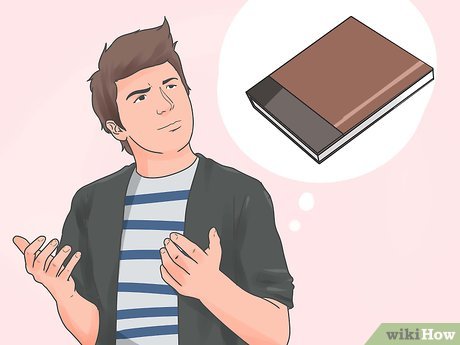}
\end{minipage}
\hfill
\begin{minipage}[t]{0.22\linewidth}
\centering
Image 1\\[2pt]
\includegraphics[width=\linewidth]{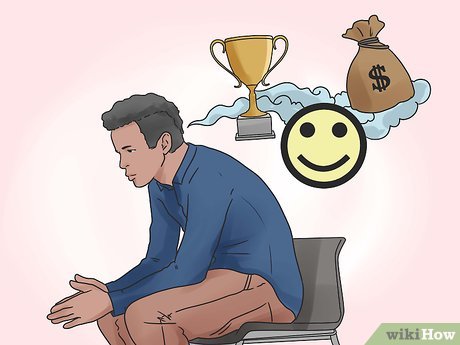}
\end{minipage}
\hfill
\begin{minipage}[t]{0.22\linewidth}
\centering
Image 2\\[2pt]
\includegraphics[width=\linewidth]{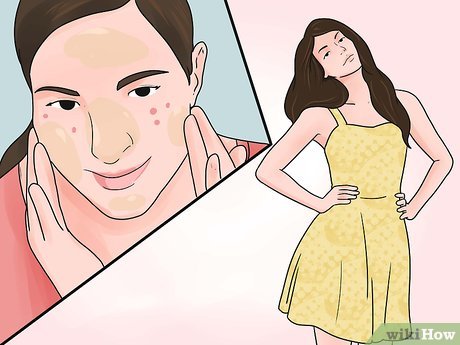}
\end{minipage}

\begin{minipage}[t]{0.22\linewidth}
\centering
Image 3\\[2pt]
\includegraphics[width=\linewidth]{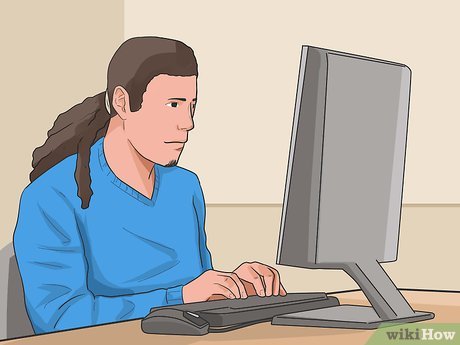}
\end{minipage}
\hfill
\begin{minipage}[t]{0.22\linewidth}
\centering
Image 4\\[2pt]
\includegraphics[width=\linewidth]{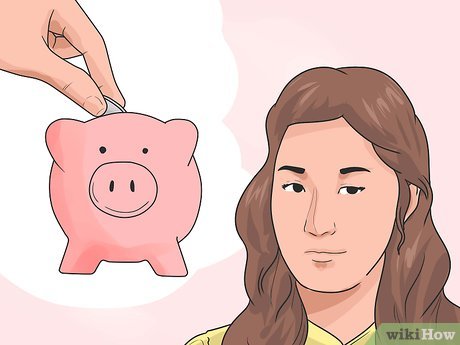}
\end{minipage}
\hfill
\begin{minipage}[t]{0.22\linewidth}
\centering
Image 5\\[2pt]
\includegraphics[width=\linewidth]{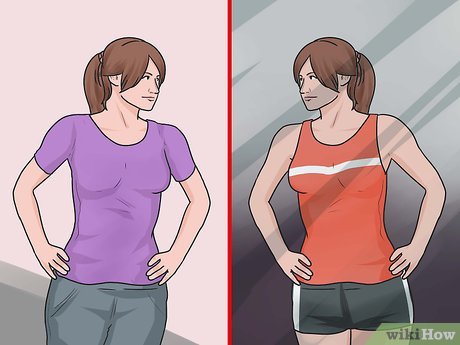}
\end{minipage}
\end{center}

\textbf{Ground Truth}

[1, 0, 4, 3, 5, 2]

\noindent\rule{\linewidth}{0.1pt}

\textbf{Model output} 

\#\#\# Step-by-Step Reasoning

1. **First placeholder**: The surrounding text is about envisioning your desired future goals (success, happiness, financial security etc.). Image 1 perfectly matches this, as it shows a person thinking about a trophy (success), money bag (financial stability), and smiley face (happiness), all typical desired outcomes for life changes.

2. **Second placeholder**: The text uses becoming a famous author as the core example for planning lifestyle change steps, centered on writing books. Image 0 shows a person thinking of a book, directly matching the author/book example in this section.

3. **Third placeholder**: The section focuses on saving money as a safety net for life changes. Image 4 depicts putting a coin into a piggy bank, which is a direct visual representation of saving money, so it matches this section.

4. **Fourth placeholder**: The text advises researching and getting educated about your planned change (e.g. looking up information online, studying your new path). Image 3 shows a person working on a computer, which aligns with researching/studying to get informed about the change.

5. **Fifth placeholder**: The section covers respectfully leaving your old life and saying goodbye to old ties without burning bridges. Image 2 shows a person waving goodbye to another person, which fits the description of polite, amicable separation from old connections.

6. **Sixth placeholder**: The text is about committing to daily action to make your new life a reality, resulting in a full transformation. Image 5 shows a before-and-after of the same person (old self vs new transformed self), which directly represents the successful life change outcome described here.

[1, 0, 4, 3, 2, 5]

\noindent\rule{\linewidth}{0.1pt}

\textbf{Error Analysis} 

"error\_reason": "The model's prediction is incorrect because it swaps the assignments for the fifth and sixth placeholders, primarily driven by a Visual Hallucination where it claims Image 2 shows a person 'waving goodbye.' This action is not present in Image 2, which actually depicts a woman examining her skin and standing in a dress, leading the model to falsely align it with the text about cutting ties. This error caused a Step-State Confusion between the final two adjacent sections, as the model mixed up the visually similar transformation-style images (Image 2 and Image 5). Additionally, the model committed a Fine-Detail Miss by failing to identify the specific content of Image 2, which would have prevented the forced alignment with the 'goodbye' context.",

"primary\_error\_type": "Visual Hallucination",

"secondary\_error\_types": [
  "Step-State Confusion",
  "Fine-Detail Miss"
]
    
\end{takeawaybox}

% \clearpage

\begin{takeawaybox}{Error Case 5: Global Assignment Drift}

\textbf{Article Text (with placeholders):}

Force your dog to gently swallow the pill. Do this if you cannot get your dog to take his medicine another way. This may feel a bit extreme to do, but in some cases it can be entirely necessary. Don't worry, you won't choke your dog. By taking your time, and by being gentle, this can be a simple, surefire way to get your dog to take his medicine.
\vspace{-0.5em}
\begin{center}
[IMAGE\_PLACEHOLDER]
\end{center}
\vspace{-0.5em}
Start opening his jaws from the back of the mouth with one hand. Then, use your second hand to lift from the roof of his mouth. Fold his lips over his teeth in order to help prevent your dog from biting. Go slow so as not to hurt your dog. Do not cover his nose.
\vspace{-0.5em}
\begin{center}
[IMAGE\_PLACEHOLDER]
\end{center}
\vspace{-0.5em}
Open your dog's mouth wide, and place the medicine inside. Try to put it as far back as possible. You want to maximize the possibility that your dog will swallow the pill. By placing it as far back as you comfortably can, you will increase the chance that this will happen. If you don't put it in far enough, the dog will spit it out.
\vspace{-0.5em}
\begin{center}
[IMAGE\_PLACEHOLDER]
\end{center}
\vspace{-0.5em}
Hold his mouth shut gently. Do this until your dog swallows. When a dog has swallowed can be difficult to perceive at first. Always check after, just to make sure the medicine is not still in your dog's mouth. It may seem tough, but hold a little bit to ensure your dog gets his medication.  Blow gently on his nose to help aid this process.  Rub your hand over the dog's throat after your dog swallows the pill to make sure that it goes down smoothly. This will stimulate his swallow reflex and your dog will have to swallow.  Give your dog a drink of water after your dog swallows the pill to ensure that the pill doesn't get stuck in its throat, which may cause an ulcer.  Be patient, calm but firm.
\vspace{-0.5em}
\begin{center}
[IMAGE\_PLACEHOLDER]
\end{center}
\vspace{-0.5em}
Give treats after your dog swallows. Use high value treats. Give lots of them before, and especially after. Your dog won't mind this experience as much if your dog gets a huge reward after. Make sure you take the time to do this, especially if you will have to give your dog pills often. If your dog learns that this will only be an unpleasant experience, it will be more difficult to work with your dog.
\vspace{-0.5em}
\begin{center}
[IMAGE\_PLACEHOLDER]
\end{center}
\vspace{-0.5em}

\vspace{0.5em}
\textbf{Candidate Images (Image 0 to Image 4):}

\begin{center}
\begin{minipage}[t]{0.15\linewidth}
\centering
Image 0\\[2pt]
\includegraphics[width=\linewidth]{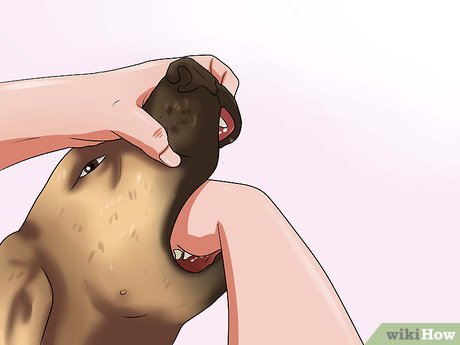}
\end{minipage}
\hfill
\begin{minipage}[t]{0.15\linewidth}
\centering
Image 1\\[2pt]
\includegraphics[width=\linewidth]{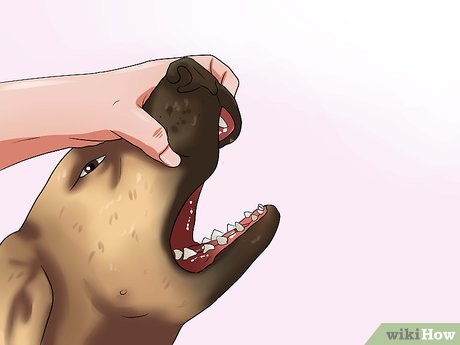}
\end{minipage}
\hfill
\begin{minipage}[t]{0.15\linewidth}
\centering
Image 2\\[2pt]
\includegraphics[width=\linewidth]{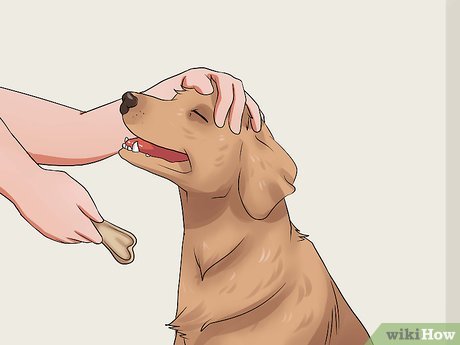}
\end{minipage}
\hfill
\begin{minipage}[t]{0.15\linewidth}
\centering
Image 3\\[2pt]
\includegraphics[width=\linewidth]{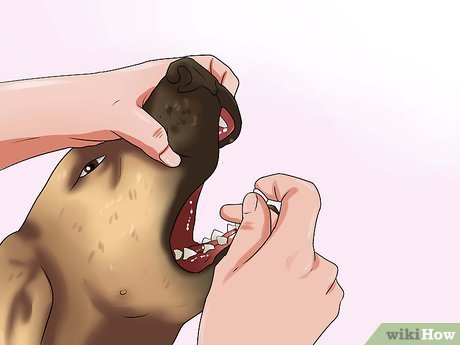}
\end{minipage}
\hfill
\begin{minipage}[t]{0.15\linewidth}
\centering
Image 4\\[2pt]
\includegraphics[width=\linewidth]{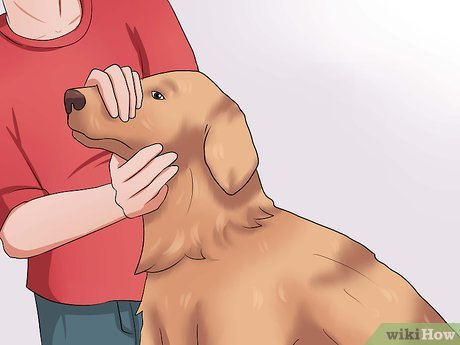}
\end{minipage}
\end{center}

\textbf{Ground Truth}

[0, 1, 3, 4, 2]

\noindent\rule{\linewidth}{0.1pt}

\textbf{Model output} 

\#\#\# Step-by-Step Reasoning
**Placeholder 1 (after “Force your dog to gently swallow the pill...”):**

Text: “Start opening his jaws from the back of the mouth with one hand. Then, use your second hand to lift from the roof of his mouth. Fold his lips over his teeth...”

This describes the initial step of opening the dog’s mouth. Image 1 shows a hand opening the dog’s mouth wide — teeth visible, jaw lifted — which matches the description of starting to open the jaws. Image 0 shows the mouth partially open but not as wide, and Image 3 shows the pill being placed. So Image 1 is best for this step.

→ **Image 1**

**Placeholder 2 (after “Open your dog's mouth wide, and place the medicine inside...”):**

Text: “Open your dog's mouth wide, and place the medicine inside. Try to put it as far back as possible.”

This is the action of placing the pill. Image 3 clearly shows a hand placing a pill into the dog’s open mouth, near the back of the throat. Image 0 shows the mouth closed or nearly closed, which is incorrect. Image 1 is just opening the mouth, not placing the pill. So Image 3 is the best match.

→ **Image 3**

**Placeholder 3 (after “Hold his mouth shut gently. Do this until your dog swallows...”):**

Text: “Hold his mouth shut gently... Blow gently on his nose... Rub your hand over the dog's throat...”

This describes holding the mouth shut and stimulating swallowing. Image 0 shows the dog’s mouth being held shut — the hand is covering the jaw, and the mouth is closed. This matches “hold his mouth shut gently.” Image 4 shows a person holding the dog’s head but not necessarily the mouth shut. So Image 0 is best.

→ **Image 0**

**Placeholder 4 (after “Give treats after your dog swallows...”):**

Text: “Give treats after your dog swallows... Use high value treats...”

This is about rewarding the dog. Image 2 clearly shows a hand giving a treat (a bone-shaped treat) to a happy-looking dog. This is the only image showing treat-giving. Perfect match.

→ **Image 2**

**Placeholder 5 (last one, after “Give treats...” — but actually, this is the final placeholder after the treat-giving paragraph):**

Wait — let’s recheck: there are 5 placeholders and 5 images. We’ve assigned:

- Placeholder 1 → Image 1
- Placeholder 2 → Image 3
- Placeholder 3 → Image 0
- Placeholder 4 → Image 2

That leaves Placeholder 5 and Image 4.

Let me verify once more:

Placeholder 1: "Start opening his jaws..." → Image 1: mouth being opened 

Placeholder 2: "Open your dog's mouth wide, and place the medicine inside." → Image 3: pill being placed 

Placeholder 3: "Hold his mouth shut gently." → Image 0: mouth held shut 

Placeholder 4: "Give treats..." → Image 2: treat being given 

Placeholder 5: after treat-giving, perhaps showing the dog is calm → Image 4: person holding dog gently 

Perfect.

Final Answer:
[1, 3, 0, 2, 4]

\noindent\rule{\linewidth}{0.1pt}

\textbf{Error Analysis} 

"error\_reason": "The model's prediction exhibits a severe global misalignment where the sequence of images does not match the chronological progression of the text instructions, shifting almost every assignment by one or more steps. Specifically, the model incorrectly assigns the 'mouth wide open' image (Image 1) to the first placeholder intended for the initial grip (Image 0), and swaps the final two steps by placing the 'treat' image (Image 2) at the 'hold mouth shut' placeholder and the 'hold mouth' image (Image 4) at the final 'treat' placeholder. This indicates a failure to track the state transitions across the full article context, compounded by misinterpreting subtle visual differences between similar states like initial grip versus holding shut.",

"primary\_error\_type": "Global Assignment Drift",

"secondary\_error\_types": [
  "Step-State Confusion",
  "Fine-Detail Miss"
]
\end{takeawaybox}

\clearpage

\begin{takeawaybox}{Error Case 6: Instruction Violation}

\textbf{Article Text (with placeholders):}

Sake in Edible Mount Fuji Cucumber Cups

This is the last, and simplest of three cocktail recipes in my Edible Cucumber Cup Cocktail series.
\vspace{-0.5em}
\begin{center}
[IMAGE\_PLACEHOLDER]
\end{center}
\vspace{-0.5em}
Fill 'em Up!Once you've hollowed your sake sized cups (for instructions on how to make the cucumber cups, click here ), using a clean baster, fill cups with your favorite kind of cold sake!  *sake is available at most places that wine and liquor are sold
\vspace{-0.5em}
\begin{center}
[IMAGE\_PLACEHOLDER]
\end{center}
\vspace{-0.5em}
Party Time!aaaand serve!
\vspace{-0.5em}
\begin{center}
[IMAGE\_PLACEHOLDER]
\end{center}
\vspace{-0.5em}
Applejack \& Cucumber Mint Cooler Recipe LinksFor the Applejack recipe, click here . For the Cucumber Mint Cooler recipe, click here .
\vspace{-0.5em}
\begin{center}
[IMAGE\_PLACEHOLDER]
\end{center}
\vspace{-0.5em}

\#\# Candidate Images (Image 0 to Image 3):

\begin{center}
\begin{minipage}[t]{0.20\linewidth}
\centering
Image 0\\[2pt]
\includegraphics[width=\linewidth]{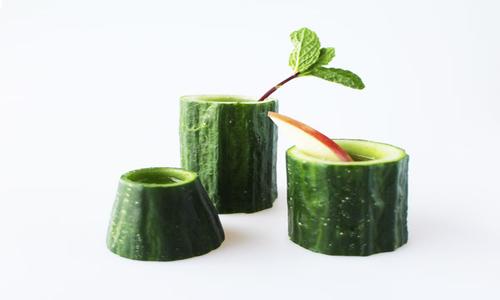}
\end{minipage}
\hfill
\begin{minipage}[t]{0.20\linewidth}
\centering
Image 1\\[2pt]
\includegraphics[width=\linewidth]{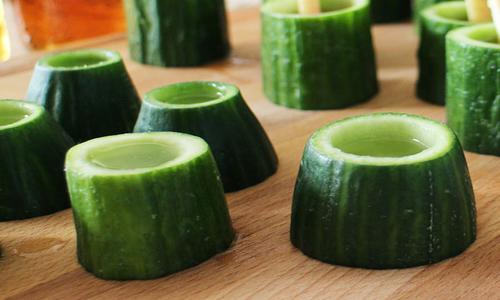}
\end{minipage}
\hfill
\begin{minipage}[t]{0.20\linewidth}
\centering
Image 2\\[2pt]
\includegraphics[width=\linewidth]{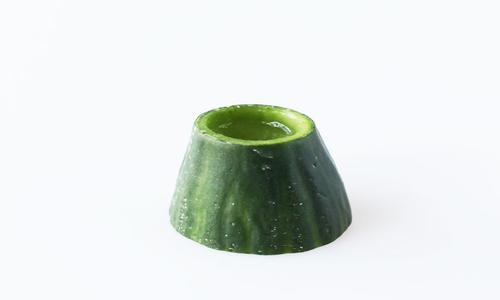}
\end{minipage}
\hfill
\begin{minipage}[t]{0.20\linewidth}
\centering
Image 3\\[2pt]
\includegraphics[width=\linewidth]{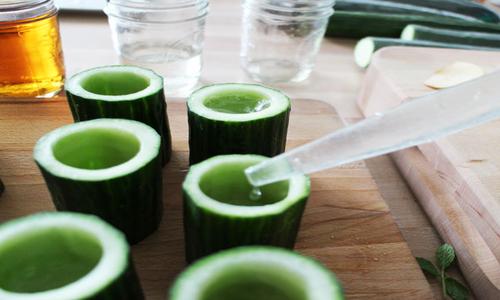}
\end{minipage}
\end{center}

\textbf{Ground Truth}

[2, 3, 1, 0]

\noindent\rule{\linewidth}{0.1pt}

\textbf{Model output} 

Let's break down the text and match it to the images.

1.  **First placeholder: "This is the last, and simplest of three cocktail recipes in my Edible Cucumber Cup Cocktail series."**
    - This is the introductory sentence. The context is about the "Edible Cucumber Cup Cocktail series" and the "last" recipe.
    - The image should show the finished product of this recipe. The text mentions "sake sized cups" and "fill 'em Up!", which implies the cups are filled with a drink.
    - Looking at the images, Image 0 shows three cucumber cups, one of which has a slice of apple (a fruit, possibly related to the "Applejack \& Cucumber Mint Cooler" mentioned later). Image 1 shows multiple cucumber cups, some with a drink inside. Image 2 shows a single cup. Image 3 shows a setup with a pipette, indicating the process of filling.
    - The description "Fill 'em Up!" and the mention of "sake" strongly suggest that the image for the first placeholder should show the cups filled with a drink, which is best represented by Image 1, as it shows the cups being filled.
    - Image 3 is the most relevant for the "filling" step, showing a pipette being used to pour liquid into a cucumber cup, which aligns perfectly with the "Fill 'em Up!" instruction.
    - Image 0 is a potential candidate, but it's a general shot of the finished product, not specifically the filling process.
    - Image 2 is too simple and does not show the filling action.
    - Therefore, the first placeholder (after "This is the last...") should be Image 3.

2.  **Second placeholder: "Fill 'em Up! Once you've hollowed your sake sized cups (for instructions on how to make the cucumber cups, click here ), using a clean baster, fill cups with your favorite kind of cold sake! *sake is available at most places that wine and liquor are sold"**
    - This text describes the action of filling the cucumber cups with sake. The key detail is "using a clean baster" and "fill cups with your favorite kind of cold sake".
    - The image should show the cups being filled. Image 3 clearly shows a pipette being used to pour liquid into a cucumber cup, which matches the "clean baster" description.
    - Image 1 shows cups that are already filled, but the filling action is not visible. Image 0 shows a cup with a slice of apple, not filled with sake. Image 2 is a single cup, not showing the filling process.
    - Therefore, the second placeholder should be Image 3.

3.  **Third placeholder: "Party Time! aaaand serve!"**
    - This text is a simple, enthusiastic closing. It says "Party Time!" and "serve!", indicating the final step after the filling.
    - The image should show the cups ready to be served. Image 0 shows the cups with a garnish, which is a common way to present a finished drink. Image 1 shows the cups with a drink, but not garnished. Image 2 shows a single cup, not garnished. Image 3 shows the cups being filled, not ready for serving.
    - The best image for "Party Time!" and "serve!" is the one that shows the cups ready to be served, which is Image 0.
    - Therefore, the third placeholder should be Image 0.

4.  **Fourth placeholder: "Applejack \& Cucumber Mint Cooler Recipe Links For the Applejack recipe, click here . For the Cucumber Mint Cooler recipe, click here ."
    - This text refers to recipes for Applejack and Cucumber Mint Cooler.
    - The image should be related to the "Applejack" recipe. The image should show the applejack drink in a glass or cup.
    - Looking at the images, Image 3 shows a wooden board with several cucumber cups, and in the background, there is a jar of what appears to be applejack (a golden liquid). This image is the most relevant for the "Applejack" recipe.
    - Image 1 shows multiple cucumber cups, but the drink is not applejack. Image 0 shows a cup with a slice of apple, not applejack. Image 2 is a single cup, not showing a drink.
    - Therefore, the fourth placeholder should be Image 3.

The final answer is [3, 3, 0, 3].

\noindent\rule{\linewidth}{0.1pt}

\textbf{Error Analysis} 

"error\_reason": "The model's prediction is fundamentally invalid because it violates the explicit constraint that each image index must be used exactly once, instead reusing Image 3 for three out of four placeholders. This structural failure results in a globally inconsistent assignment where Images 1 and 2 are completely ignored, breaking the logical narrative flow from preparation to serving. Additionally, the model exhibits step-state confusion by placing the active filling image at the introduction stage, failing to distinguish between the initial empty cup state and the subsequent liquid filling action.",
"primary\_error\_type": "Instruction Violation",
"secondary\_error\_types": [
  "Global Assignment Drift",
  "Step-State Confusion"
]
    
\end{takeawaybox}

\end{document}